\pdfoutput=1
\documentclass[11pt, a4paper, oneside, openany]{report}

\usepackage[T1]{fontenc}
\usepackage[utf8]{inputenc}
\usepackage{amsmath}
\usepackage{newpxtext}
\usepackage{newpxmath} 
\usepackage[scale=0.95]{sourcesanspro}
\usepackage{dsfont}
\usepackage{microtype}

\usepackage[a4paper, left=2.5cm, right=2.5cm, top=2.5cm, bottom=2.5cm]{geometry}
\usepackage{setspace}
\usepackage{bm}
\usepackage{graphicx}
\usepackage[export]{adjustbox}
\usepackage{booktabs}
\usepackage{multirow}
\usepackage[table]{xcolor}
\usepackage{enumitem}
\usepackage[font=small, labelfont=bf]{caption}
\usepackage{subcaption}
\usepackage{pifont}
\usepackage{pgfplots}
\pgfplotsset{compat=1.16}
\usepackage[nottoc]{tocbibind}

\newcommand{\cmark}{\ding{51}}%
\newcommand{\xmark}{\ding{55}}%

\graphicspath{{figures/}}

\PassOptionsToPackage{hyphens}{url}
\usepackage[hidelinks]{hyperref}

\usepackage[explicit]{titlesec}
\titleformat{\chapter}[display]
  {\fontsize{27}{30}\selectfont\sffamily\bfseries\filleft}{\thechapter}{0pt}{#1}
\titleformat{name=\chapter, numberless}[display]
  {\fontsize{27}{30}\selectfont\sffamily\bfseries\filleft}{}{0pt}{#1}
\titlespacing*{\chapter}{0pt}{0pt}{30pt}
\titleformat{\section}{\Large\bfseries\sffamily}{\thesection}{0.5em}{\MakeUppercase{#1}}
\titleformat{\subsection}{\large\bfseries\sffamily}{\thesubsection}{0.5em}{\MakeUppercase{#1}}
\titleformat{\subsubsection}{\sffamily}{\thesubsubsection}{0.5em}{\MakeUppercase{#1}}

\newcommand{\ttitle}{Where, How, and When: Query-Guided Grounding for Video Understanding}
\newcommand{\authorname}{Yogesh Kumar}
\newcommand{\supname}{Dr. Anand Mishra}
\newcommand{\degree}{Doctor of Philosophy}
\newcommand{\deptname}{Computer Science and Engineering}
\newcommand{\univname}{Indian Institute of Technology Jodhpur}
\newcommand{\thesisdate}{October 2026}

\hypersetup{
  pdftitle={\ttitle},
  pdfauthor={\authorname},
  pdfsubject={PhD Thesis, \univname},
  pdfkeywords={video understanding, visual grounding, few-shot learning, one-shot object localization, referring relationships, video moment retrieval}
}

\begin{document}

\pagenumbering{roman}
\begin{titlepage}
\sffamily
\setstretch{1.0}
\begin{flushright}
{\Huge\bfseries \ttitle\par}
\vfill
{\Large\itshape A thesis submitted by\par}
\vspace{0.4em}
{\huge\bfseries \authorname\par}
\vfill
{\Large\itshape in partial fulfillment of the requirements for the award of the degree of\par}
\vspace{0.4em}
{\huge\bfseries \degree\par}
\vfill
{\Large\itshape under the supervision of\par}
\vspace{0.4em}
{\Large\bfseries \supname\par}
\vfill
\includegraphics[width=0.25\textwidth]{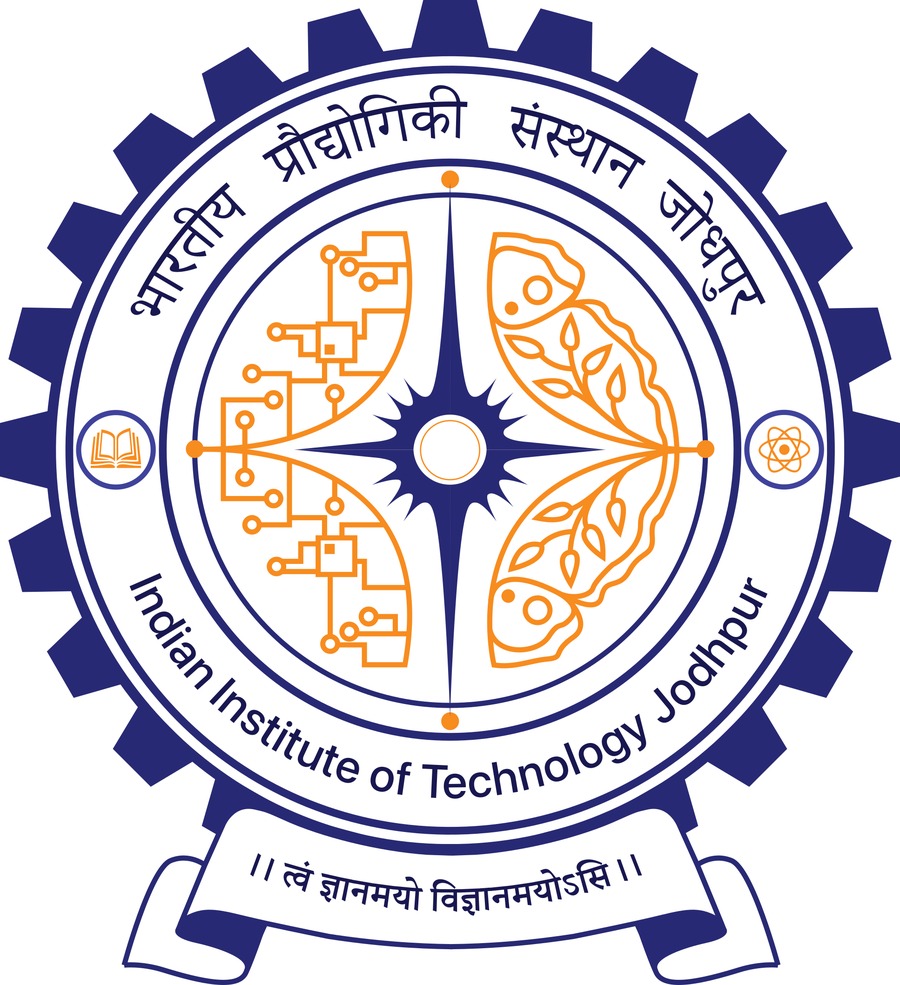}\par
\vspace{1em}
{\Large\bfseries \univname\par}
{\Large\bfseries Department of \deptname\par}
\vspace{0.4em}
{\Large\itshape \thesisdate\par}
\end{flushright}
\end{titlepage}

\setcounter{page}{2}
\chapter*{Abstract}
\addcontentsline{toc}{chapter}{Abstract}

Understanding video content requires reasoning across multiple levels of granularity, from identifying individual objects to interpreting their relationships and localizing specific temporal actions or moments. This thesis explores three distinct grounding tasks that correspond to fundamental questions in query-guided video understanding: \textit{Where} key objects are present (object-level grounding), \textit{How} these objects interact (relationship-level grounding), and \textit{When} specific actions occur (moment-level grounding). These tasks collectively enable a structured interpretation of dynamic scenes in videos, progressing from fine-grained object localization to coarse temporal localization.

For object-level grounding, we present QDETRv (Query-guided Detection Transformer for Videos), a novel extension of Detection Transformer that enables one-shot localization of unseen objects in videos. QDETRv employs (i) query-context fusion, where query features are integrated with contextual video frames to enable effective temporal reasoning; (ii) memory module to maintain temporal consistency across frames; and (iii) unsupervised pretraining strategy using synthetic object trajectories with recurrent queries and feature-reconstruction objectives for a better initialization, and thereby achieves substantial gains over competitive baselines. Taking it forward, to address the limitations of QDETRv, we propose the Temporal Object-Aware Vision Transformer (TOViT), which leverages a language-aligned, large-scale pre-trained open-set vision encoder. This enhances semantic discrimination, enabling better separation of visually similar objects through rich linguistic priors. Additionally, TOViT introduces a filtering mechanism that propagates high-confidence object features across frames, allowing temporal consistency to support the detection of small and partially visible objects.

For relationship-level grounding, we propose few-shot referring relationships in videos  where given a query visual relationship, i.e. (subject, predicate, object) and a target video, the task is to localize the entities using only a few support videos sharing the same predicate. Our method formulates this as an optimization over a T-partite random field, with candidate bounding boxes as vertices across T frames. A relation network with query-conditioned translational embeddings is trained via episodic meta-learning. Two aggregation techniques, Global Semantic Aggregation (GSA) and Local Localization Aggregation (LLA), incorporate temporal context, and belief propagation-based message passing produces subject-object trajectories. Experiments show strong performance, effectively addressing the long-tail distribution of visual relationships.

For moment-level grounding, we propose Moment Alignment Transformer (MATR) for video-to-video moment retrieval. MATR employs dual-stage Dynamic Time Warping for pre- and post-fusion alignment within a transformer, capturing fine-grained temporal correlations. Learnable query vectors guide moment localization via classification and boundary prediction heads. A self-supervised pretraining strategy with temporal augmentations further enhances generalization. We further extended MATR to support text-to-video moment retrieval and achieved comparable, if not better performance to the state-of-the-art methods.

In short, this thesis progressively advances the understanding of where, how, and when in query-guided video comprehension by proposing a series of principled solutions spanning spatial, relational, and temporal grounding. We believe that the ideas and frameworks presented here will open up promising research avenues in various application domains such as sports analytics, manufacturing, and skill understanding, among others.

\chapter*{Acknowledgements}
\addcontentsline{toc}{chapter}{Acknowledgements}

The completion of this doctoral dissertation represents not just a personal milestone but the culmination of years of collaboration, support, and encouragement from numerous individuals that have profoundly shaped my academic journey.

First and foremost, I extend my deepest gratitude to my PhD advisor, Dr. Anand Mishra, whose guidance and mentorship have been essential in shaping both this research and my development as a scholar. His pursuit of excellence, dedication to rigorous research methodology, and promotion of innovative thinking have been a constant source of inspiration throughout my doctoral journey. Dr. Mishra's ability to balance constructive criticism with encouragement, his open-door policy for discussions ranging from technical challenges to career guidance have been invaluable. His hard work and dedication to maintaining the highest standards of research quality continue to motivate me, and I am grateful for the excellent opportunities he provided that allowed me to excel across multiple dimensions of academic research. I fondly remember how, during my first publication, Dr. Mishra spent late night with me in the lab, patiently guiding me through the process of clear academic writing, thoughtful experimental design, and effective presentation of research. I am also thankful for his generous support for my participation in national and international conferences, which greatly enriched my academic exposure. Furthermore, his efforts in facilitating collaborations with industry partners provided me with invaluable real-world insights and broadened the practical relevance of my research.

I am deeply grateful to my PhD committee members, Dr. Suman Kundu, Dr. Ravi Kiran, and Dr. Angshuman Paul, for their thoughtful engagement with my research and their invaluable suggestions that helped refine this work. I extend special appreciation to Dr. Manish Gupta for his consistent feedback and guidance during our weekly meetings and paper writing. I am also thankful to Dr. Anutosh Maitra, Dr. Sowmya Rasipuram, and Roshni Ramnani for their collaboration and constructive feedback during our weekly discussions.

This research would not have been possible without the financial support of our funding agencies and industry collaborators. I acknowledge the University Grants Commission (UGC) for providing the fellowship that supported my doctoral studies. I am particularly thankful to Microsoft and Accenture for their research grants and collaborative partnerships that provided computing resources for our lab and also offered valuable insights into real-world applications.

I extend my sincere appreciation to the faculty, staff, and administration of IIT Jodhpur for creating an environment conducive to high-quality research and learning. The institution's commitment to enabling innovation and providing state-of-the-art research facilities has been fundamental to the success of this work. I am particularly grateful for the technical support staff (Rimpesh) who ensured that our computational infrastructure remained robust and accessible throughout the demanding phases of this research.

My doctoral journey has been greatly enriched by the opportunity to work alongside talented students and research associates in our lab. I have been fortunate to collaborate with several undergraduate and research associates, including Prajwal, Saswat, Shreya, Abu, Kranti, and Uday. The doctoral journey, with its inevitable challenges and moments of uncertainty, has been made more manageable and enjoyable through the friendship with my fellow graduate students. I am grateful to my friends, fellow graduates and lab-mates, Abhirama, Rishab, Rekhansh, Vishnu, Kartik, Surbhi, Somraj, Abhishek, Niveddita, Neelu, Akshita, Satyam, Prince, Pushkar, and Anoop who have been my companions through both the challenging and successful moments of this journey. Our discussions, ranging from technical debates to philosophical conversations about research and life, have provided both intellectual stimulation and emotional support.

Perhaps most importantly, I owe my deepest gratitude to my family, whose support, sacrifice, and belief in my aspirations have been the foundation upon which this achievement stands. My brother deserves special recognition for shouldering additional family responsibilities, allowing me to focus entirely on my studies. My mother and sister have been pillars of emotional support throughout this journey, providing comfort during difficult times and celebrating each milestone along the way. Their unconditional love, patience during my long absences, and faith in my abilities have sustained me through the challenging phases of this work.

I also wish to acknowledge my young nieces, Pankhudi and Miradhya, whose joyful presence during family visits provided much needed respite from the intensity of research work. Their innocent curiosity and their ability to bring laughter and lightness to family gatherings have been precious gifts that reminded me of the importance of maintaining balance in life.

At this point, I also wish to remember my late father, whose teachings continue to guide me even in his absence. His belief in never losing hope and his lessons on perseverance during difficult times have been a constant source of strength throughout my doctoral journey. The values he instilled in me have not only shaped my character but also carried me through the most challenging phases of my life.

Finally, I acknowledge that this doctoral dissertation represents not just my individual effort, but the collective contribution of all these individuals and institutions. While my name appears on the cover, the ideas, insights, and innovations presented within these pages have been shaped by many interactions, discussions, and collaborations. Any errors or shortcomings in this work remain entirely my responsibility, while its contributions to the field of computer vision and video understanding reflect the collective wisdom and support of this remarkable community of scholars, friends, and family members who have accompanied me on this transformative journey.

\begin{flushright}
\emph{Yogesh Kumar}
\end{flushright}

\begingroup
\setlength{\parskip}{0pt}
\tableofcontents
\listoffigures
\listoftables
\endgroup

\clearpage
\pagenumbering{arabic}
\setcounter{page}{1}

\chapter{Introduction}
\label{ch:introduction}

The digital era has witnessed explosive growth in the generation of video data. Recent studies show that video accounts for more than 80\% of global internet traffic~\cite{cisco2023}, with platforms like YouTube receiving more than $20$ million uploads per day\footnote{\url{https://blog.youtube/press/}}. Surveillance systems generate petabytes of footage daily~\cite{surveillance2023}, while billions of smartphone users continuously create and share videos. This massive volume of visual data presents unprecedented opportunities for insight in domains such as sports analytics, manufacturing, skill understanding, healthcare, education, security, and scientific research. However, this abundance also presents major challenges. The sheer scale of video content far exceeds what humans can view and analyze manually. To enable scalable and timely understanding, there is a pressing need for automatic tools that can flexibly interpret video content, adapt to novel scenarios, and extract user-relevant insights with minimal supervision.

Crucially, such tools must go beyond generic summarization and support targeted video understanding, where users seek specific insights into particular tasks or events rather than exhaustive analysis of entire videos. For instance, in the sports domain, one may wish to identify a specific object such as a player, a football, or a goalpost; an interaction like a player kicking the football; or a moment/action such as a shot on goal or a successful save. These examples underscore the need for query-based grounding systems that can parse dynamic gameplay on demand and retrieve user-defined information with minimal supervision and effort.

\begin{figure*}[t!]
    \centering
      \includegraphics[width=\textwidth]{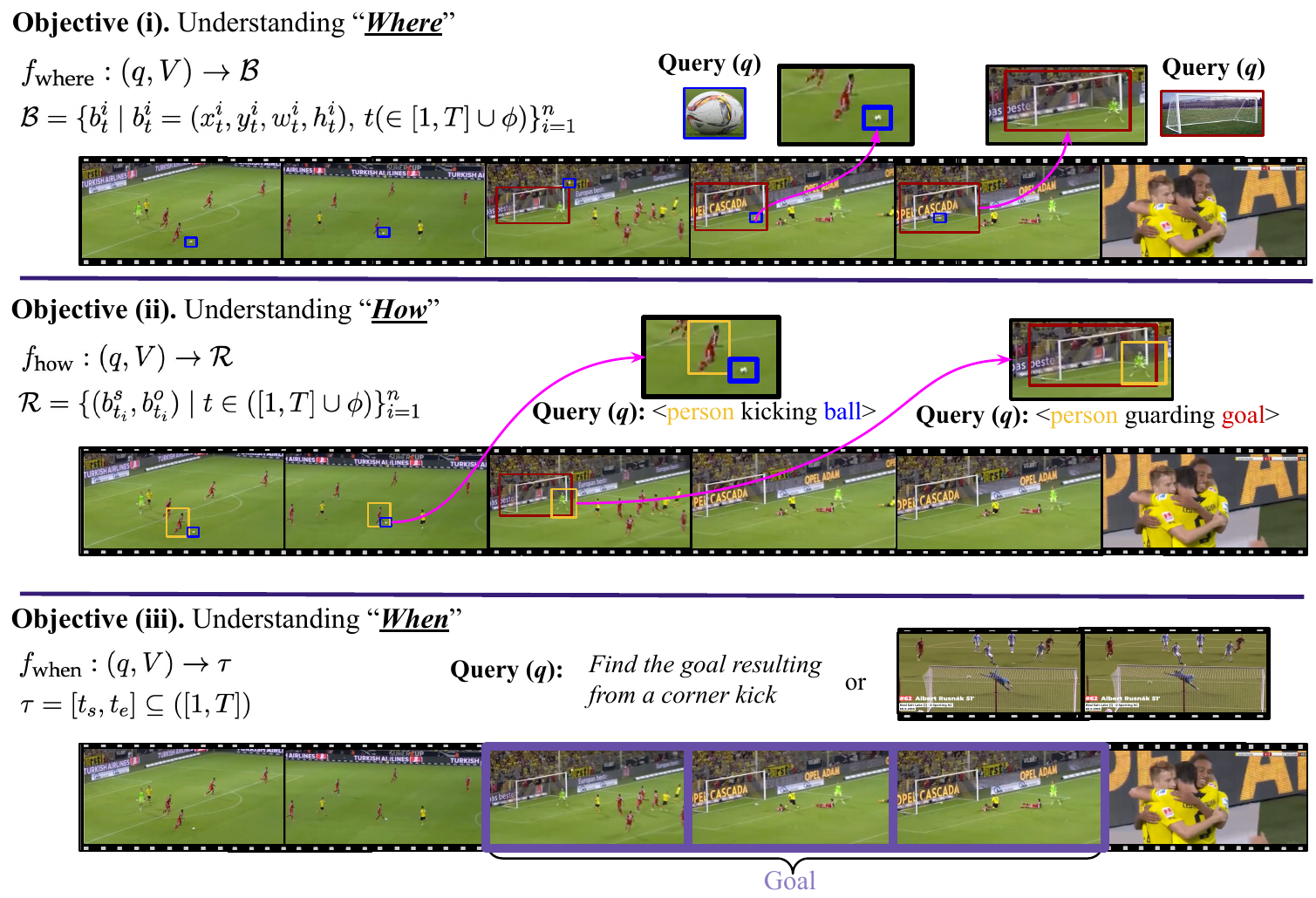}
    \caption[Thesis goal]{\label{fig:t_goal} \textbf{Thesis Objectives}.
    This work aims to advance query-guided video understanding by addressing three core questions: \textit{where} objects are present, \textit{how} they interact, and \textit{when} specific actions occur. Consider a football match: understanding the play requires (i) identifying key entities such as the ball, goalposts, and the players (where important objects are), (ii) recognizing interactions like ``player kicking the ball to a teammate" (how objects interact), and (iii) localizing key events such ``goa" moment (when critical actions occur). Together, these tasks enable structured interpretation of complex, fast-paced sports scenes, moving beyond frame-level recognition toward temporally grounded, interaction-aware understanding.
    }
\end{figure*}

In this thesis we study query-based grounding tasks and introduce methods that enable flexible and targeted interpretation of video content with minimal supervision, even in open-world settings. Toward this end, the goal of this thesis is illustrated in Figure~\ref{fig:t_goal}. Consider a video of a football match. Understanding the unfolding play requires addressing three core questions while handling potentially unseen players, formations, or gameplay scenarios. Specifically: (\textbf{Objective (i). Understanding where}) identifying key entities such as players, the ball, and goalposts, even when player appearances, jerseys, or camera perspectives differ from training examples. This objective is achieved through query-guided object localization using visual queries; (\textbf{Objective (ii). Understanding how}) recognizing interactions such as ``player kicking the ball”, even when the interaction patterns or strategies differ from those observed during training, which is addressed through relationship grounding using visual-relationship queries; and (\textbf{Objective (iii). Understanding when}) detecting key temporal events such as a ``goal" even when the exact action instance or context was not seen during training, handled via video moment retrieval using video queries.

This query-driven formulation enables the system to interpret video content based on user-defined interests, eliminating the need to manually view entire videos. Each task is designed to handle novel visual elements that may not have been encountered during training, ensuring adaptability in open-worldd settings. Collectively, the three grounding tasks, object-level, relationship-level, and moment-level, capture the multidimensional nature of video understanding: objects serve as the foundational elements of scenes, relationships encode meaningful interactions that define activities, and temporal grounding pinpoints when key events unfold. By tackling these objectives through targeted queries, our approach empowers machines to retrieve specific user-relevant information from unstructured videos with minimal supervision. This provides scalable and user-centric video interpretation systems that flexibly generalize to unseen concepts and deliver actionable insights in dynamic environments.

The following sections present a discussion of the key challenges, prior work, and our core contributions.

\section{Challenges}
While addressing the proposed-task towards query-guided video understanding we tackled following challenges.

\subsection{The Scalability Challenge}

Many video understanding methods face a major limitation: they rely heavily on detailed annotations for every possible scenario. Traditional supervised approaches demand large-scale training datasets that cover all object types~\cite{bertasius2018object}, relationships~\cite{xiao2020visual}, actions, and moments~\cite{ji2020action} that may arise during deployment or test time making them difficult to scale in real-world settings.

This challenge is particularly pronounced when systems are expected to generalize from limited supervision. In many practical cases, models must detect unfamiliar objects, interpret novel interactions, or localize complex actions using only a handful of examples. For instance, identifying an unexpected road obstacle, understanding a rare interaction like ``plane flying above truck,'' or retrieving fine-grain actions such as ``bicycle kicks'' from a footwall match video. The annotation burden becomes even more severe due to the compositional nature of visual understanding, where the number of possible object-relationship combinations grows exponentially.

To address the scalability challenge, the chapters~\ref{chapone},  \ref{chaptwo}, and \ref{chapthree} focus on methods that can be learned with one or a few examples. Instead of relying on exhaustive annotations, we develop methods to generalize from one or a few examples, allowing scalable models. This shift from annotation-heavy pipelines to adaptive learning frameworks provides a step toward building practical, flexible video AI systems.

\subsection{Open-Set Generalization}

Real-world environments are inherently open-set: new objects, interactions, and events frequently emerge without prior labels or limited training instances. Systems designed under closed-set assumptions, where all categories are known in advance, struggle to adapt, especially in dynamic domains like autonomous driving, surveillance, and content retrieval. This mismatch between training and deployment distributions creates a critical need for open-set generalization, where models must recognize and reason about novel concepts at test time.

In video understanding, this challenge is amplified by the temporal and compositional complexity of visual scenes. Unlike static images, videos require consistent interpretation across frames, making it harder to detect and track unseen categories, relationships, or actions. Traditional pipelines with fixed category heads or frame-wise processing often fail to maintain temporal coherence or calibrate unknowns. Recent open-vocabulary approaches~\cite{owl,video-owlvit23} leverage language-aligned encoders to extend recognition to novel classes, but typically lack object-aware temporal modeling, resulting in fragmented or inconsistent detections for unseen concepts.

To address these limitations, chapters~\ref{chapone}, ~\ref{chaptwo}, and~\ref{chapfour} propose query-conditioned formulations that enable open-set generalization in video tasks. Our methods are designed to adapt for unseen examples at test time while leveraging temporal structure for robust grounding.

\subsection{Beyond Static Analysis}

Moving from static image understanding to video analysis introduces a new dimension of complexity, time. Unlike images, videos contain evolving scenes where objects move, interact, and change appearance across frames. This temporal axis brings several challenges; models must handle motion blur, occlusions, deformations, and viewpoint shifts while maintaining consistent predictions over time. Frame-wise methods that ignore temporal context often suffer from fragmented tracks and unstable boundaries, especially when objects are briefly missed or partially visible.

Effective video understanding requires spatiotemporal reasoning: systems must integrate information across frames to preserve continuity and suppress noise. For localization and detection, temporal fusion and cross-attention mechanisms help propagate high-confidence signals and maintain object identity. For temporal action/moment grounding, precise temporal alignment is essential, must match query and target sequences despite variations in speed, duration, and execution style, placing accurate start–end boundaries rather than relying on isolated frame-level matches.

To address these challenges, our work incorporates video-specific architectural designs that go beyond per-image processing. We introduce temporal cross-attention modules, memory-based fusion strategies (chapter~\ref{chapone}), and sequence-level alignment objectives (chapter~\ref{chapfour}) that enable robust spatio-temporal localizations and retrieval across time. These components allow our models to maintain coherence, adapt to dynamic content, and deliver temporally consistent predictions, advancing video understanding beyond static analysis.

\section{Prior Works}

\subsection{Object Detection and Grounding}

Object detection and localization in images have evolved from traditional CNN-based approaches like Faster R-CNN~\cite{FasterRCNN}, which use region proposal networks for candidate generation, and YOLO~\cite{zhang2021vit}, which formulates detection as a single-stage regression problem for real-time performance, to transformer-based methods like DETR~\cite{detr} that treat detection as a set prediction task. These methods typically rely on predefined object categories and extensive supervised training on large datasets~\cite{FasterRCNN,zhang2021vit}. Query-based object localization represents a paradigm shift, using visual queries such as single images of target objects to localize instances in target scenes. Methods like CoAE~\cite{hsieh2019one} employ co-attention mechanisms to establish correspondences between query images and target regions, while OS2D~\cite{osokin2020os2d} provides versatile one-stage detection, enabling detection of novel object categories without extensive retraining.

Extending object detection from images to videos introduces significant additional challenges beyond static image detection. The temporal nature of videos creates complexities, including appearance variations due to illumination changes, motion blur, viewpoint shifts, and occlusions where objects disappear and reappear across frames. Videos require temporal consistency to maintain coherent object identities and smooth trajectories across consecutive frames. Many existing methods process frames independently or use simple temporal aggregation that fails to capture complex object dynamics effectively~\cite{MEGA,DBLP:journals/corr/abs-2208-09686}. In this thesis (chapter~\ref{chapone}), we developed a query-guided detection framework~\cite{Kumar2024} specifically designed for video object localization using image queries, incorporating cross-attention mechanisms that capture information from neighboring frames to enable robust localization while maintaining temporal consistency.

\subsection{Visual Relationship Detection and Grounding}

Visual relationship detection in images focuses on understanding interactions between objects by identifying subject-predicate-object triplets such as ``person riding bicycle" or ``bird flying above car". Early approaches concentrated on spatial relationships and human-object interactions, with scene graphs introduced to represent multiple visual relationships in structured formats. Traditional methods~\cite{liang2018visual,yao2018exploring} rely on strong supervision with extensive annotations for every possible subject-predicate-object combination, limiting scalability to diverse relationship types. Query-based relationship grounding~\cite{krishna2018referring}, particularly referring relationships, represents a more flexible paradigm where systems must localize subject and object entities given specific relationship queries, typically using iterative message-passing mechanisms and language priors.

Extending visual relationship detection and grounding from images to videos introduces substantial temporal complexities beyond static image analysis~\cite{xiao2020visual,yang2022tubedetr}. Videos contain dynamic relationships that evolve over time, such as temporal predicates like ``chasing" or ``following" that have limited understanding in static images. Objects involved in relationships undergo appearance changes, occlusions, and motion patterns that must be tracked consistently across frames, while maintaining relationship consistency to avoid temporal discontinuities. Recent video relationship detection methods have incorporated transformer architectures and temporal attention mechanisms~\cite{yang2022tubedetr}, but most struggle with novel relationship types not seen during training. In this thesis (chapter~\ref{chapthree}), we address few-shot referring relationships in videos, where the model localizes subject and object entities corresponding to unseen relationship queries using only a few support examples, formulating this as an optimization problem over a temporal random field with belief propagation.

\subsection{Video Action or Moment Grounding}
Video moment retrieval majorly focused on text-based approaches that locate temporal segments in videos matching natural language descriptions~\cite{moment_detr,QD-DETR}. Early methods employed CNN-based video encoders combined with RNN-based text encoders~\cite{FFI-SRM,GDP}, using simple alignment mechanisms to match textual queries with video content. These approaches often struggled with complex semantic understanding and temporal reasoning, relying on basic feature concatenation or attention mechanisms. The introduction of transformer architectures revolutionized this field, with methods like Moment-DETR~\cite{moment_detr}, QD-DETR~\cite{QD-DETR}, and UniVTG~\cite{univtg} leveraging self-attention and cross-attention mechanisms to achieve better alignment between text queries and video segments, significantly improving localization accuracy through end-to-end learning.

However, text-based approaches face inherent limitations when describing complex visual moments, temporal dynamics, or nuanced actions that are difficult to articulate precisely in natural language. This challenge motivated the development of video-to-video moment retrieval (Vid2VidMR), where actual video clips serve as queries instead of text descriptions. Video-based approaches like GDP~\cite{GDP}, FFI+SRM~\cite{FFI-SRM}, and SRL~\cite{srl} enable more intuitive querying by allowing users to specify target moments using videos, particularly valuable for complex activities that are hard to describe verbally. These methods face unique challenges in aligning video sequences and capturing temporal correlations between query and target videos, requiring sophisticated temporal modeling to handle variations in motion, appearance, and timing. Our work (chapter~\ref{chapfour})~\cite{kumar2025matr,kumar2026matrcviu} has explored dual-stage sequence alignment, self-supervised pre-training objectives, and transformer-based architectures specifically designed for video-to-video alignment to improve performance in this challenging task.

\subsection{Transformer-Based Video Understanding}
Transformer architectures~\cite{vaswani2017attention} were originally introduced for natural language processing, utilizing self-attention mechanisms to model long-range dependencies without recurrent connections. Their success in NLP motivated adaptation to computer vision, starting with Vision Transformer (ViT)~\cite{dosovitskiy2020transformers} for image classification by treating image patches as tokens. DETR~\cite{detr} further revolutionized object detection by formulating it as a set prediction problem using transformer encoder-decoder architectures, eliminating hand-crafted components like non-maximum suppression. Adapting transformers for video understanding introduced unique temporal modeling challenges beyond static image analysis, requiring mechanisms to capture relationships across frames while maintaining computational efficiency.

Video-specific transformer variants~\cite{QD-DETR,moment_detr} incorporated spatiotemporal attention mechanisms and cross-attention to align information between different temporal segments or between query and target video content. Methods like Trans-VOD~\cite{DBLP:journals/corr/abs-2201-05047} addressed video challenges by designing attention patterns that capture both short-term motion and long-term temporal dependencies while handling variable video lengths. Vision-language transformers pretrained on large-scale image-text datasets, such as Video-OWLViT~\cite{video-owlvit23}, a CLIP-based model, have proven particularly effective for video tasks requiring semantic understanding of novel concepts, transferring rich semantic knowledge for few-shot and open-vocabulary understanding. In this thesis, we leverage transformer architectures through cross-attention mechanisms that capture temporal context (chapter~\ref{chapone}) from neighboring frames, dual-stage sequence alignment for video-to-video matching (chapter ~\ref{chapfour}), and query-guided attention that enables detection and localization of novel objects or relationships using minimal supervision.

\begin{figure*}[t!]
    \centering
      \includegraphics[width=0.83\textwidth]{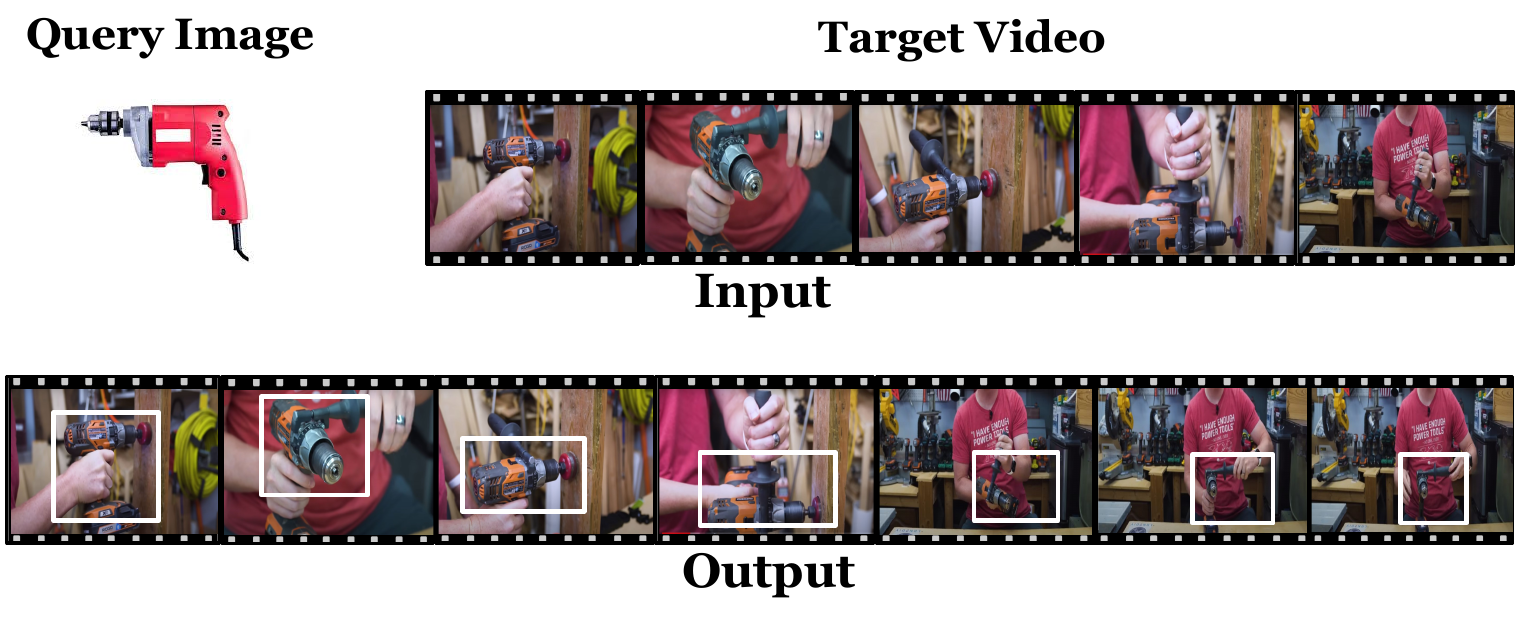}
    \caption[One-shot object localizations task]{\label{fig:aaai24_goal} Given an unseen query image and a target video, the model is expected to localize all instances of objects present in the query image on the target video.}

\end{figure*}

\section{Thesis Objectives and Contributions}

\subsection{Objective-1}

\textbf{\textit{Understanding ``Where''}: Develop a query-guided approach for object grounding that operates in open-set settings using one or a few visual examples.}

\noindent \textbf{Query-Guided Object Grounding.}
Given a visual query $q$ (e.g., an image or visual embedding of an object) and a target video
$V = \{f_1, f_2, \ldots, f_T\}$ consisting of $T$ frames, the goal of query-guided object localization is to determine a set of spatio-temporal regions
\[
\mathcal{B} = \{b_t^i \mid b_t^i = (x_t^i, y_t^i, w_t^i, h_t^i), \, t (\in [1, T] \cup \phi)\}_{i=1}^{n},
\]
that correspond to the visual entity represented by $q$, where $n$ represents the number of instances of the query object. Formally, the task can be viewed as learning a function
\[
f_{\text{where}}: (q, V) \rightarrow \mathcal{B},
\]
where $f_{\text{where}}$ localizes the queried object across frames as shown in Figure~\ref{fig:aaai24_goal}.

Unlike conventional object grounding systems limited to a fixed set of categories, our aim is to identify any visual entity specified through a query, thereby enabling open-world generalization. The problem becomes particularly challenging in videos, where objects exhibit temporal dynamics, occlusions, and appearance variations, requiring consistent localization across frames with little to no task-specific supervision.

To address this, we study the problem in two related settings: \textit{one-shot} and \textit{few-shot} video object localization. In the one-shot setting, the model receives a single query image and localizes the corresponding object in the video. In contrast, the few-shot setting provides a small support set of examples that guide the localization of novel object instances.

For the one-shot scenario, we introduce QDETRv (Query-Guided DETR for Videos)~\cite{Kumar2024} (chapter~\ref{chapone}), a transformer-based architecture that extends DETR with temporal modeling  capabilities. QDETRv employs (i) query-context fusion, where query features are integrated with contextual video frames to enable effective temporal reasoning; (ii) memory module to maintain temporal consistency across frames; and (iii) unsupervised pretraining strategy using synthetic object trajectories with recurrent queries and feature-reconstruction objectives, and thereby achieves substantial gains over competitive baselines. Taking it forward, we extended QDETRv for few-shot object grounding.

Building on this, we proposed a temporal object-aware Vision Transformer~\cite{kumar2026fsvit} (chapter~\ref{chaptwo}) that addresses QDETRv’s limitations by leveraging language aligned large-scale pre-trained open-set detectors. It integrates semantic knowledge from vision-language models like OWL-ViT to enable effective knowledge transfer and robust object generalization across time. The method introduces a filtering mechanism that propagates high-confidence object features across frames, enabling efficient progression and reduced noise. It integrates feature interaction between current frames and past detections to maintain temporal coherence without explicit object-tube proposals. Parallel classification and detection heads are optimized for few-shot settings, achieving state-of-the-art performance.

Together, these contributions advance the understanding of \textit{``where''} objects are located in videos, establishing a robust foundation for open-world and query-driven object grounding under limited supervision.

\begin{figure*}[t!]
    \centering
      \includegraphics[width=0.8\textwidth]{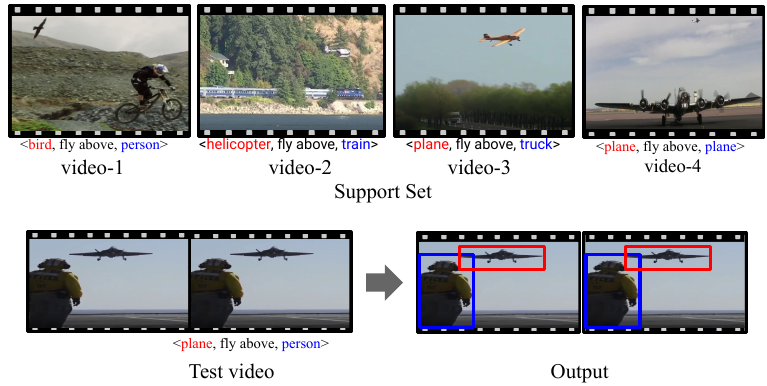}
    \caption[Few-shot Referring Relationship task]{\label{fig:cvpr_goal} Given a query visual relationship as $<$subject, predicate, object$>$ and a test video, our goal is to localize the subject and object on the test video with the help of a support set containing a few videos with the same predicate. In this example, the goal is to spatio-temporally localize the \emph{plane} (subject) and \emph{person} (object) that are connected via \emph{fly above} (predicate), with the help of a support set containing only four videos sharing the predicate \emph{fly above}. It should be noted here that \emph{fly above} is unseen during training. We refer to this problem as \emph{few-shot referring relationship in videos}. }
\end{figure*}

\subsection{Objective-2}

\textbf{\textit{Understanding ``How''}: Develop a query-guided method to ground unseen object interactions (visual relationships) spatio-temporally using few examples.}

\noindent \textbf{Query-Guided Relationship Grounding.}
Given a visual relationship query $q = \langle s, p, o \rangle$, where $s$ and $o$ denote subject and object visual entities, and $p$ represents a predicate describing their interaction, and a target video $V = \{f_1, f_2, \ldots, f_T\}$ consisting of $T$ frames,  the goal of relationship grounding is to identify spatio-temporal trajectories
\[
\mathcal{R} = \{(b_{t_i}^s, b_{t_i}^o) \mid t \in ([1, T] \cup \phi)\}_{i=1}^n
\]
such that the localized regions correspond to entities engaged in interaction $p$, where $n$ is the number of instances of the query visual relation. Formally, the task can be defined as learning a mapping
\[
f_{\text{how}}: (q, V) \rightarrow \mathcal{R},
\]
where $f_{\text{how}}$ grounds relationships in videos by jointly reasoning over spatial and temporal dependencies of entities, as illustrated in Figure~\ref{fig:cvpr_goal}.

While object localization answers \textit{``where''} entities are, understanding \textit{``how''} they interact provides a richer, more compositional understanding of dynamic scenes. For example, the relationships ``person near bicycle'' and ``person riding bicycle'' represent semantically distinct interactions despite involving the same entities. Modeling such relational semantics is crucial for high-level reasoning and event understanding. However, existing relationship detection methods rely on exhaustive annotation of subject-predicate-object triplets~\cite{xiao2020visual}, which severely limits scalability and generalization in realistic, open-world video settings. In videos, these challenges are further amplified by temporal evolution, occlusions, and discontinuities, as relationships appear, vanish, and transform over time.

To address these challenges, we propose the task of few-shot referring relationship grounding in videos~\cite{Kumar023-2} (\ref{chapthree}), where the goal is to ground novel subject-object interactions with limited supervision. The problem is formulated as an optimization over a $T$-partite random field, where $T$ denotes the number of frames in the video. Each vertex represents a candidate subject-object pair, and the objective function combines frame-level detection potentials with relationship similarity potentials learned through episodic meta-training. This design allows the model to generalize to unseen predicates while maintaining temporal coherence across frames.

Our approach introduces query-conditioned translational relationship embeddings that capture semantic interactions between entities in a structured embedding space. These embeddings are enhanced using two complementary attention-based aggregation mechanisms: (i) Global Semantic Aggregation (GSA), which fuses I3D features across all frames using learnable attention weights to incorporate long-range contextual cues, and (ii) Local Localization Aggregation (LLA), which focuses on spatial and ROI features from neighboring frames within a temporal window, thereby addressing occlusions and partial visibility.

A meta-learned Relation Network computes similarity scores between support and query relationships, enabling robust matching even for unseen predicates. The resulting optimization is efficiently solved via belief propagation-based message passing on the random field, producing consistent and temporally aligned subject–object trajectories.

Together, these contributions advance the understanding of ``how'' objects interact in videos, bridging the gap between object-level detection and event-level reasoning, and enabling scalable few-shot grounding of visual relationships under dynamic video conditions.

\begin{figure*}[t!]
    \centering
      \includegraphics[width=0.9\textwidth]{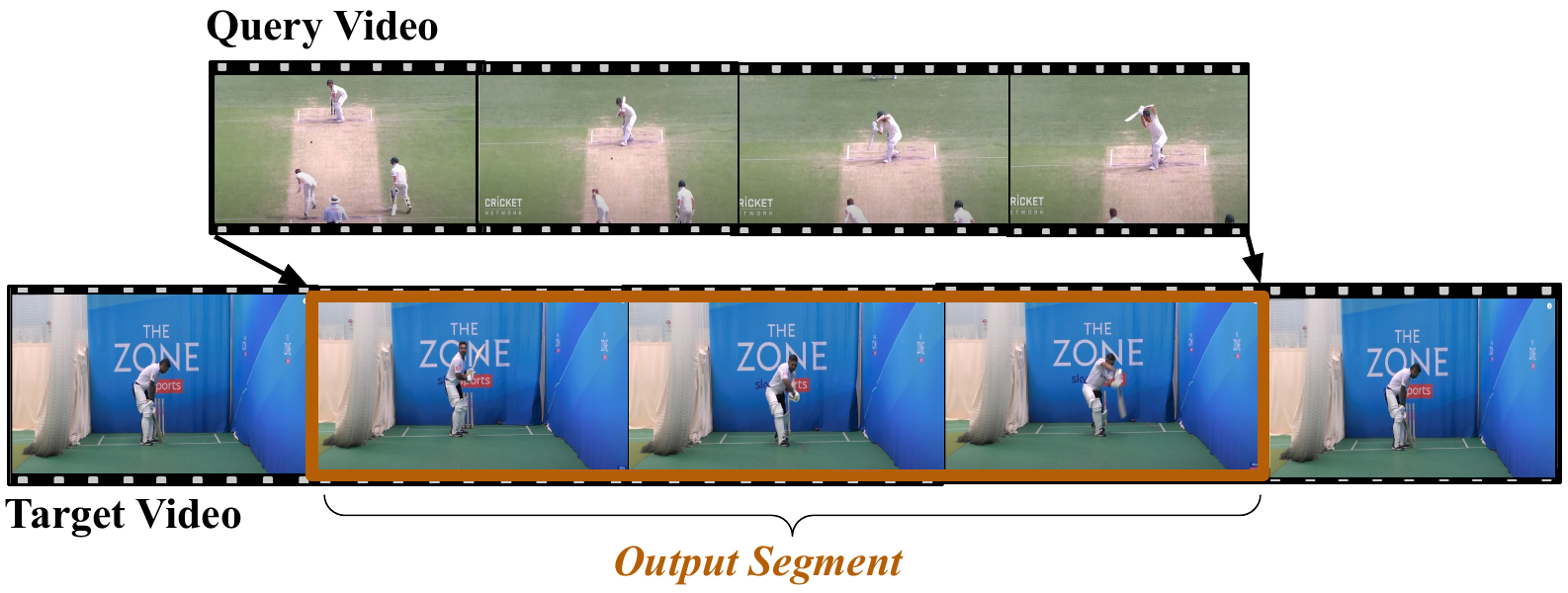}
    \caption[Video-to-video Moment Retrieval Task]{\label{fig:iccv_goal} \textbf{} Given a query video depicting an unseen moment or action, the task is to temporally localize the corresponding moment or action within a target video.}
\end{figure*}

\subsection{Objective-3}

\textbf{\textit{Understanding ``When''}: Develop an approach that, given a text or video query, can ground unseen temporal events in videos.}

\noindent \textbf{Video Moment Grounding.}
Given a query video $q = \{q_1, q_2, \ldots, q_{T_q}\}$ (text or a video) and a target video
$V = \{v_1, v_2, \ldots, v_{T}\}$ consisting of $T$ frames, the goal of video moment grounding is to identify
the temporal segment
\[
\tau = [t_s, t_e] \subseteq ([1, T])
\]
in $V$ that is semantically most similar to $Q$. Formally, the task can be expressed as
learning a function
\[
f_{\text{when}}: (q, V) \rightarrow \tau,
\]
where $f_{\text{when}}$ aligns visual and temporal representations between $Q$ and $V$ to retrieve the corresponding moment as illustrated in Figure~\ref{fig:iccv_goal}.

While the previous objectives addressed \textit{where} objects appear and \textit{how} they interact, this objective focuses on understanding \textit{when} key events occur within a video. Moment-level grounding thus extends video understanding from spatial and relational reasoning to temporal reasoning, pinpointing the exact time intervals corresponding to meaningful actions or events. Traditional approaches typically employ text-based queries for temporal localization; however, such methods often face limitations in descriptive precision, temporal ambiguity, and modality mismatch, which hinder their ability to capture subtle visual cues such as motion dynamics, execution speed, or style variations.

To overcome these limitations, we introduce the Moment Alignment Transformer (MATR) for video-to-video moment retrieval (Vid2VidMR)~\cite{kumar2025matr,kumar2026matrcviu} (Chapter~\ref{chapfour}). In this formulation, a short video clip serves as a visual query to locate semantically similar moments within a longer target video. This visual-to-visual alignment paradigm leverages the full richness of motion and appearance information, offering a more precise and fine-grained retrieval framework than text-based approaches. Furthermore, we extended MATR to handle text queries and achieved performance comparable to state-of-the-art methods.

MATR employs a dual-stage sequence alignment strategy within a transformer architecture to model temporal dependencies between query and target sequences. Specifically, the method integrates Dynamic Time Warping (DTW) at two levels: (i) pre-fusion alignment, which enhances semantic correlation between query and target representations before transformer encoding, and (ii) post-fusion alignment, which refines target video features after fusion for more accurate temporal correspondence. This two-stage alignment enables MATR to capture both coarse-to-fine temporal relationships across varying time scales.

The architecture combines encoder representations with decoder-refined features through learnable query vectors that guide moment localization. Two prediction heads perform (i) foreground–background classification and (ii) boundary regression to predict the temporal boundaries $(t_s, t_e)$ of the retrieved segment. MATR is trained using a multi-objective loss that jointly optimizes alignment cost, classification accuracy, and boundary precision.

To further enhance model initialization and temporal reasoning, we introduce a self-supervised pretraining strategy, where MATR learns to localize randomly sampled clips within the same video. This auxiliary task aligns closely with the main Vid2VidMR objective and encourages robust temporal correspondence learning.

Together, these contributions advance the understanding of \textit{``when''} events occur in videos, enabling temporally grounded, query-driven retrieval of meaningful moments. This complements the spatial (\textit{where}) and relational (\textit{how}) understanding objectives, thereby completing a holistic framework for query-guided video comprehension.

\section{Thesis Organization:}
This thesis is organized as follows.

Chapter \ref{chapbackground} provides a background of Transformer and Few-shot learning relevant to our work. This chapter establishes the foundation and contextualizes our contributions by highlighting key advancements and limitations in the field.

Chapter \ref{chapone} presents a one-shot video object localization framework using Query-Guided DETR. We extend DETR with cross-attention, where queries stem from visual exemplars and key-value pairs encode spatio-temporal video features. The chapter introduces an unsupervised pre-training strategy leveraging synthetic object trajectories, recurrent memory, feature reconstruction, and temporal consistency. Key innovations include query-conditioned object tokens, multi-scale temporal fusion, and differentiable trajectory generation, enabling efficient localization of unseen objects across video sequences via selective attention. The work presented in this chapter has been published in the following publication: \textbf{Yogesh Kumar}, Saswat Mallick, Anand Mishra, Sowmya Rasipuram, Anutosh Maitra, Roshni Ramnani,
    \textit{QDETRv: Query-Guided DETR for One-Shot Object Localization in Videos},
    In Proceedings of the AAAI Conference on Artificial Intelligence (AAAI), pages 2831--2839, 2024~\cite{Kumar2024}.

Chapter \ref{chaptwo} introduces the Temporal Object-Aware Few-Shot Video Object Detection Transformer, which enhances object-level understanding by integrating OWL-ViT a large-scale vision-language model with specialized temporal modeling. We propose a filtering mechanism that selectively propagates high-confidence object features across frames using adaptive thresholding and cross-attention between current frame embeddings and previously detected object representations. The work presented in this chapter has been published in the following publication: \textbf{Yogesh Kumar}, Anand Mishra,
    \textit{Temporal Object-Aware Vision Transformer for Few-Shot Video Object Detection}, In Proceedings of the AAAI Conference on Artificial Intelligence (AAAI), pages 5764--5772, 2026~\cite{kumar2026fsvit}.

In chapter \ref{chapthree} we proposed Few-Shot Referring Relationships that extends object level grounding to relationship level grounding through our T-partite random field formulation with belief propagation-based optimization. The chapter details query-conditioned translational relationship embeddings, Global Semantic Aggregation (GSA) and Local Localization Aggregation (LLA) mechanisms for enhanced relationship representations, and meta-learning frameworks using episodic training. The work presented in this chapter has been published in the following publication: \textbf{Yogesh Kumar}, Anand Mishra,
    \textit{Few-Shot Referring Relationships in Videos},
    In Proceedings of IEEE/CVF Conference on Computer Vision and Pattern Recognition (CVPR), pages 2289--2298, 2023~\cite{Kumar023-2}.

Chapter \ref{chapfour} introduces MATR, a transformer-based model for Video-to-Video Moment Retrieval using dual-stage Dynamic Time Warping (DTW) for sequence alignment. It features learnable queries, multi-scale encoder-decoder fusion, and self-supervised clip localization. Key innovations include differentiable DTW, boundary-aware classification, and temporal augmentations to improve generalization. The work presented in this chapter has been published in the following publications: (i) \textbf{Yogesh Kumar}, Uday Agarwal, Manish Gupta, Anand Mishra,
    \textit{Aligning Moments in Time using Video Queries},
    In Proceedings of International Conference on Computer Vision (ICCV), pages 20215--20225, 2025~\cite{kumar2025matr}, (ii) \textbf{Yogesh Kumar}, Uday Agarwal, Manish Gupta, Anand Mishra,
    \textit{Moment Alignment Transformer for Video-to-Video Moment Retrieval},
    Computer Vision and Image Understanding (CVIU), article 104959, 2026~\cite{kumar2026matrcviu}.

Chapter \ref{chap7conclusionFW} concludes the thesis by summarizing key contributions, acknowledging current limitations, and proposing future research directions in multi-granular video understanding. It highlights opportunities to extend the presented frameworks toward more scalable, generalizable, and interpretable video grounding systems.

\chapter{Background}
\label{chapbackground}

This chapter presents the foundational concepts and prior research that support the methods proposed in this thesis. We highlight key prior works that inform or are directly integrated into our contributions. Readers already familiar with these topics may choose to proceed directly to the next chapter, where our core methodologies are introduced.
\section{Transformers}
\label{sec:transformer}

The Transformer~\cite{vaswani2017attention} represents a paradigm shift in deep learning architectures, replacing recurrent and convolutional operations with self-attention as the primary building block. Its introduction was motivated by the limitations of recurrent neural networks (RNNs)~\cite{medsker2001recurrent} and long short-term memory (LSTM)~\cite{hochreiter1997long} models, which struggle with long-range dependencies and suffer from sequential processing bottlenecks. In contrast, the Transformer processes entire sequences in parallel and models pairwise interactions between tokens via attention, resulting in superior scalability and performance.

Formally, given an input sequence $\mathbf{x}=(x_1,\dots,x_n)$, the Transformer produces contextualized representations $(h_1,\dots,h_n)$ by allowing each token to attend to all others. The ability to capture global dependencies without recurrence or convolution has led to state-of-the-art performance across natural language processing (NLP) tasks such as machine translation, language modeling, and text generation.

Despite its strengths, the Transformer faces key limitations. The computational and memory cost of self-attention is $\mathcal{O}(n^2)$ with respect to sequence length $n$, making it expensive for long inputs (e.g., documents, videos). Furthermore, unlike CNNs, Transformers lack strong inductive biases such as locality and translation equivariance, requiring vast datasets for effective training. Overfitting and instability in low-data regimes remain challenges, although pretraining on massive corpora (e.g., BERT~\cite{bert}, GPT~\cite{achiam2023gpt}) followed by fine-tuning has mitigated these issues in practice.

The flexibility of the Transformer has inspired numerous domain-specific adaptations across text, vision, multimodal, and video understanding.
Table~\ref{tab:transformer_summary} provides a comparative summary of representative Transformer models, highlighting their key innovations, targeted modalities, and efficiency trade-offs.

\subsection{Technical Details}
The Transformer consists of two major components: an \emph{encoder} and a \emph{decoder}, each composed of a stack of identical layers~\cite{vaswani2017attention}. The encoder is responsible for mapping an input sequence to a contextualized representation, while the decoder autoregressively generates output tokens by attending both to previously generated tokens and the encoder output.

At the heart of the architecture lies the \emph{scaled dot-product attention} mechanism, which enables each token to selectively focus on relevant parts of the input sequence. It operates on three components: queries ($Q$), keys ($K$), and values ($V$), computed by projecting each input embedding $x_i$ using learnable weight matrices $W_Q$, $W_K$, and $W_V$. Intuitively, queries represent what each token is looking for, keys encode what each token offers, and values carry the actual information to be aggregated. The attention scores are computed by comparing queries against keys, and these scores are used to weight the values:

\[
\text{Attention}(Q,K,V) = \text{softmax}\!\left(\frac{QK^\top}{\sqrt{d_k}}\right)V,
\]

where $d_k$ is the dimensionality of the keys. This formulation allows the model to dynamically integrate context from the entire sequence, enhancing representation quality.

To improve expressiveness, \emph{multi-head attention (MHA)} extends this mechanism by projecting inputs into $h$ different subspaces, computing attention independently in each, and concatenating the results:
\[
\text{MHA}(Q,K,V) = \text{Concat}(\text{head}_1,\dots,\text{head}_h)W^O, \quad
\text{head}_i = \text{Attention}(QW_Q^i,KW_K^i,VW_V^i).
\]
This allows the model to capture diverse types of relationships simultaneously (e.g., syntactic and semantic dependencies in language, or local and global structures in vision).

Each encoder block contains:
\begin{enumerate}
    \item Multi-head self-attention, where every token attends to all others in the input.
    \item A position-wise feed-forward network (FFN), consisting of two linear layers with nonlinearity (ReLU or GELU).
    \item Residual connections and layer normalization applied around each sub-layer for stable optimization.
\end{enumerate}
Formally, for input $H^{(l)}$ at layer $l$:
\[
\tilde{H}^{(l)} = \text{LN}(H^{(l)} + \text{MHA}(H^{(l)})), \quad
H^{(l+1)} = \text{LN}(\tilde{H}^{(l)} + \text{FFN}(\tilde{H}^{(l)})).
\]

The decoder mirrors this structure, but includes two additional mechanisms:
\begin{itemize}
    \item \emph{Masked self-attention}, which prevents access to future tokens and ensures autoregressive generation.
    \item \emph{Cross-attention}, where queries come from the decoder and keys/values are taken from the encoder output, enabling the decoder to condition its predictions on the source sequence.
\end{itemize}

Since self-attention is permutation-invariant, positional encodings are added to input embeddings to encode sequence order. Vaswani et al.~\cite{vaswani2017attention} introduced sinusoidal encodings:
\[
PE_{(pos,2i)} = \sin\!\left(\frac{pos}{10000^{2i/d}}\right), \quad
PE_{(pos,2i+1)} = \cos\!\left(\frac{pos}{10000^{2i/d}}\right),
\]
where $pos$ is the position index and $d$ is embedding dimension. Later variants include learned embeddings and rotary position encodings.

By stacking $L$ layers of encoders and decoders, the Transformer scales to highly expressive models. A task-specific output head (e.g., linear projection with softmax for classification, autoregressive token generation for language modeling, bounding box regression for detection) is applied on top of the final decoder or encoder outputs.

A key property of the Transformer is its modality-agnostic nature; as long as data can be represented as a sequence of embeddings, the architecture can process it. This includes words in a sentence, image patches, audio frames, or spatio-temporal video tokens. Thus, the Transformer has evolved into a general-purpose architecture for sequence-to-sequence modeling across diverse domains.

Despite these strengths, the quadratic cost $\mathcal{O}(n^2)$ in sequence length $n$ remains a bottleneck for long sequences. This has motivated research into sparse, linear, or hierarchical attention mechanisms, such as Longformer~\cite{beltagy2020longformer}, Performer~\cite{choromanski2020rethinking}, and hierarchical variants designed for vision and video tasks.

\subsection{Extensions to Vision and Multimodality}
The extension of Transformers to vision was pioneered by Vision Transformer (ViT)~\cite{dosovitskiy2020transformers}, which views an image as a sequence of patches rather than pixels. Specifically, an image $I \in \mathbb{R}^{H\times W \times C}$ is divided into $N$ non-overlapping patches of size $P\times P$, each flattened and linearly projected into an embedding:
\[
z_i = E(x_i) + p_i, \quad i=1,\dots,N,
\]
where $E(\cdot)$ is a learnable projection matrix, $x_i \in \mathbb{R}^{P^2C}$ is the flattened patch, and $p_i$ is a positional embedding that encodes spatial order. A learnable \texttt{[CLS]} token $z_{\text{cls}}$ is prepended to the sequence, and its representation after the Transformer encoder is used for classification. Thus, the Transformer input is:
\[
Z = [z_{\text{cls}}; x_1E; x_2E; \dots; x_NE] + P,
\]
where $P$ denotes positional encodings. ViT demonstrates that Transformers can match or surpass CNNs when pretrained on sufficiently large datasets (e.g., JFT-300M~\cite{jft300}).

To adapt Transformers more effectively for vision, several innovations were introduced. DeiT~\cite{touvron2022deit} showed that careful regularization and knowledge distillation from CNN teachers can enable ViT training on smaller datasets such as ImageNet-1k. Swin Transformer~\cite{liu2021swin} introduced a hierarchical design by computing attention within shifted local windows and merging patches across stages, which reintroduces locality and reduces complexity. CvT~\cite{wu2021cvt} integrates convolutional projections into patch embedding and attention layers to recover inductive bias. Positional encoding strategies were also adapted for 2D spatial data, ranging from sine-cosine embeddings on a 2D grid to learnable embeddings and relative positional biases. These design choices illustrate how structural priors (e.g., locality, hierarchy) are injected back into the otherwise permutation-invariant Transformer.

\paragraph{Modality Adaptation.}
Transformers require each modality to be expressed as a sequence of tokens. Different modalities are adapted as follows:

\textbf{Vision.}
\begin{itemize}
    \item \textit{Patch embedding:} Instead of words, images are split into patches $x_i \in \mathbb{R}^{P^2C}$ and projected into $d$-dimensional embeddings, analogous to NLP token embeddings.
    \item \textit{Class token:} A learnable \texttt{[CLS]} token is prepended for classification tasks.
    \item \textit{Hierarchical variants:} Swin, CvT, and others employ patch merging, local windows, and convolutional tokenization to reintroduce inductive bias.
    \item \textit{Positional encoding:} Essential to preserve 2D spatial order, typically via sine-cosine 2D embeddings, learnable embeddings, or relative position bias.
\end{itemize}

\textbf{Multimodality.}
\begin{itemize}
    \item \textit{Text:} Standard subword tokenization (e.g., BPE) with embeddings plus positional encodings.
    \item \textit{Image (CLIP):} A ViT-style vision encoder produces patch embeddings, while a Transformer text encoder handles subword tokens. Both are aligned with a contrastive loss:
    \[
    \mathcal{L}_{\text{CLIP}} = -\frac{1}{N}\sum_{i=1}^N
    \Bigg(
    \log \frac{\exp(\langle v_i, t_i \rangle / \tau)}{\sum_j \exp(\langle v_i, t_j \rangle / \tau)}
    +
    \log \frac{\exp(\langle t_i, v_i \rangle / \tau)}{\sum_j \exp(\langle t_i, v_j \rangle / \tau)}
    \Bigg),
    \]
    where $v_i$ and $t_i$ are vision and text embeddings, and $\tau$ is a temperature parameter.
    \item \textit{Cross-attention (ViLBERT, LXMERT):} Separate unimodal encoders produce features for image regions and words. Cross-attention layers then exchange information, e.g.,
    \[
    \text{Attn}(Q_{\text{text}}, K_{\text{vision}}, V_{\text{vision}}).
    \]
    \item \textit{Unified tokenization (PaLI, Flamingo):} All modalities are projected into a shared token space, allowing a single Transformer to attend jointly across them. For video/audio, temporal patches or spectrogram frames are tokenized in the same way as image patches.
\end{itemize}

Recent large-scale multimodal foundation models (e.g., Flamingo~\cite{alayrac2022flamingo}, PaLI~\cite{chen2022pali}) demonstrate that with unified tokenization and sufficient scale, Transformers can serve as a backbone for diverse multimodal reasoning tasks.

Adapting Transformers to vision and multimodality involves three main components: (i) modality-specific tokenization (e.g., image patches, spectrogram frames, video cubes), (ii) positional encoding schemes to capture modality structure (1D for text, 2D for images, temporal encodings for video), and (iii) interaction mechanisms such as self-attention, cross-attention, or contrastive alignment. This flexibility makes the Transformer a unifying architecture across diverse modalities.

\subsection{Transformers for Object Detection}
Transformers have been adapted to object detection through DETR (DEtection TRansformer)~\cite{detr}, which formulates detection as a direct set prediction problem. Unlike anchor-based detectors (e.g., Faster R-CNN~\cite{faster2015towards}, RetinaNet~\cite{wang2019automatic}), DETR uses a fixed set of learnable object queries, which the Transformer decoder maps to bounding box coordinates and class labels. A CNN backbone extracts image features, which are flattened and fed into the Transformer encoder. The decoder cross-attends between object queries and encoder features, refining predictions iteratively.

The loss function combines a bipartite matching cost and a set-based prediction loss:

\[
\mathcal{L} = \sum_{(y,\hat{y})\in \pi^*} \mathcal{L}_{cls}(y,\hat{y}) + \lambda \mathcal{L}_{box}(y,\hat{y}),
\]

where $\pi^*$ denotes the optimal one-to-one assignment between predicted outputs $\hat{y}$ and ground-truth targets $y$. This assignment is computed using the Hungarian algorithm, which minimizes a matching cost. Unlike traditional detectors that rely on heuristic post-processing (e.g., non-maximum suppression), this formulation treats object detection as a direct set prediction problem, ensuring permutation-invariant supervision and eliminating duplicate predictions.

DETR eliminates the need for non-maximum suppression (NMS) and anchor design, but requires longer training to converge. Deformable DETR~\cite{ddetr} improves efficiency by restricting attention to sparse spatial locations using learned offsets. Variants such as DAB-DETR~\cite{liu2022dab} and DN-DETR~\cite{dn-detr} further accelerate convergence and improve localization.

\setlength{\tabcolsep}{5pt}
\begin{table}[t!]
\renewcommand{\arraystretch}{1.25}
\scriptsize
\centering
\begin{tabular}{>{\arraybackslash}m{3cm}
                >{\arraybackslash}m{4cm}
                m{6.5cm}}
\toprule
\textbf{Model} & \textbf{Modality / Task} & \textbf{Key Contributions and Innovations} \\
\midrule
Transformer~\cite{vaswani2017attention} & Text (Seq2Seq) & Introduced self-attention and encoder-decoder architecture for sequence modeling. \\
BERT~\cite{bert}, GPT~\cite{achiam2023gpt} & Text (Pretraining) & Masked LM (BERT) and autoregressive LM (GPT); fine-tuning for downstream tasks. \\
ViT~\cite{dosovitskiy2020transformers} & Image (Classification) & Treats image as patch sequence; shows Transformers rival CNNs with large-scale pretraining. \\
DeiT~\cite{touvron2022deit} & Image (Classification) & Data-efficient ViT training via distillation and augmentation. \\
Swin Transformer~\cite{liu2021swin} & Image (Dense Tasks) & Hierarchical features; shifted window attention for scalable vision tasks. \\
DETR~\cite{detr} & Image (Detection) & Set prediction via object queries; removes anchors and NMS. \\
Deformable DETR~\cite{ddetr} & Image (Detection) & Sparse deformable attention over sampled keypoints; faster training. \\
CLIP~\cite{clip} & Image-Text (Retrieval) & Contrastive dual-encoder trained on web-scale image-text pairs. \\
ViLBERT~\cite{lu2019vilbert} & Image-Text (VQA) & Cross-attention fusion for multimodal reasoning. \\
TimeSformer~\cite{timesformer} & Video (Classification) & Factorized spatial-temporal attention for efficient video modeling. \\
ViViT~\cite{arnab2021vivit} & Video (Classification) & Explores multiple spatio-temporal factorization strategies. \\
Video Swin~\cite{vidswimtrans} & Video (Detection/Action) & Extends Swin to 3D; hierarchical spatio-temporal features. \\
Flamingo~\cite{alayrac2022flamingo}, PaLI~\cite{chen2022pali} & Image-Text-Video (Multimodal) & Unified multimodal models with few-shot capabilities and large-scale pretraining. \\
Moment-DETR~\cite{moment_detr} & Text-Video (Moment Retrieval) & Moment prediction as direct-set problem. \\
VRDFormer~\cite{zheng2022vrdformer} & Video (Relation Detection) & End-to-end query-based framework with static and recurrent queries for spatio-temporal relation detection and tagging. \\
Tube-DETR~\cite{yang2022tubedetr} & Spatio-Temporal Video Grounding & Space-time transformer that jointly models temporal, spatial, and multi-modal interactions for spatio-temporal video grounding. \\
EGTR~\cite{im2024egtr} & Image (Scene Graph Generation) & Extracts relation graph directly from DETR’s self-attention; introduces adaptive smoothing and connectivity prediction for efficient one-stage SGG. \\

STTran~\cite{cong2021spatial} & Video (Dynamic Scene Graph Generation) & Spatial encoder for intra-frame context and temporal decoder for inter-frame dependencies; introduces multi-label margin loss and semi-constrained graph generation. \\
OWL-ViT~\cite{owl} & Text and Image (Open-Vocabulary Detection) & Simplifies DETR by removing decoder; enables open-vocabulary and one-shot detection via contrastively pre-trained image-text encoders. \\
Video OWL-ViT~\cite{video-owlvit23} & Text and Video (Open-World Localization and Tracking) & Extends OWL-ViT to video by adding a temporal transformer decoder that propagates object queries across frames for end-to-end open-world tracking with improved temporal consistency. \\

\bottomrule
\end{tabular}
\caption[Representative Transformer architectures]{Representative Transformer architectures across modalities and tasks. We summarize their core innovations and modality scope.}
\label{tab:transformer_summary}
\end{table}

\subsection{Transformers for Open-World and Open-Vocabulary Detection}
Traditional object detectors are typically trained under a closed-world assumption, where the set of object categories is fixed and fully annotated. However, real-world environments are open-ended: models must detect known categories, localize novel or unseen classes, and continually adapt to new semantic spaces. This motivates two closely related paradigms: \textit{open-world object detection} (OWOD) and \textit{open-vocabulary object detection} (OVOD).

\paragraph{Problem Setup.}
In OWOD, let $\mathcal{C}_{\text{train}}$ denote the set of known classes during training, and $\mathcal{C}_{\text{test}} = \mathcal{C}_{\text{train}} \cup \mathcal{C}_{\text{novel}}$ represent test-time classes, where $\mathcal{C}_{\text{novel}}$ are unseen. The model must: (i) detect and classify $\mathcal{C}_{\text{train}}$, (ii) localize instances of $\mathcal{C}_{\text{novel}}$ as ``unknown,'' and (iii) incrementally integrate $\mathcal{C}_{\text{novel}}$ with minimal supervision while avoiding catastrophic forgetting.

In OVOD, the objective is to enable detectors to recognize categories beyond $\mathcal{C}_{\text{train}}$ by leveraging external semantic knowledge from language models or image-text pretraining, without explicit bounding-box annotations for novel classes.

\paragraph{Transformer-Based Approaches in OWOD.}
Transformers naturally support OWOD due to their query-centric design and compatibility with multimodal supervision:
\begin{itemize}
    \item \textbf{DETR-style extensions:} DETR~\cite{detr} reformulates detection as set prediction with object queries. OW-DETR~\cite{gupta2022ow} extends this idea with attention-driven pseudo-labeling, a novelty classification head, and an objectness branch to separate unknowns from background. Fast-OWDETR~\cite{chen2022fast} reduces overhead via lightweight pseudo-label selection and improved bounding-box refinement.
    \item \textbf{Incremental learning:} Open World DETR~\cite{dong2022open} introduces exemplar replay buffers, multi-view consistency, and knowledge distillation to expand classification heads while mitigating catastrophic forgetting. PROB~\cite{zohar2023prob} models objectness using Gaussian distributions in embedding space, explicitly separating objectness from classification and facilitating smoother class expansion.
    \item \textbf{Language-supervised detection:} MAVL~\cite{maaz2022class} and related models replace fixed classifier weights with CLIP-style text embeddings. This enables zero-shot transfer by aligning region features with category names or natural language descriptions, bridging OWOD and OVOD.
\item \textbf{Prompt-based adaptation:} Lightweight prompt-tuning strategies allow frozen vision-language backbones (e.g., CLIP~\cite{clip}, PaLI~\cite{chen2022pali}, Flamingo~\cite{alayrac2022flamingo}) to adapt their detection heads to new categories with minimal fine-tuning, improving scalability.
\end{itemize}

\paragraph{Transformer-Based Approaches in OVOD.}
In parallel, OVOD methods focus on leveraging vision-language pretraining and semantic supervision to enable vocabulary expansion without explicit bounding-box annotations:
\begin{itemize}
    \item \textbf{Knowledge distillation:} ViLD~\cite{gu2021open} distills knowledge from CLIP into detection heads via dual-branch learning. HierKD~\cite{ma2022open} and LP-OVOD~\cite{pham2024lp} extend this with loss refinements and pseudo-labeling schemes to improve transferability.
    \item \textbf{Region-text pretraining:} OVR-CNN~\cite{zareian2021open} aligns region features with text embeddings, while GLIP~\cite{li2022grounded} scales this to billions of image-text pairs with grounding supervision. RO-ViT~\cite{kim2023region} augments positional embeddings and replaces cross-entropy with focal loss for better region-level discrimination.
    \item \textbf{Prompt modeling:} PromptDet~\cite{feng2022promptdet} and DetPro~\cite{du2022learning} design structured or continuous prompts to inject semantic priors. Context grading and background-aware prompts further refine alignment with novel categories.
\end{itemize}

\paragraph{OWOD vs. OVOD.}
While both paradigms extend closed-set detection, their emphases differ. OVOD leverages language and large-scale pretraining to broaden the detectable vocabulary in a zero- or few-shot fashion. OWOD additionally requires \emph{incremental adaptation}: localizing unknowns, integrating them as novel known classes, and preserving prior knowledge. Thus, OVOD focuses on semantic expansion via language grounding, whereas OWOD addresses the dual challenges of \emph{unknown discovery} and \emph{continual learning}.

\subsection{Transformers for Video Understanding}
Extending Transformers to video understanding requires modeling spatio-temporal dependencies efficiently. A straightforward approach is to tokenize video frames into spatio-temporal patches, but the resulting sequence length grows quadratically with both spatial and temporal resolution.

TimeSformer~\cite{timesformer} addresses this by factorizing attention into temporal and spatial dimensions, reducing complexity from $\mathcal{O}((HW T)^2)$ to $\mathcal{O}(T^2 + (HW)^2)$. ViViT~\cite{arnab2021vivit} explores several variants, including factorized encoder design and spatio-temporal pooling. Video Swin Transformer~\cite{vidswimtrans} adapts shifted window attention to 3D, producing hierarchical spatio-temporal representations.

For action detection and temporal localization, ActionFormer~\cite{zhang2022actionformer} uses a Transformer encoder to aggregate long temporal context efficiently. VideoBERT~\cite{sun2019videobert} adapts BERT-style masked prediction to video-text data, leveraging large-scale pretraining (e.g., HowTo100M~\cite{howto100m}).  TubeDETR~\cite{yang2022tubedetr} extends DETR-style architectures to spatio-temporal video grounding by introducing a dual-stream video-text encoder and a space-time transformer decoder. It avoids handcrafted proposals by using time-aligned queries and achieves efficient end-to-end grounding with reduced memory footprint.

Key challenges for video Transformers include: (1) handling long temporal data, (2) high memory/computation cost, (3) scarcity of labeled video data, and (4) robustness to variable frame rates and motion. Common implementation strategies involve using CLIP~\cite{clip}, 2D CNNs (e.g., ResNet~\cite{resnet}, Swin~\cite{liu2021swin}) for spatial encoding, followed by a temporal Transformer for sequence modeling. End-to-end video Transformers are increasingly feasible with advances in hardware and efficient attention mechanisms (e.g., sparse, linear, or axial attention).

Overall, Transformers have become central to video understanding, powering state-of-the-art results in classification, detection, retrieval, and captioning tasks, though scalability and efficiency remain open research directions.

\section{Few-Shot Learning}
\label{sec:fsl}
Few-shot learning (FSL) addresses the fundamental challenge of enabling models to generalize to novel categories or tasks when only a handful of labeled examples are available~\cite{wang2020generalizing}. Formally, let $\mathcal{C}_{\text{train}}$ and $\mathcal{C}_{\text{test}}$ denote disjoint sets of base and novel classes, respectively. A model is trained on a large-scale dataset $\mathcal{D}_{\text{train}}=\{(x_i,y_i)\}_{i=1}^N$, with $y_i \in \mathcal{C}_{\text{train}}$, and evaluated on $\mathcal{D}_{\text{test}}$, where each class $c \in \mathcal{C}_{\text{test}}$ is represented by only $K$ labeled support samples $\mathcal{S}_c=\{(x_j,y_j)\}_{j=1}^K$. The learner must predict labels for a query set $\mathcal{Q}$ drawn from $\mathcal{C}_{\text{test}}$. The typical setting is $K \ll |\mathcal{C}_{\text{train}}|$, with $K=1$ (one-shot) or $K=5$ being the most common.

While the FSL framework naturally extends beyond classification to tasks such as object detection, segmentation, reinforcement learning, or cross-modal reasoning (e.g., vision–language), in this section we primarily focus on the \emph{classification setting}. This restriction allows for a clearer comparison of methods under a standardized benchmark, while acknowledging that many techniques have broader applicability.

In practice, FSL is often framed as an $N$-way $K$-shot classification problem: given $N$ novel classes and $K$ support samples per class, the task is to predict the label of queries from those $N$ classes. Training usually follows an \emph{episodic paradigm}~\cite{vinyals2016matching}, where each training episode simulates a few-shot task by sampling $N$ classes and $K$ supports per class from $\mathcal{C}_{\text{train}}$, encouraging the model to generalize to $\mathcal{C}_{\text{test}}$.

The motivation for FSL arises from real-world scenarios where collecting large annotated datasets is infeasible, expensive, or even impossible. For example, rare diseases in medical imaging may have only a handful of expert-annotated scans, while autonomous driving must handle corner cases (e.g., animals crossing or unusual weather conditions) with extremely limited data. In natural language processing, low-resource languages often lack parallel corpora. Unlike conventional supervised learning, where abundant data supports statistical generalization, FSL emphasizes \emph{inductive bias}, \emph{transfer of prior knowledge}, and \emph{meta-level learning} to overcome the data scarcity bottleneck.

\subsection{Approaches}
FSL research has produced multiple families of methods, each motivated by different perspectives on how humans generalize from few examples. These are summarized in Table~\ref{tab:back_few-shot}, and expanded below.

\textbf{(i) Metric-based approaches.}
These methods learn an embedding function $f_\theta: \mathcal{X} \rightarrow \mathbb{R}^d$ that maps inputs into a metric space, where $\mathcal{X}$ is the input space and $d$ is the embedding dimension. Classification is then performed by comparing query embeddings with class prototypes.
Formally, given a query $x_q$ and a support set $\mathcal{S}_c = \{(x_j,y_j)\}_{j=1}^K$ for class $c$, the prototype $\mu_c$ is the mean of the support embeddings:
\[
\mu_c = \frac{1}{K} \sum_{(x_j,y_j) \in \mathcal{S}_c} f_\theta(x_j).
\]
The query is assigned to the closest prototype according to a distance metric $d(\cdot,\cdot)$:
\[
c^\ast = \arg\min_{c \in \mathcal{C}_{\text{test}}} d\!\left(f_\theta(x_q), \mu_c\right).
\]
Siamese networks~\cite{koch2015siamese} pioneered pairwise similarity learning, while Matching Networks~\cite{vinyals2016matching} used attention over support examples. Relation Networks~\cite{sung2018learning} introduced non-linear metric learning. More recent methods like Meta-Baseline~\cite{chen2021meta} integrate metric learning with pretrained backbones. These methods are computationally efficient and interpretable but degrade under domain shift when support and query distributions differ.

\begin{table}[t!]
\scriptsize
\centering
\label{tab:fsl_approaches}
\renewcommand{\arraystretch}{1.25}
\begin{tabular}{p{2.6cm} p{4.3cm} p{3.6cm} p{3.7cm}}
\toprule
\textbf{Approach} & \textbf{Core Principle} & \textbf{Representative Methods} & \textbf{Strengths / Limitations} \\
\midrule
\textbf{Metric-based} & Learn a shared embedding space where classification is performed via similarity to class prototypes or exemplars. Often trained episodically. &
Prototypical Nets~\cite{snell2017prototypical}, Matching Nets~\cite{vinyals2016matching}, Siamese Nets~\cite{koch2015siamese}, RelationNet~\cite{sung2018learning} &
+ Simple, fast inference; non-parametric adaptation. \newline -- Sensitive to embedding quality; vulnerable to domain shift. \\
\midrule
\textbf{Optimization-based} & Meta-learn an initialization or update rule that enables rapid adaptation to new tasks with few gradient steps. &
MAML~\cite{finn2017model}, Reptile~\cite{nichol2018reptile}, Meta-SGD~\cite{metasgd}, ANIL~\cite{anil}, ProtoMAML~\cite{triantafillou2019meta} &
+ Strong task-level generalization. \newline -- Computationally heavy; unstable in high dimensions. \\
\midrule
\textbf{Data Augmentation / Generative} & Enrich support data via synthetic samples or feature-level perturbations using generative models or semantic mixing. &
FeatGen~\cite{featgen}, GAN-based~\cite{ganfew}, VAE-based~\cite{schonfeld2019generalized}, Diffusion-based~\cite{difffewshot}, CLIP-based ~\cite{tang2024amu} &
+ Mitigates scarcity; improves diversity. \newline -- Risk of unrealistic or biased samples. \\
\midrule
\textbf{Self-Supervised / Contrastive} & Learn transferable representations via contrastive or self-supervised pretraining, then adapt to few-shot tasks. &
MoCo~\cite{he2020momentum}, SimCLR~\cite{chen2020simple}, BYOL~\cite{grill2020bootstrap}, DINO~\cite{caron2021emerging}, SwAV~\cite{caron2020swav} &
+ Strong embeddings; reduces label reliance. \newline -- Pretraining requires massive unlabeled data. \\
\midrule
\textbf{Transformer-based / Prompting} & Use attention-based architectures or prompts to model task structure and leverage foundation models. &
Few-shot Transformers~\cite{dong2022self}, Prompt-tuning~\cite{Zhou_2022}, GPT-4V-assisted FSL~\cite{openai2023gpt4}, Flamingo~\cite{alayrac2022flamingo} &
+ Scales with LLMs; interpretable task conditioning. \newline -- Prompt sensitivity; high compute demand. \\
\bottomrule
\end{tabular}
\caption[Few-shot Learning Approaches]{Major approaches in few-shot learning (FSL), with principles, representative methods, and trade-offs. \label{tab:back_few-shot}}
\end{table}
\textbf{(ii) Optimization-based approaches.}
In this family, few-shot learning is framed as \emph{meta-learning} across a distribution of tasks $\mathcal{T} \sim p(\mathcal{T})$. Each task $\mathcal{T}$ is split into a \emph{support set} $\mathcal{S}$ (used for adaptation) and a \emph{query set} $\mathcal{Q}$ (used for evaluation). The key idea is to learn model parameters $\theta$ that can be quickly adapted to a task-specific parameter set $\theta'$ using only $\mathcal{S}$, while still achieving good generalization on $\mathcal{Q}$.

For example, Model-Agnostic Meta-Learning (MAML)~\cite{finn2017model} learns an initialization $\theta$ that adapts to each task with a few gradient steps:
\[
\theta' = \theta - \alpha \nabla_\theta \mathcal{L}_{\mathcal{S}}(f_\theta),
\]
where $\alpha$ is the adaptation step size and $\mathcal{L}_{\mathcal{S}}$ is the task-specific loss computed on the support set $\mathcal{S}$. The \emph{meta-objective} is then to minimize the query loss after adaptation:
\[
\min_\theta \; \mathbb{E}_{\mathcal{T} \sim p(\mathcal{T})} \big[ \mathcal{L}_{\mathcal{Q}}(f_{\theta'}) \big],
\]
where $\mathcal{L}_{\mathcal{Q}}$ denotes the loss evaluated on the query set $\mathcal{Q}$. This formulation explicitly enforces that the learned initialization $\theta$ is such that, after fine-tuning on $\mathcal{S}$, the adapted model $f_{\theta'}$ performs well on unseen samples $\mathcal{Q}$ from the same task.

Variants include Reptile~\cite{nichol2018reptile}, which simplifies the meta-update by averaging task-specific updates; ANIL~\cite{anil}, which adapts only the final classification head; and ProtoMAML~\cite{triantafillou2019meta}, which integrates prototypes into the MAML framework. Latent Embedding Optimization (LEO)~\cite{rusu2018meta} further improves scalability by learning low-dimensional latent codes for each task. While optimization-based approaches provide strong task-level generalization, they can be computationally expensive and unstable in high-dimensional parameter spaces, requiring careful design and regularization.

\textbf{(iii) Data augmentation and generative approaches.}
Generative methods aim to enrich the support set by synthesizing new examples. Early work used GANs~\cite{wang2018low} and VAEs~\cite{pu2016variational} to produce class-consistent samples. More recently, diffusion models~\cite{difffewshot} have been employed for controlled generation.
At the feature level, interpolation-based augmentation (e.g., Mixup) is widely used. Given two embeddings $f_\theta(x_a)$ and $f_\theta(x_b)$, a mixed feature $\tilde{f}$ is created as
\[
\tilde{f} = \lambda f_\theta(x_a) + (1-\lambda) f_\theta(x_b), \quad \lambda \sim \text{Beta}(\alpha,\alpha),
\]
where $\alpha$ is a hyperparameter controlling the interpolation.
Recent advances leverage text-guided models such as CLIP, enabling semantically controlled hallucination~\cite{tang2024amu}. While these methods increase diversity, they risk introducing unrealistic or biased samples.

\textbf{(iv) Self-supervised and contrastive approaches.}
Self-supervised pretraining aims to learn transferable visual representations without requiring annotated labels, which can then be adapted to few-shot learning tasks. A dominant paradigm is \emph{contrastive learning}, where augmented views of the same instance are treated as positives, and embeddings from other instances serve as negatives. The objective encourages invariance to data augmentations while enforcing instance discrimination. A widely used formulation is the InfoNCE loss:
\[
\mathcal{L}_{\text{NCE}} = -\log \frac{\exp(f(x)\cdot f(x^+)/\tau)}{\sum_{x^-}\exp(f(x)\cdot f(x^-)/\tau)},
\]
where $f(\cdot)$ is the encoder, $x^+$ is a positive (augmented) sample of $x$, $\{x^-\}$ are negatives drawn from the batch or memory bank, and $\tau$ is a temperature hyperparameter. This contrastive setup naturally aligns with few-shot settings by learning discriminative embeddings that generalize well across tasks.

\emph{Instance-level contrastive learning.}
SimCLR~\cite{chen2020simple} relies on large batch sizes to provide sufficient negatives and employs strong augmentations (cropping, color jitter, blur) along with an MLP projection head. MoCo~\cite{he2020momentum} alleviates the need for large batches by maintaining a dynamic memory queue of negatives and using a momentum encoder for stable representations. These methods yield highly discriminative embeddings suitable for metric-based FSL methods.

\emph{Bootstrap and redundancy-reduction approaches.}
BYOL~\cite{grill2020bootstrap} removes explicit negatives by using an online encoder and a momentum target encoder, where the online network learns to predict the target's representations. Despite lacking negatives, BYOL avoids representational collapse by architectural asymmetry. Similarly, SimSiam~\cite{chen2021exploring} shows that predictor asymmetry alone suffices. SwAV~\cite{caron2020swav} departs from pairwise contrast by performing online clustering with a "swapped prediction" loss, aligning assignments of different augmentations, thereby learning more semantically structured embeddings.

\emph{Transformer-based SSL.}
Vision Transformer (ViT)-based self-supervised learning has emerged as particularly powerful for transferring to FSL tasks.
DINO~\cite{caron2021emerging} employs a self-distillation framework, where a \emph{student} network is trained to match the representations of a \emph{teacher} network under different augmentations.
The teacher parameters are updated as an \emph{exponential moving average (EMA)} of the student parameters, ensuring stable training and preventing collapse.
This design enables alignment of both patch-level and global representations, capturing fine-grained semantic correspondences.

Building on this idea, EsViT (Efficient Self-supervised Vision Transformers)~\cite{li2022esvit} introduces token-level contrastive learning, where individual patch tokens are contrasted across different views in addition to global features.
By encouraging local token consistency, EsViT enhances the contextual modeling capability of ViTs, which is particularly valuable for open-set and cross-domain few-shot learning.
Overall, transformer-based SSL methods leverage contextual, patch-level interactions to produce richer representations, narrowing the gap between pretraining and the demands of real-world FSL.

\textbf{(v) Transformer-based and prompt-oriented approaches.}
Transformers naturally model support–query interactions. Few-shot Transformers~\cite{dong2022self} use cross-attention between support and query embeddings for direct adaptation.
Prompt-based methods, such as CoOp~\cite{Zhou_2022}, learn continuous text prompts that adapt pretrained vision–language models (e.g., CLIP) for FSL. Large multimodal models like Flamingo~\cite{alayrac2022flamingo} and GPT-4V~\cite{openai2023gpt4} demonstrate few-shot reasoning via in-context learning, unifying few-shot and zero-shot paradigms. However, these methods remain sensitive to prompt design and require substantial computational resources.

Overall, the concepts reviewed in this chapter establish the technical foundations for our thesis. We discussed how Transformers evolved into a unified architecture for text, vision, and multimodal reasoning; how their query-based formulations enable object detection, open-world recognition, and video understanding; and how few-shot learning provides mechanisms for generalizing under limited supervision. These ideas directly underpin the design choices in our subsequent chapters, where we leverage Transformer-based models and low-supervision learning to develop effective methods for query-guided video understanding.

\newcommand{\model}{QDETRv}

\chapter{One-shot Object Grounding in Videos}
\label{chapone}

In this chapter, we study one-shot video object localization problem that aims to localize instances of unseen objects in the target video using a single query image of the object. Toward addressing this challenging problem, we extend a popular and successful object detection method, namely DETR (Detection Transformer), and introduce a novel approach -- query-guided detection transformer for videos (\model{}). A distinctive feature of \model{} is its capacity to exploit information from the query image and spatio-temporal context of the target video, which significantly aids in precisely pinpointing the desired object in the video. We incorporate cross-attention mechanisms that capture temporal relationships across adjacent frames to handle the dynamic context in videos effectively. Further, to ensure strong initialization for \model{}, we also introduce a novel unsupervised pretraining technique tailored to videos. This involves training our model on synthetic object trajectories with an analogous objective as the query-guided localization task. During this pretraining phase, we incorporate recurrent object queries and loss functions that encourage accurate patch feature reconstruction. These additions enable better temporal understanding and robust representation learning.
Our experiments show that the proposed model significantly outperforms the competitive baselines on two public benchmarks, VidOR and ImageNet-VidVRD, extended for one-shot open-set localization tasks.

The subsequent sections present the introduction, a review of related work, the proposed methodology, and a detailed analysis of results and discussions.
\section{Introduction}
\begin{figure}[t!]
    \centering
      \includegraphics[width=0.75\textwidth]{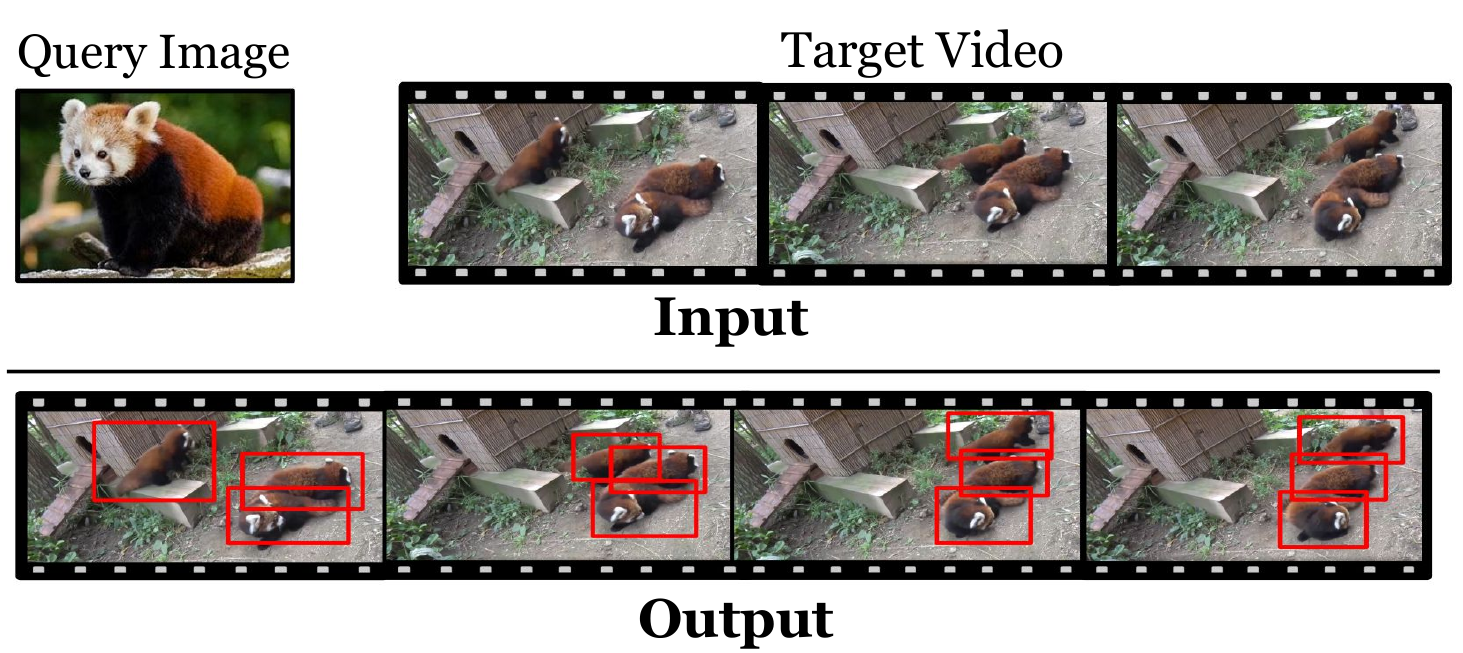}
    \caption[Illustration of the proposed task]{\label{fig:aaai24goal} Given an unseen query image and a target video, the model is expected to localize all instances of objects present in the query image on the target video. [Best viewed in color]. }
\end{figure}

The field of computer vision has long been engaged in the pursuit of localizing objects of interest within videos. In the past, the primary emphasis has been localizing the specified object within a single frame on the entire video~\cite{sivic2003video}. Additionally, efforts have been made towards recognizing individuals based on their facial features~\cite{sivic2005person} or clothing characteristics~\cite{brunelli1995person}. In a more novel direction, some research has explored using natural language or sketch-based queries to precisely locate objects within images~\cite{owl,sadhu2019zero,Kumar023-2,tripathisketch,tripathi2023multimodal}. As visual comprehension continues to advance, there is a need for localization methodologies that can effectively pinpoint unfamiliar objects in the video that were not encountered during the training phase and unexpectedly appeared during testing. This is particularly relevant in various application areas, including surveillance systems and industrial automation. We illustrate our open-set one-shot object localization goal in Figure~\ref{fig:aaai24goal}.

To address the challenge posed by open-set one-shot object localization, our goal is to develop a model that, once trained, can localize any object query that might arise unexpectedly during testing. As noted in some of the recent works~\cite{hsieh2019one,osokin2020os2d}, such a model should maintain its class-agnostic nature even during the training process. Our requisites, distinct from approaches, necessitate prior access to the list of category names during both training and testing~\cite{fan2022few,du2022unknown,liang2023unknown}. To this end, we extend a popular and successful object detection method, namely DETR (Detection Transformer)~\cite{detr}, and introduce a novel query-guided detection transformer for videos -- \model{}. The \model{} leverages the visual representation of the query image to better localize the target object in the video frames. To handle the temporal context in videos, we incorporate cross-attention that captures information from neighboring frames of the video. The key and value matrices of the cross-attention are obtained from the contextual and target frames, and the query matrix is derived from the query image. The cross-attention operation produces the final context-aware feature representation, which inputs the query-guided DETR. To ensure stronger initialization of \model{} for our targeted task, we introduced an unsupervised, video-specific pretraining approach. During the pretraining phase, we employed a feature reconstruction loss combined with recurrent object queries to enhance representation quality and temporal learning.

To facilitate our study, we have extended two existing datasets, VidOR~\cite{vidor} and ImageNet-VidVRD~\cite{vidvrd}, by splitting them into train and test sets such that there are no common object categories between the two sets. We obtain query images from each object category from the Google open-images~\cite{OpenImages} dataset. This dataset preparation ensures a realistic evaluation of one-shot learning performance on videos containing unseen objects. Here, we must highlight that we employ category-wise splits to conduct experiments. However, our model remains entirely class-agnostic during the training and testing phases. Our experimental results demonstrate the effectiveness of our proposed one-shot video object localization approach on the extended VidOR~\cite{vidor} and ImageNet-VidVRD~\cite{vidvrd} datasets. By combining the query-guided DETR with the video-specific context module, our method exhibits robust performance in localizing instances of unseen objects in target videos.

The contributions of this chapter are twofold: (i) We proposed a novel category-free and query-guided extension to DETR. Our proposed model -- \model{} incorporates a temporal module, capitalizing on temporal context for superior object localization within videos. (ii) We introduced a novel video-specific unsupervised pretraining for \model{}. This pretraining objective is analogous to our downstream one-shot localization task, resulting in a notable improvement in the localization performance.

\section{Related Work}
\subsection{One/Few-Shot Object Detection}
Over the last few years, there has been significant progress in one/few-shot object detection, largely due to the adoption of sophisticated strategies such as attention mechanisms, transformers, and few-shot learning techniques~\cite{sun2021fsce,fan2020few,wu2020multi,kang2019few,wang2020frustratingly,fan2022few}. These strategies have been successful in a variety of domains, encompassing both image-based and video-based object detection, thus broadening the scope and application of these advanced detection methods. Recently, CoAE~\cite{hsieh2019one} and OS2D~\cite{osokin2020os2d} have shown significant advancements toward one-shot object detection in images. The CoAE~\cite{hsieh2019one} makes ingenious use of co-attention and co-excitation mechanisms to generate region proposals and highlight correlated feature channels, thereby improving the accuracy and efficiency of object detection. On the other hand, the OS2D~\cite{osokin2020os2d} model stands out as a versatile one-stage system that simultaneously performs localization and recognition tasks, proving its effectiveness in various detection scenarios. In the realm of few-shot object detection, recent innovations have resulted in the development of models like the feature reweighting-based model~\cite{kang2019few}, the Multi-level Feature Enhancement (MFE) model~\cite{wu2020multi}, the Counting-DETR~\cite{nguyen2022few} model for few-shot object counting and detection, and the Fast Hierarchical Learning~\cite{she2022fast} model designed to address the catastrophic forgetting issue. These models, each exploring different aspects of few-shot learning, have contributed to improved performance across various datasets and settings. In the context of video object detection, models like the Tube Proposal Network with Temporal Matching Network~\cite{nguyen2022few} and the Thaw method~\cite{yu2022few} for few-shot learning in video object detection have been proposed. Both models harness the power of attention mechanisms, transformers, and few-shot learning techniques to achieve compelling results. There have been several successful advances towards developing DETR~\cite{detr} variants in recent years for addressing different objectives; for example, in~\cite{Dong2022IncrementalDETRIF}, authors proposed an incremental DETR that can generalize to
novel classes with finetuning on a few examples. In~\cite{Jia2022DETRsWH} proposed different variants of object query, and~\cite{Jia2022DETRsWH} proposed an auxiliary one-to-many matching branch during training to enhance the performance of DETR for object detection in images. In~\cite{dai2021up} proposed UP-DETR method that enhances the original DETR~\cite{detr} model by introducing unsupervised pretraining with a random query patch detection pretext task, demonstrating considerable performance improvement. The advancements in these areas suggest that the continuous refinement of these techniques can lead to substantial improvements in object detection across image and video domains with limited supervision.
\begin{figure*}[t!]
    \centering
      \includegraphics[width=\textwidth]{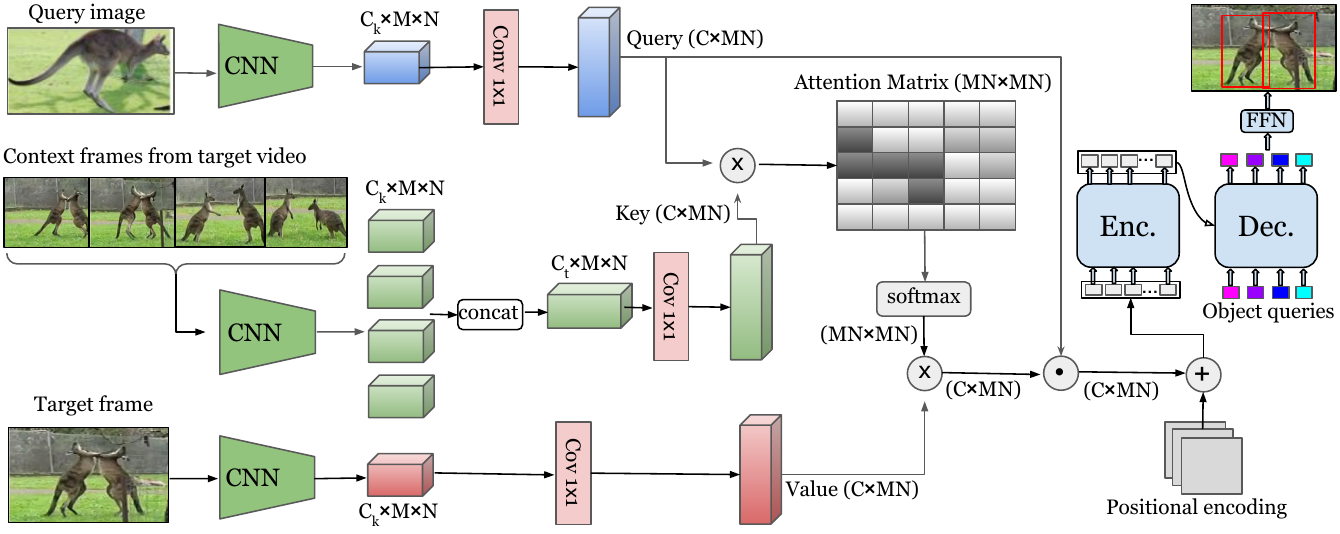}
  \caption[Illustration of the proposed \model]{\label{fig:aaai24model} Illustration of the proposed \model. The process begins with the feature extraction of a query image and video frames using a CNN encoder. A cross-attention mechanism and dot-product attention are used to create an attention map, transforming target frame features. The output is integrated into DETR's encoder, and predictions for bounding boxes are generated using the DETR decoder. [Best viewed in color]. }

\end{figure*}
\subsection{Transformer-Based Pretraining}
Pretraining methods based on transformers, such as BERT~\cite{bert} and GPT~\cite{brown2020language} for natural language processing, and ViT~\cite{dosovitskiy2020transformers} for computer vision, have revolutionized the landscape of various tasks, including object detection. These methods leverage self-attention mechanisms to learn high-level semantic features from large-scale data, resulting in models with robust generalization capabilities. By pretraining on large-scale data with diverse modalities, such as image-text pairs~\cite{owl}, these models can enhance the quality of representation learning, and also enable the transfer of this knowledge to downstream tasks, even when object-level data are scarce, thereby circumventing some of the limitations of traditional methods. Furthermore, unsupervised pretraining techniques, such as the random query patch detection proposed by UP-DETR~\cite{dai2021up}, have shown promise in enhancing the performance of object detection models. By capitalizing on the strengths of the transformer architecture in capturing long-range dependencies and spatial localization, these techniques offer exciting avenues for future research and development in object detection.
\section{Query-Guided DETR for Videos (\model)}
A key motivation behind QDETRv is that one-shot video localization requires identifying unseen objects from just a single query image, making it essential to exploit both the query’s visual cues and the temporal structure of the video. QDETRv achieves this by integrating query-guided cross-attention with frame-to-frame temporal context, enabling precise matching even when the target object appears with variations across time. Moreover, its video-specific unsupervised pretraining equips the model with strong spatio-temporal priors, allowing it to generalize beyond seen categories. Together, these capabilities make QDETRv particularly effective for open-set one-shot localization in videos.

Following, we discuss the formulation of the problem and architectural details.

Given a query image $I_q$ featuring an unseen object, the objective is to spatiotemporally localize all instances of the unseen object within a target video $V$. Videos, with their intricate interplay of spatial and temporal dimensions, present both opportunities and challenges for localization. Harnessing this complexity effectively ensures high precision and reduces false localization. To this end, in this work, we proposed \model{} that effectively harnesses the temporal context from the neighboring frames for better localization of unseen objects.

In this work, we utilize a query image $I_q$, a frame $I_t$, and a set of adjacent frames $S$ from the target video. These frames are defined as $S = \{I_{t+j}\}$, where $j=-k~to~k$ with temporal context window $2k+1$ for the $t^{th}$ frame of the video $V$.

\noindent\textbf{Feature Encoding:}\
The query image, target, and contextual frames are processed through a Convolutional Neural Network (CNN) encoder to generate corresponding feature representations. This results in the dimensions $I_q, I_t \in \mathbb{R}^{C_k\times M\times N}$ and $S \in \mathbb{R}^{C_t\times M\times N}$, where $M$ and $N$ are the height and width of the CNN feature maps and $C_k$, and $C_t$ are the channel dimensions.

\noindent\textbf{Cross-Attention Mechanism for Temporal Context:}\
In our work, we utilize the cross-attention mechanism~\cite{vaswani2017attention} to encode temporal context within the target frame and enable interaction between the query image and the target frame. As part of this process, we first generate Query $Q \in \mathbb{R}^{C\times MN}$, Key $K \in \mathbb{R}^{C\times MN}$, and Value $V \in \mathbb{R}^{C\times MN}$ matrices using convolution operations of kernel size $1\times 1$. Subsequently, we apply the dot-product attention~\cite{vaswani2017attention} between Query and Key, which results in an attention map. The latter serves as a storage of correspondences between the visual feature of the query image and the temporal context extracted from the video. This attention map is then used to transform the target frame features represented by Value. This process leads to output features $O \in \mathbb{R}^{C\times MN}$, which can be defined by the following equations:

\begin{equation}
W = \text{Softmax}\left(\frac{Q^TK}{\sqrt{C}}\right),
\end{equation}

\begin{equation}
O = VW^T \odot Q.
\end{equation}

\begin{figure*}[t!]
    \centering
      \includegraphics[width=\textwidth]{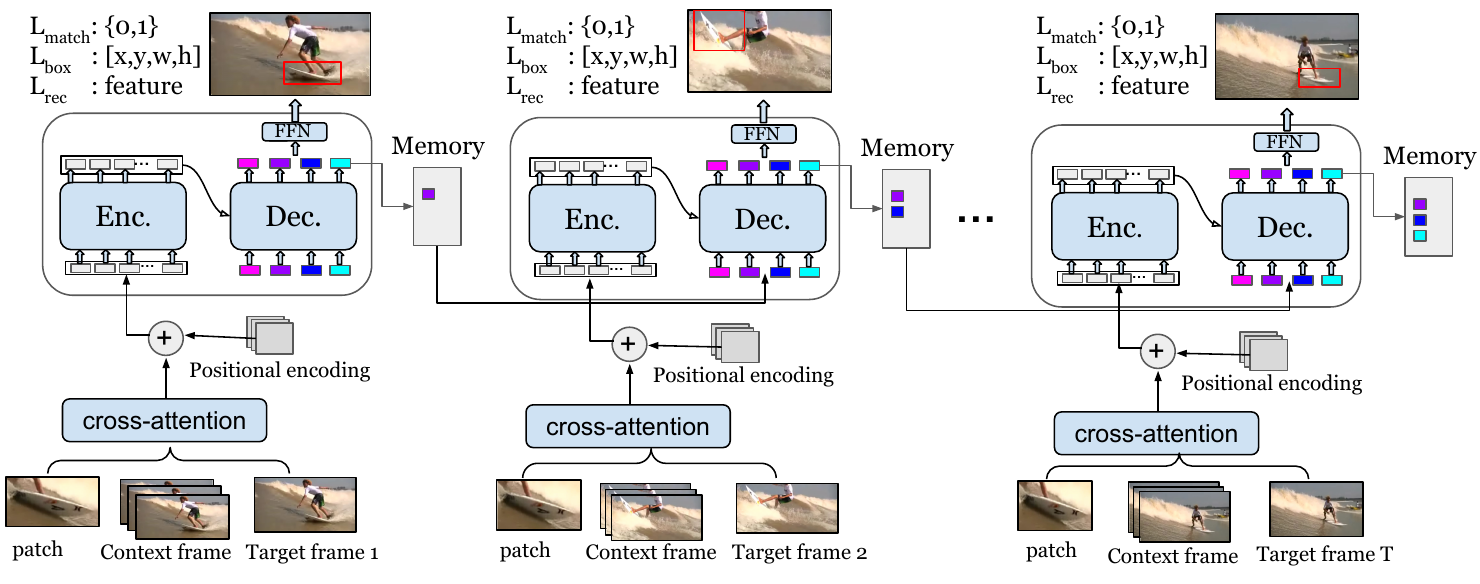}
    \caption[Unsupervised Video Pretraining]{\label{fig:pre} Unsupervised Video Pretraining: Each frame, along with its associated query patch and context frame, is fed into \model{}. The model detects the patch on every frame using decoded object queries responsible for predicting the bounding box. These decoded queries are then forwarded to the next frame as object queries for the decoder. Containing the semantic information extracted from previous frames for patch objects, they facilitate the detection in subsequent frames. During this phase, an unsupervised query reconstruction loss is also applied as an extra supervisory mechanism. The model has been pre-trained on a synthetically created dataset to enhance its performance. [Best viewed in color].}
\end{figure*}

Here, $\odot$ represents element-wise multiplication. The output features are enriched by adding positional embedding and are then integrated into the encoder of DETR~\cite{detr}. This allows the localization of unseen objects within the target frame, which is treated as a direct set problem similar to DETR~\cite{detr}. The decoder's role is to generate $N$ pairs of predictions $y = \{\hat{y_i}\}_{i=1}^N$ to accurately determine the bounding box of the query object in the current frame. Further, we leverage the Hungarian algorithm to calculate the matching cost between the prediction $\hat{y}_{\hat{\sigma}(i)}$ and the ground truth $y_i$. Here, $\hat{\sigma}(i)$ signifies the index of $y_i$ as computed by optimal bipartite matching.\\
\noindent\textbf{Loss Formulation:}
The predicted result $\hat{y_i} = (\hat{c_i} \in \mathbb{R}^2, \hat{b_i} \in \mathbb{R}^4, \hat{p_i} \in \mathbb{R}^C)$ comprises three elements: (a) $\hat{c_i}$, this binary classification determines whether a match with the query object is found ($c_i = 1$) or not ($c_i = 0$) for each object query. (b) $\hat{b}_i$, this vector is responsible for defining the box center coordinates along with its width and height, represented as ${x, y, w, h}$. (c) $\hat{p_i}$ is feature reconstruction used during pretraining for additional supervision. Taking into account these definitions, we define the Hungarian loss for all matched pairs as:

\begin{equation}
\mathcal{L}(y, \hat{y}) = \sum_{t=1}^{T}\sum_{i=1}^{N} \Bigg[ \lambda_{c_i} \mathcal{L}_{\text{match}}(c_i^{t}, \hat{c}_{\hat{\sigma}(i)}^{t}) + \mathds{1}_{{c_i=1}} \mathcal{L}_{\text{box}}(b_i^t, \hat{b}_{\hat{\sigma}(i)}^t) ) \Bigg].
\end{equation}

In this equation, $T$ is the number of frames, $\mathcal{L}_{match}$ refers to the binary cross-entropy loss over two classes (match the query object vs. not match), and $\lambda$ symbolizes the class balance weight. The $\mathcal{L}_{box}$ is a combination of $l_1$ loss and the generalized IoU loss, with the weight hyper-parameters being the same as those in DETR~\cite{detr}.

\subsection{Unsupervised Pretraining of \model{}}
Inspired by UP-DETR~\cite{dai2021up}, we introduce a video-specific pretraining strategy suitable for \model{}. This pretraining strategy is designed to locate a specific query patch within each frame, a process that aligns closely with our main objective. We devised a synthetic dataset based on the UCF101~\cite{soomro2012ucf101} dataset inspired by~\cite{wang2021unsupervised}. Our video-specific pretraining aims to identify and locate the query patch in every video frame, mimicking our main video frame localization task, as visualized in Figure~\ref{fig:pre}.
To ensure our pretraining approach is effective and aligns well with our primary task, we have added two additional mechanisms:
\noindent \textbf{(i)} We integrated an unsupervised query reconstruction loss, as suggested in~\cite{dai2021up}. This additional loss improves training signals and ensures high-quality embeddings are formed. The combined loss for this pretraining phase is represented as:

\begin{equation}
\mathcal{L}(y, \hat{y}) = \sum_{t=1}^T\sum_{i=1}^{N} \Bigg[ \lambda_{c_i} \mathcal{L}_{\text{match}}(c_i^t, \hat{c}_{\hat{\sigma}(i)}^t) + \mathds{1}_{{c_i=1}} \mathcal{L}_{\text{box}}(b_i^t, \hat{b}_{\hat{\sigma}(i)}^t) + \mathds{1}_{c_i=1}\mathcal{L}_{rec}(p_i^t, \hat{p}_{\hat{\sigma}(i)}^t) \Bigg].
\end{equation}

Here, $\mathcal{L}_{rec}$ represent the reconstruction loss defined below:

\begin{equation}
\mathcal{L}_{\text{rec}}(p_i^t, \hat{p}_{\hat{\sigma}(i)}^t) = \left\lVert \frac{p_i^t}{|p_i^t|_2} - \frac{\hat{p}_{\hat{\sigma}(i)}^t}{|\hat{p}_{\hat{\sigma}(i)}|_2} \right\rVert_2^2.
\end{equation}

\noindent \textbf{(ii)} To enhance the temporal context further, we used recurrent object queries. We fed decoded object queries from the decoder (which localizes the query patch) into the decoder again for predicting the patch in the subsequent frame, as depicted in Figure~\ref{fig:pre}. To incorporate recurrent queries, we used a memory module. The memory is a $256 \times 100$ matrix where $100$ is the number of object queries used in DETR, with each object query having a dimension of $256 \times 1$. During pretraining, for every frame, it stores decoded object queries (recurrent queries) responsible for localization. The recurrent queries carry temporal context from the previous frames and help to localize the object in the subsequent frames. Their utility is quantitatively evaluated in Table~\ref{tab:abl_recuurent_query}.\\
We also incorporated an unsupervised image-level pretraining for UP-DETR~\cite{dai2021up}. During this image-level pretraining phase, the primary objective is to localize a randomly selected patch within the input image. For this pretraining stage, we employed the ImageNet~\cite{ILSVRC15} dataset.

\begin{table}[t!h]
\centering
\small
\begin{tabular}{@{}lcccc@{}}
\toprule
Split & \multicolumn{2}{c}{Vid-OR} & \multicolumn{2}{c}{VidVRD} \\
\cmidrule(lr){2-3} \cmidrule(lr){4-5}
       & Seen & Unseen & Seen & Unseen \\
\midrule
\#Train Videos & 4,795 & 6,164 & 574 & 758 \\
\#Train Object Queries & 53,277 & 57,599 & 3,413 & 4,395 \\
\#Train Object Categories & 75 & 75 & 30 & 30 \\
\#Test Videos & 1,319 & 32 & 184 & 42 \\
\#Test Object Queries & 29,108 & 407 & 1,088 & 112 \\
\#Test Object Categories & 75 & 9 & 30 & 5 \\
\bottomrule
\end{tabular}
\caption[Overview of the Vid-OR and VidVRD datasets]{Overview of the Vid-OR~\cite{vidor} and VidVRD~\cite{vidvrd} datasets, summarizing the distribution of training and testing videos, object queries, and object categories for both seen and unseen splits.}
\label{tab:aaai26dataset_summary}
\end{table}
\section{Experiments and Results}
\subsection{Datasets and Performance Measure}
\label{sec:aaai24dataset}
In this work, we employed three primary datasets: VidOR~\cite{vidor}, ImageNet-VidVRD~\cite{vidvrd}, and Open Images~\cite{OpenImages}. We further refined VidOR and ImageNet-VidCRD annotations, segregating super-classes like \textit{fruits} and \textit{vegetables}. Extracted target frames from both datasets were paired with Open Images. Some low-frequency classes were excluded during training, but all frames were used in testing. Note that our method is class-agnostic and does not use the class information during training and testing. Dataset statistics are provided in Table \ref{tab:aaai26dataset_summary}.\\
\noindent{\textbf{Dataset for Pretraining:}} For effective visual representation learning, synthetic datasets are crucial given the challenging task of annotating real object movements in videos. We begin by generating a simulated trajectory resembling object movement. Consistency is ensured using bounding boxes at chosen keyframes, with in-between positions filled by linear interpolation. Patches from real video frames are then superimposed on this trajectory. The UCF101~\cite{soomro2012ucf101} dataset, featuring 13,320 videos with diverse complexities, was employed for pretraining, given its suitability to improve our approach.

\noindent{\textbf{Performance Measure:}}
We have utilized the mean Average Precision (mAP) at an Intersection over Union (IoU) of 0.5 as our primary evaluation metric. Our evaluation operates framewise, considering each frame as a separate detection problem.

\subsection{Baselines and Implementation Details}
The absence of prior work on class-agnostic one-shot video-based object localization methods inclines us to benchmark our proposed method against image-based techniques in a frame-wise fashion. However, such models might not be able to capture temporal aspects of the video.

\subsubsection{Retrieval-based approach}
A naive approach would involve combining an object detector that detects all objects as one class coupled with an image retrieval system for one-shot detection. This system utilizes the object detector's detections as a database and class objects as queries to search for relevant images. Inspired by baselines in~\cite{osokin2020os2d}, we utilized object detectors with identical architectures.

\subsubsection{CoAE~\cite{hsieh2019one}}
It uses non-local operations to explore the co-attention embodied in each query-target pair and yield region proposals accounting for the one-shot situation. It then formulates a squeeze-and-co-excitation that adaptively emphasizes correlated feature channels to help uncover relevant proposals and, eventually, the target objects. This method intentionally casts the learning formulation such that it does not solely rely on the label information of training data but instead explores correlated evidence revealed by the query-target pairs.

\subsubsection{OWL-ViT~\cite{owl}}
We used it as the one-shot image-conditioned object detection method for the video by applying it at each frame. OWL-ViT used a Vision Transformer architecture and contrastive image-text pretraining.

\subsubsection{UP-DETR~\cite{dai2021up}}
It refines the DETR model via unsupervised pretraining using a random query patch detection task. By selecting random image patches as queries and training the model to detect them, UP-DETR addresses challenges like balancing classification with localization by freezing the CNN backbone and managing multiple query patches with attention masks. This approach significantly boosts UP-DETR's performance in one-shot object detection, leveraging pretraining to achieve enhanced accuracy with minimal labeled data.

\subsubsection{Implementation Details}
We pre-train and fine-tune our models using the Adam optimizer~\cite{kingma2014adam} with an initial learning rate = 1e-5. The initialization of our frame and query encoders leverages weights of ResNet-50 pre-trained on the ImageNet dataset~\cite{ILSVRC15}. We train the model for $200$ epochs with batch size = $350$. Our implementation was done using the PyTorch library. We trained the model on three Nvidia-RTX A6000 GPUs.

\begin{table*}[t!]
\small
\centering
 \begin{tabular}{@{}lcccccc@{}}
\toprule
Method & Pretraining & \multicolumn{2}{c}{Vid-OR~\cite{vidor}} & \multicolumn{2}{c}{VidVRD~\cite{vidvrd}} \\
\cmidrule(lr){3-4} \cmidrule(lr){5-6}
       &              & Seen  & Unseen & Seen  & Unseen \\
\midrule
CoAE~\cite{hsieh2019one}     & -         & 35.3  & 32.8   & 29.2  & 27.1    \\
Retrieval-Based~\cite{osokin2020os2d}  &  - & 3.11    & 2.47     & 2.08  &  1.48       \\
OWL-ViT~\cite{owl}  & -         & -     & 28.6      & -  & 22.5       \\
UP-DETR~\cite{dai2021up}    & \xmark        & 37.4  & 35.4   & 33.7  & 28.5       \\
UP-DETR~\cite{dai2021up}    & \cmark       & 41.7  & 38.1   & 39.4  & 37.1       \\
Ours\\
~~\model   & \xmark        & 40.2  & 38.6   & 38.9  & 35.7       \\
~~\model   & \cmark     & \textbf{43.1}  & \textbf{41.6}      & \textbf{42.6}  & \textbf{38.5}       \\
\bottomrule
\end{tabular}
\caption[Comparative performance of different baselines]{Comparative performance of different baselines and our method on Vid-OR~\cite{vidor} and VidVRD~\cite{vidvrd} datasets. The metric for evaluation is mAP. \model~methods outperform other baselines across both datasets and splits, with pertaining significantly boosting the performance.}
\label{tab:performance_comparison}
\end{table*}

\section{Results and Discussions}
The results presented in Table~\ref{tab:performance_comparison} showcase the performance comparison of various methods applied for both of the datasets, Vid-OR~\cite{vidor} and VidVRD~\cite{vidvrd}. The methods assessed include baselines Retrieval-Based~\cite{osokin2020os2d}, OWL-ViT~\cite{owl}, CoAE~\cite{hsieh2019one}, UP-DETR~\cite{dai2021up} (\model{}~without temporal module), and our \model, with the latter two being tested both with and without pretraining.

\subsection{Performance of \model{}} For the Vid-OR dataset, \model{} with video pretraining yielded the best performance with $43.1$\% for seen and $41.6$\% for unseen classes, surpassing its performance without pretraining, which scored $38.6$\% and $40.2$\%, respectively. It's important to note that unsupervised video pretraining improved the results significantly, highlighting its effectiveness as a strategy for this task. We get similar observations for the ImageNet-VidVRD dataset.
\begin{figure}
\centering
\begin{tikzpicture}
    \begin{axis}[
        width=\linewidth,
        height=3.5cm,
        title={},
        xlabel={Temporal Context (Window Size)},
        ylabel={mAP},
        xmin=2, xmax=10,
        ymin=0.38, ymax=0.45,
        xtick={3,5,7,9},
        ytick={.35, .38,...,.45},
        legend pos=south east,
        ymajorgrids=true,
        grid style=dashed,
        tick label style={font=\small},
        label style={font=\small},
        legend style={font=\small},
    ]

    \addplot[
        color=green,
        mark=*,
    ]
    coordinates {
        (3, .40)
        (5, .404)
        (7, 0.418)
        (9, 0.427)
    };

    \end{axis}
\end{tikzpicture}
\caption[Ablation study on the temporal context window size]{\label{fig:tempa} Ablation study on the temporal context window size using the VidOR dataset. A positive correlation is observed between the window size and the mAP.}
\end{figure}
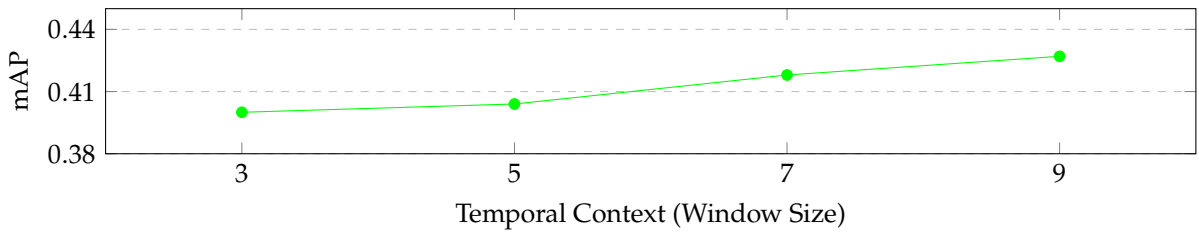

\begin{table}[t!]
\small
    \centering
    \begin{tabular}{c|cc}
    \hline
    Recurrent Query & VidVRD & VidOR\\
    \toprule
    \xmark          &  37.9  &  40.3\\
    \cmark          &  \textbf{38.5}  &  \textbf{41.6} \\
    \bottomrule
    \end{tabular}
    \caption[Ablation studies on recurrent queries during pretraining]{Ablation studies on recurrent queries during pretraining showed improved performance. This highlights the importance of adding temporal context in the pretraining phase.}
    \label{tab:abl_recuurent_query}
\end{table}

\subsection{Ablation Study on Temporal Context Module} To evaluate the effectiveness of the temporal context module integrated into \model{}, we conducted an ablation study where this module was removed. We exclusively adopted an image-level approach for unsupervised pretraining, eliminating the recurrent queries and leveraging the ImageNet~\cite{ILSVRC15} dataset. Without a context module with image-level pretraining, our method becomes similar to UP-DETR~\cite{dai2021up}. The results from this ablation demonstrated a notable $3.2$\% performance boost for \model{} without pretraining. Furthermore, with pretraining, there was a $1.4$\% enhancement in performance on VidOR~\cite{vidor} dataset for the unseen test set as shown in Table~\ref{tab:performance_comparison}. These findings confirm the significant contribution of the temporal context module to the model's efficacy, both in the presence and absence of pretraining.

\subsection{Fine-tuning with varied Temporal Context Windows} As we progressed to the fine-tuning stage, we explored different temporal context windows, spanning a range from $3$ to $9$, as depicted in Figure~\ref{fig:tempa}. Interestingly, our results indicated that as the temporal context window expanded, there was a corresponding enhancement in performance. This trend underscores the pivotal role of temporal context in improving localization accuracy. This accuracy curve should be read alongside its cost. For a window of size $2k+1$, the cross-attention in the temporal context module is computed over the target frame together with its $2k$ context frames, so both the FLOPs and the activation memory of that module grow linearly in the window size, while the encoder--decoder cost is unchanged. Widening the window therefore buys accuracy at a linear price, and the practical ceiling is set by memory rather than by any accuracy plateau. At the operating point we adopt, \model{} runs at 11 FPS with 7.6\,GB of peak memory, benchmarked under identical conditions against the methods reported in Chapter~\ref{chaptwo}. That comparison also indicates the preferable direction for deployment: rather than widening a fixed window, the framework of Chapter~\ref{chaptwo} propagates only high-confidence object features across frames and is both faster (14 FPS) and more accurate for a modest 0.5\,GB of additional memory, so selective propagation dominates window widening.

\begin{figure}[t!]
    \centering
      \includegraphics[width=0.95\textwidth]{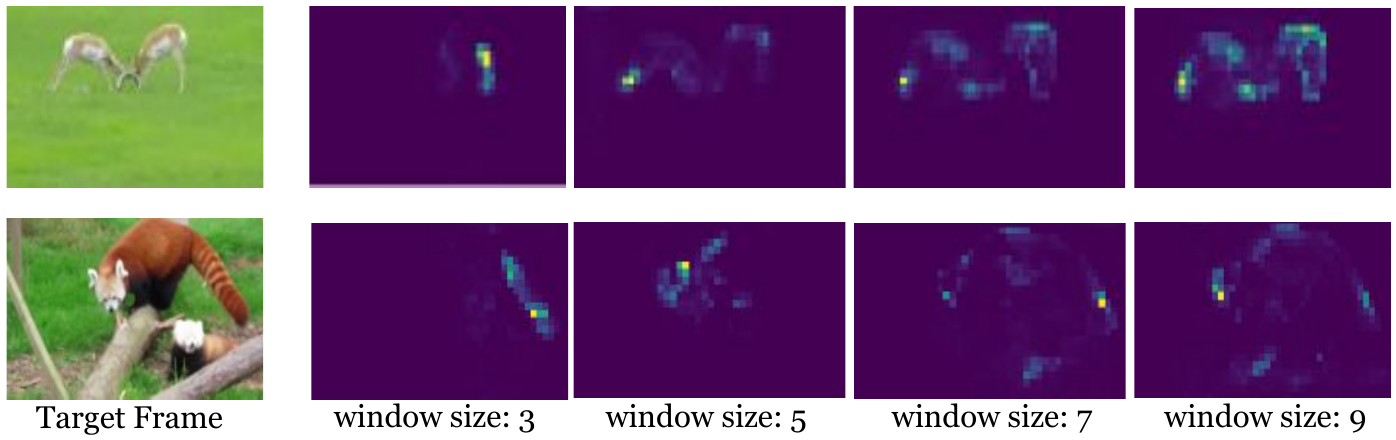}
    \caption[Cross-attention visualization on test frame]{\label{fig:att_viz} Cross-attention visualization on test frame with varying window size. Attention improves on objects of interest with an increase in window size.}
\end{figure}

\subsection{Cross-attention Visualization for varied temporal context} Figure~\ref{fig:att_viz} shows the cross-attention visualization on the test frame for different window sizes. The attention to the object of interest improves with an increase in the window size, showing the efficacy of temporal context being utilized from context frames of the video.\

\subsection{Ablation on Recurrent Queries} Table~\ref{tab:abl_recuurent_query} presents an ablation study focusing on the impact of recurrent queries during the pretraining phase. When recurrent queries are employed (indicated by \cmark), there's a noticeable performance improvement on both datasets: $0.6$\% on ImageNet-VidVRD~\cite{vidvrd} and $1.3$\% on VidOR~\cite{vidor}, compared to when these queries are omitted (\xmark). While seemingly modest, these increments emphasize the added value of recurrent queries in capturing intricate temporal information during pretraining.
\begin{table*}[t!]
\small
\centering
\setlength{\tabcolsep}{3.8pt}
\begin{tabular}{@{}lcccccccccc@{}}
\toprule
Method       & Bear & Carrot & Coconut & Frisbee & Kangaroo & Lemon & Leopard & Melon & Orange & Stop sign \\
\midrule

CoAE       & 37.8 & 19.2   & 33.6    & 25.4    & 42.6     & 21.3  & 46.9    & 22.8  & 28.1 & 30.6      \\
OWL-ViT       & 35.7 & 20.3   & 28.5    & 21.7    & 38.6     & 22.8  & 42.8    & 20.3  & 25.3 & 27.4      \\
UP-DETR         & 47.5 & 32.6   & 36.1    & 32.8    & 50.1     & 30.2  & 50.7    & 31.4  & 30.2 & 35.3      \\
\model{}        & \textbf{47.9} & \textbf{35.3}  &  \textbf{38.6}    &  \textbf{33.4}   & \textbf{51.4}    & \textbf{30.9}  & \textbf{52.3}  &  \textbf{33.2} & \textbf{30.8}    & \textbf{35.8}         \\
\bottomrule
\end{tabular}
\caption[Class-wise performance comparison]{Class-wise performance comparison. Each column indicates performance for distinct object categories. Larger objects like bears, kangaroos, and leopards are better detected than smaller ones like coconuts and lemons. }
\label{tab:class_wise}
\end{table*}
\subsection{Class-wise Performance}
Table~\ref{tab:class_wise} shows the class-wise performance on the VidOR dataset. Intriguingly, larger-sized objects like bear, kangaroo, and leopard generally achieve better performance compared to relatively smaller objects such as coconut and lemon. The model performs better on larger objects because they occupy more pixels and provide stronger visual features, making them easier to localize reliably. In contrast, small objects offer limited resolution and weaker feature cues, which reduces detection accuracy.

A promising direction to mitigate this issue within our current framework is to incorporate multi-scale attention or lightweight FPN-style feature fusion. By leveraging multi-level feature maps with higher spatial granularity, the model can better preserve fine details and improve localization of small objects. \\

\begin{figure*}[t!]
    \centering
      \includegraphics[width=0.95\textwidth]{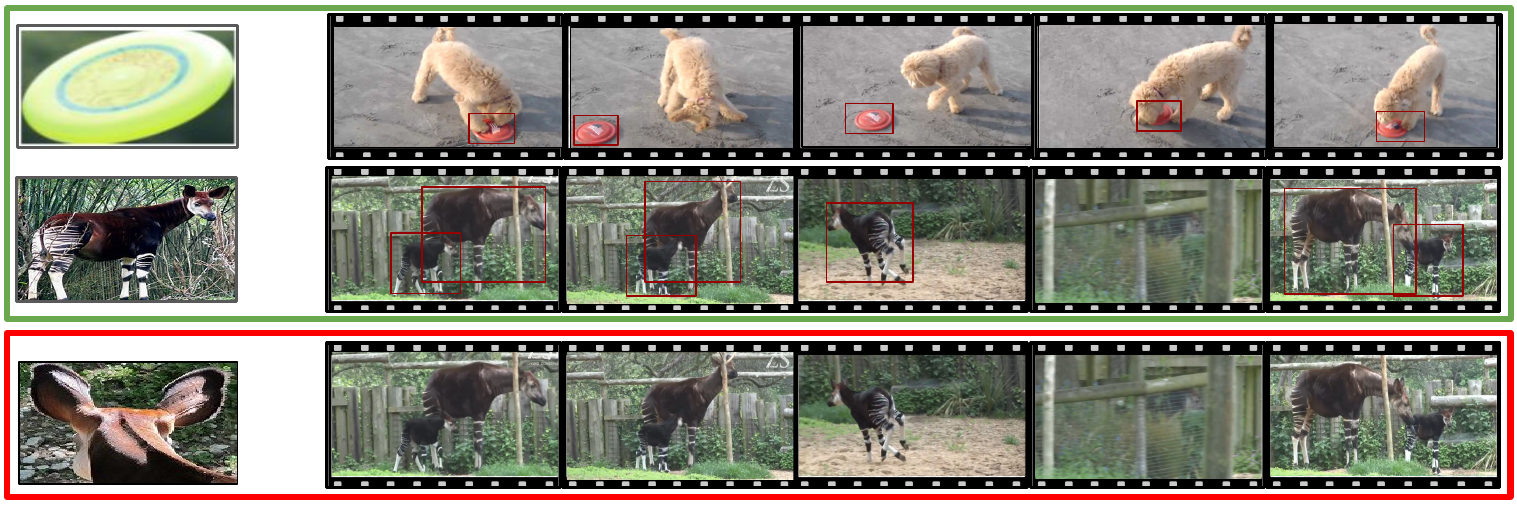}
    \caption[Qualitative results]{\label{fig:aaai24_qual} Qualitative results. In the green box results, \model{} accurately localizes objects from the query image on the left. The red box highlights the model's limitations, with missed localization in videos where the object is only partially visible in the query image.}
\end{figure*}

\begin{figure*}[t!]
    \centering
      \includegraphics[width=\textwidth]{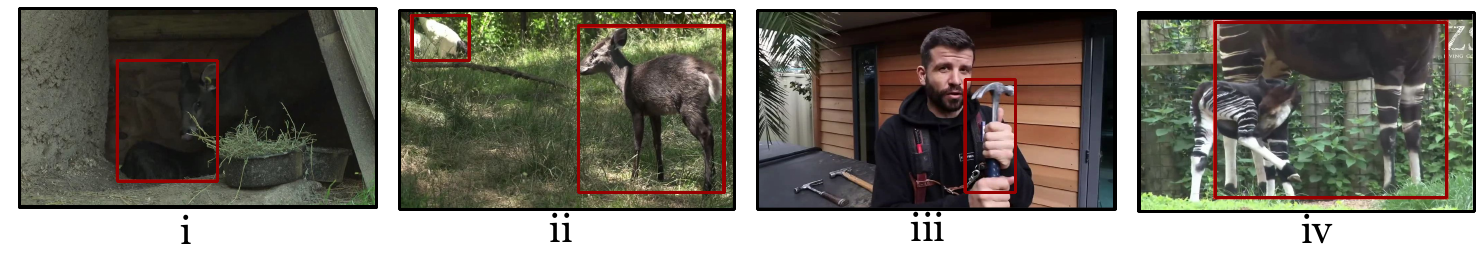}
    \caption[Failure Cases]{\label{fig:fail_case} Failure Cases. We observed the following major failure types: (i) The occluded object is missed, (ii) A visually similar object is detected as dear on the top left, (iii) a few small instances of the hammer are missed, (iv) an overlapping instance is commonly detected as a single object.}
\end{figure*}

\subsection{Qualitative Results}
\model{} effectively localizes query objects both spatially and temporally within the videos, as shown in the first two rows of Figure~\ref{fig:aaai24_qual}. Notably, in the second video, \model{} adeptly identifies multiple instances of the object. \\
\subsection{Failure Case Analysis}
While our proposed methodology has shown promising results, it does come with a few limitations: our model \model{} highly relies on the quality of the query image for object localization in videos. As shown in Figure~\ref{fig:aaai24_qual} (red box), occluded query images often lead to weak or missing localization. Further, we observe the following major failure cases: (i) occlusion, (ii) visually similar object, (iii) small object, and (iv) overlapping instances on randomly selected video frames. Figure~\ref{fig:fail_case} shows a selection of these examples. We leave addressing these limitations of the current model as the future scope of this work.
\section{Summary}
In this chapter, we introduced the novel class-agnostic query-guided DETR for videos (\model{}), effectively enhancing video object localization using a query image and temporal context. We showed evaluations on two extended datasets that displayed superior and consistent performance of \model{}, particularly when paired with video-specific unsupervised pretraining. While showing promise, our approach has a few limitations, including limited success in localizing small or occluded object instances. Further, it struggles to differentiate between visually similar objects. We leave addressing these limitations as future work and firmly believe this work will open up new research avenues in open-set one-shot video object localization.

In the following chapter, we generalize the one-shot object grounding paradigm to a few-shot setting and introduce a novel framework.

\chapter{Few-Shot Object Grounding in Videos}
\label{chaptwo}

Few-shot Video Object Detection (\textsc{FSVOD}) addresses the challenge of detecting novel objects in videos with limited labeled examples, overcoming the constraints of traditional detection methods that require extensive training data. This task presents key challenges, including maintaining temporal consistency across frames affected by occlusion and appearance variations, and achieving novel object generalization without relying on complex region proposals, which are often computationally expensive and require task-specific training. Our novel object-aware temporal modeling approach addresses these challenges by incorporating a filtering mechanism that selectively propagates high-confidence object features across frames. This enables efficient feature progression, reduces noise accumulation, and enhances detection accuracy in a few-shot setting. By utilizing few-shot trained detection and classification heads with focused feature propagation, we achieve robust temporal consistency without depending on explicit object tube proposals. Our approach achieves performance gains, with AP improvements of 3.7\% (FSVOD-500), 5.3\% (FSYTV-40), 4.3\% (VidOR), and 4.5\% (VidVRD) in the 5-shot setting. Further results demonstrate improvements in 1-shot, 3-shot, and 10-shot configurations.

The subsequent sections present the introduction, a review of related work, the proposed methodology, and a detailed analysis of the results and discussions.
\section{Introduction}
\begin{figure}[!t]
    \centering
      \includegraphics[width=0.85\textwidth]{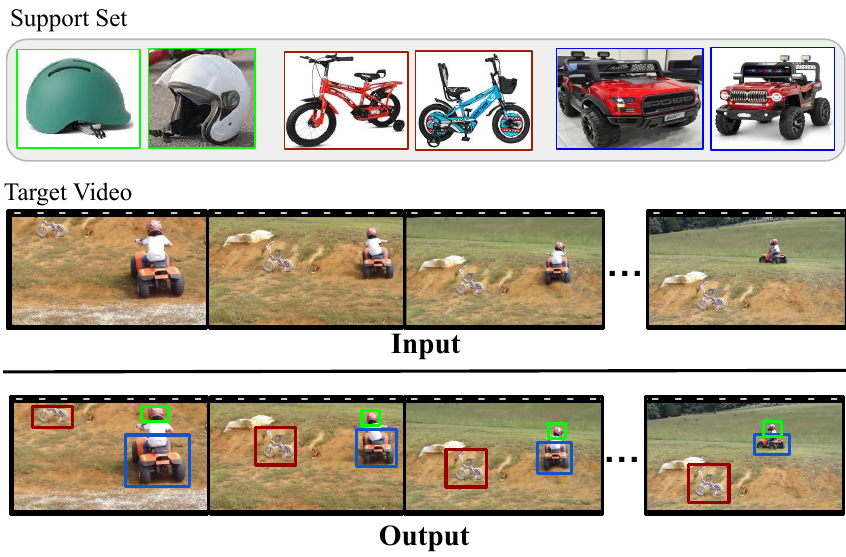}
    \caption[Few-shot Object Detection]{\label{fig:aaai26goal} Given a support set containing novel objects and a target video, the goal of \textsc{FSVOD}~\cite{fan2022few} is to detect all the instances of novel objects on the target video. We propose object-aware temporally consistent few-shot object detection framework that significantly improves state of the art for this task. [\textbf{Best viewed in color}].}

\end{figure}

Object detection in videos has witnessed significant progress with the advent of deep learning-based methods in the last few years~\cite{zhu2017flow,wu2019sequence,yu2022few,fan2022few}. However, most traditional methods focus on closed-set detection, where models are trained on predefined object categories with extensive labeled samples per category. This makes them impractical for many real-world applications that require the detection of novel object categories with limited training samples. Few-Shot Video Object Detection (\textsc{FSVOD})~\cite{fan2022few} addresses this challenge by enabling models to detect novel objects in videos using only a few examples as shown in Figure~\ref{fig:aaai26goal}.

Unlike widely-studied few-shot object detection in images~\cite{Dong2022IncrementalDETRIF,wang2020frustratingly,kang2019few,wu2020multi,sun2021fsce}, \textsc{FSVOD} must account for temporal consistency, motion blur, occlusion, and variations in object appearance between frames, making the task substantially more difficult~\cite{fan2022few}. Existing approaches for \textsc{FSVOD}~\cite{fan2020few,CenterTrack,han2023temporal,Kumar2024} face several limitations, such as they often process frames independently without leveraging temporal information~\cite{fan2020few}, struggle with false positives due to weak temporal matching~\cite{fan2022few}, use region or tube proposal networks not optimized for few-shot scenarios~\cite{fan2020few,fan2022few,han2023temporal}, and have difficulties with visually-similar objects and occlusions~\cite{Kumar2024}.

On the other hand, recent advances in vision-language models like Video OWL-ViT~\cite{video-owlvit23} have demonstrated impressive capabilities in open-vocabulary video object detection by leveraging CLIP's vision-language pre-training. Video OWL-ViT can be naturally adapted for few-shot video object detection by replacing text prompts with visual prompts from support images, enabling the detection of novel objects using only a few examples. However, this adaptation faces a key limitation, i.e., lack of object-aware temporal modeling, as it processes frames independently on the detected objects, resulting in inconsistent object detection across frames.

Our approach leverages vision-language pretrained frame encoder to transfer semantic knowledge from image-text data, enabling recognition of novel object categories. Furthermore, contrastively-trained vision encoder is more effective in distinguishing between visually similar objects across different categories and handles partially visible or occluded objects compared to CNN backbones (like ResNet~\cite{resnet}) used in prior methods~\cite{fan2022few,han2023temporal,Kumar2024}. This robust feature representation serves as the foundation for our temporal consistency mechanism. To enhance this consistency, we fuse representations of previously detected objects with current frame detections, preserving relevant temporal information and adapting to appearance changes over time. Unlike methods that process frames independently, our approach maintains memory of past observations for better temporal reasoning. Our detection is directly optimized to condition on few-shot visual examples, providing better results compared to traditional region proposal networks. Further, unlike conventional tracking-by-detection pipelines~\cite{fan2022few,han2023temporal}, our end-to-end learning of video-specific object representations leads to superior generalization.

In summary, in this chapter we make the following contributions: (i) We revisit the relatively under-explored task of Few-Shot Video Object Detection (\textsc{FSVOD}) and introduce simple yet effective task-specific innovations. Our key contribution is a novel object-aware temporal modeling approach, that selectively propagates high-confidence object features across frames. This enables efficient feature progression, reducing noise accumulation, and improving detection accuracy in few-shot video scenarios.
(ii) For the first time, we adapt large-scale pre-trained vision-language models for \textsc{FSVOD}, aligning with recent trends in open-world detection. This integration enables novel object detection by leveraging broad visual-linguistic knowledge, facilitating efficient adaptation and strong generalization to unseen categories. (iii) We perform extensive evaluations against baselines and state-of-the-art methods on four public \textsc{FSVOD} benchmarks. Our approach achieves significant performance gains, with AP improvements of 3.7\% (FSVOD-500), 5.3\% (FSYTV-40), 4.3\% (VidOR), and 4.5\% (VidVRD) in the 5-shot setting. Further, we observe consistent improvements across 1-shot, 3-shot, and 10-shot setups, demonstrating robust generalization across different few-shot scenarios.

\section{Related Work}
\noindent{\textbf{Few-Shot Object Detection}}:
Recent advances in few-shot object detection (\textsc{fsod}) have been driven by attention mechanisms, transformers, and meta-learning techniques. Approaches like CoAE~\cite{hsieh2019one} and OS2D~\cite{osokin2020os2d} have pushed the boundaries of one-shot object detection in images through co-attention mechanisms and versatile one-stage approach, respectively. For traditional \textsc{fsod}, methods such as feature reweighting~\cite{kang2019few}, contrastive proposal encoding~\cite{sun2021fsce}, and hierarchical learning~\cite{she2022fast} have demonstrated improved performance across various benchmarks. In addition, several DETR variants~\cite{Dong2022IncrementalDETRIF,Jia2022DETRsWH,dai2021up} have emerged and shown promising results for few-shot object detection in images.
Although \textsc{fsod} in images provides the foundational techniques and can be trivially extended to videos by performing few-shot detection on frames independently.
Our experiments suggest that they often fall short. The inherent temporal nature of videos introduces additional complexities that require specialized approaches to maintain consistency across frames while preserving the few-shot learning paradigm. We introduce object-aware temporal consistency to overcome the inherent challenges specific to video.

While few-shot object detection in images has been extensively studied, its video counterpart remains relatively under explored. Qi et al.~\cite{fan2022few} have formally introduced a few-shot video object detection task and accompanying datasets. They proposed a proposal-based approach, where object detection trajectories are first proposed and later refined using the matching network. Kumar et al.~\cite{Kumar2024} have specifically studied this problem for one-shot settings using a novel query-guided variant of DETR. More recently, Wentano et al.~\cite{han2023temporal} enhanced temporal information by context fusion. Our work addresses key limitations of existing \textsc{FSVOD} methods by utilizing a transformer-based backbone (OWL-ViT) pre-trained on large-scale image-text pairs, replacing the CNN-based backbones used in prior approaches and improves semantic knowledge transfer. In addition, unlike proposal-based networks not optimized for few-shot scenarios, our proposal-free method conditions detection directly on few-shot-trained heads and improves generalization to novel object categories.

\noindent{\textbf{Open-World Object Detection}}: There is a fundamental shift in computer vision literature toward detecting novel visual concepts driven by models pre-trained on large-scale vision and language data~\cite{llm_fsod,mml_clas}. These large models help in detecting novel objects by leveraging their broad visual and linguistic knowledge, efficient adaptation, and improved generalization to unseen categories. Recent works like OWL-ViT~\cite{owl} and Video OWL-ViT~\cite{video-owlvit23} leverage vision-language models pretrained on large-scale data to detect open-vocabulary concepts. Methods like GLIP~\cite{li2022grounded} and OpenSeed~\cite{zhang2023simple} further advance this paradigm by incorporating grounded pretraining and structured knowledge distillation.  Open-world object detectors can be extended for image-conditioned detection, enabling users to detect objects by providing a few visual examples. In this work, we leverage OWL-ViT encoders for image-conditioned detection in a few-shot setup and enhance the existing Video OWL-ViT~\cite{video-owlvit23} by introducing an object-aware decoder that significantly improves temporal consistency across video frames, and thereby the few-shot video object detection performance.

\section{Method}
Few-shot video object detection (\textsc{FSVOD}) aims to detect novel object categories that were not encountered during training. The task involves a \textit{support set} containing a few images of novel object classes and a \textit{target video} consisting of multiple frames. The goal is to identify and localize all instances of the objects from the support set across every frame of the target video, as shown in Figure~\ref{fig:aaai26goal}.

\subsection{Problem Formulation}
In the \textsc{FSVOD} task, we aim to locate and classify all instances of objects belonging to novel classes in a target video using only a few support examples. Formally, given an $N$-way $K$-shot support set $\mathcal{S} = \{(\mathbf{I}_{i,j}, c_i) | i = 1,\cdot,N; j = 1,\cdot,K\}$ where $\mathbf{I}_{i,j}$ represents the $j$-th support image of class $c_i$, and a target video $\mathcal{V} = \{\mathbf{F}_t | t = 1,...,T\}$ with $T$ frames, our goal is to detect all instances of the $N$ novel classes across all frames. For each frame $\mathbf{F}_t$, the model predicts a set of object bounding boxes along with their class labels.

\subsection{Design Motivation and Architectural Overview}
Our proposed approach is designed to address the following unique challenges associated with \textsc{FSVOD} task: (i) the limited supervision requiring efficient knowledge transfer, (ii) the necessity for temporal consistency across sequential frames despite appearance variations due to motion, occlusion, and viewpoint shifts, and (iii) the open-set recognition capability to distinguish between novel objects and background elements without explicit training.

\begin{figure}[t!]
    \centering
      \includegraphics[width=0.68\linewidth]{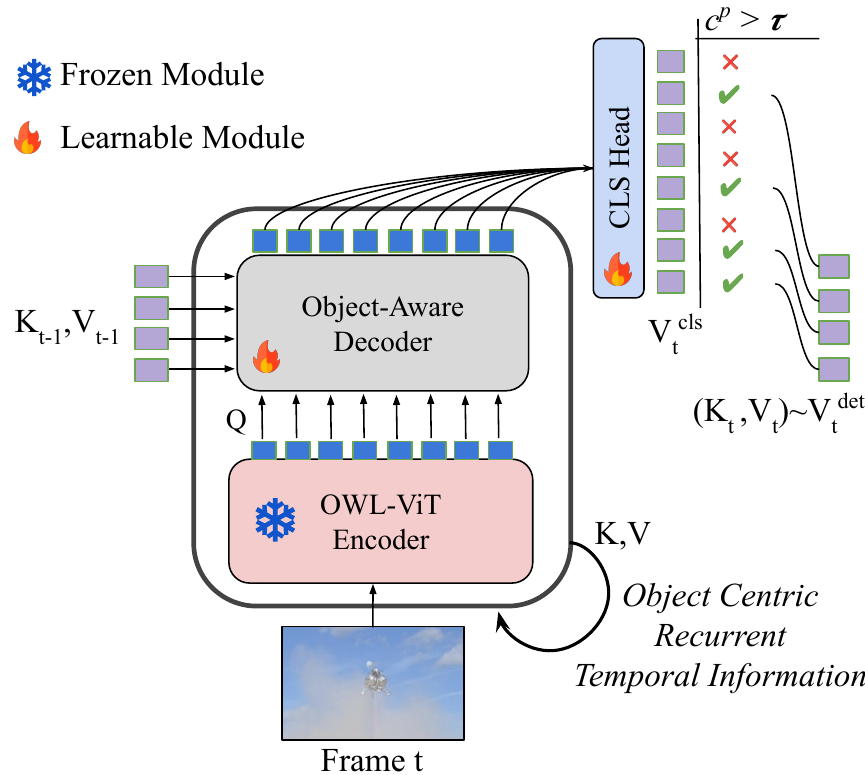}
  \caption[Object-aware Temporal Consistency]{\label{fig:model1}Object-aware Temporal Consistency. We utilize the OWL-ViT encoder output tokens as queries but, critically, only forward the matched object tokens ($c^p > \tau$) as key-value pairs to the next frame's decoder. This mechanism selectively propagates high-confidence object features across frames. This enables efficient feature progression, reducing noise accumulation. We empirically found a threshold ($\tau$) and compared it with class probability ($c^p$) scores for selective propagation. This selective propagation mechanism significantly reduces noise accumulation across frames and maintains focused, consistent visual representations of detected objects throughout the video sequence.}
\end{figure}

\begin{figure}[!t]
    \centering
    \includegraphics[width=0.7\linewidth]{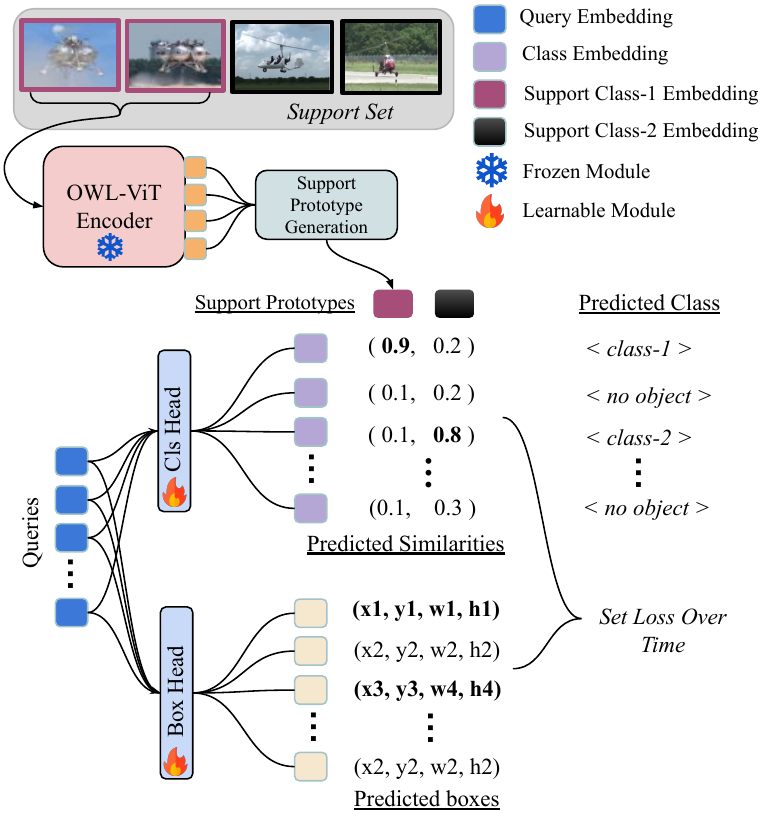}
    \caption[Few-shot heads]{\label{fig:model2}Few-shot Classification and Detection Heads. Our architecture processes object queries through parallel projection heads for classification and localization. Classification embeddings are compared against support embeddings via cosine similarity, while the detection head predicts bounding box coordinates. The temporal consistency is maintained by propagating matched object queries across frames.}
\end{figure}

To address these challenges, we propose an object-aware, temporally consistent few-shot video object detection framework. Our approach consists of the following main components: a language-aligned vision encoder that provides semantically rich visual representations, a temporal fusion decoder that selectively propagates high-confidence object features across frames and, few-shot matching and detections heads that aligns query frame features with support examples. Figures~\ref{fig:model1},\ref{fig:model2} illustrates our overall framework.

\subsection{Object-aware Temporally consistent FSVOD framework}
\subsubsection{Language-Aligned Vision Encoder:}
\label{sec:l_a_enc}
Traditional CNN backbones like ResNet~\cite{resnet}, trained solely on visual classification tasks, often struggle to generalize to novel classes with limited examples. To overcome this limitation, the recent trend~\cite{llm_fsod,video-owlvit23} is to use a pretrained vision encoder that has been aligned with language semantics through large-scale vision-language pretraining.
We adopt this by utilizing an OWL-ViT encoder pretrained on large-scale image-text pairs. This encoder transforms each input image $\mathbf{I}$ into a grid of patch embeddings $\mathbf{E} \in \mathbb{R}^{P \times D}$, where $P$ is the number of patches and $D$ is the embedding dimension.

\subsubsection{Support Set Encoding:}

\textbf{(i) Support Feature Extraction.}
Given a support set $\mathcal{S}$ containing $K$ examples for each of the $N$ novel classes, we extract patch-level features from each support image. For each support image $\mathbf{I}_{i,j}$, the encoder generates both patch embeddings $\mathbf{E}_{i,j} \in \mathbb{R}^{P \times D},$ and their corresponding objectness scores for each patch,  $\mathbf{s}_{i,j} \in \mathbb{R}^P$.
A higher objectness score indicates a higher probability of an object being present in that particular patch. To obtain an object-centric representation, we select the patch with the highest objectness score.
\begin{equation}
p^* = \arg\max_p s_{i,j}^p, \quad \mathbf{z}_{i,j} = \mathbf{e}_{i,j}^{p^*},
\end{equation}
where $\mathbf{z}_{i,j}$ is the selected patch embedding that best represents the target object in the support image.
\textbf{(ii) Support Prototype Generation.} After extracting object-centric features ($\mathbf{z}_{i,j}$) from each support image, we aggregate information from the $K$ support examples to form a class prototype. For each class $c_i$, the final class prototype $\mathbf{z}_i$ is computed by averaging across the $K$ selected object-centric features.
\noindent This aggregation creates a robust representation for each novel class that captures essential visual characteristics across multiple support examples, allowing the model to handle intra-class variations more effectively.

\subsubsection{Object-Aware Temporal Frame Decoder:}
\label{sec:o_temp_mod}
\textbf{(i) Target Frame Feature Extraction.} For each frame $\mathbf{F}_t$ in the target video, we extract patch embeddings using the OWL-ViT image encoder, $\mathbf{F}_t \in \mathbb{R}^{P \times D}$.
\textbf{(ii) Temporal Fusion.} To ensure temporal consistency in object detection across frames, we propose a filtering mechanism that selectively propagates high-confidence object features from one frame to the next. This approach facilitates efficient feature progression and minimizes noise accumulation by leveraging the embeddings of successfully detected objects from previous frames. By doing so, it enhances temporal consistency and improves the overall accuracy of object detection in video sequences.

For the first frame ($t=0$), we process the frame embeddings directly. For subsequent frames ($t>0$), we incorporate information from previous frame detections via a cross-frame interaction that selectively integrates object-level features to enhance temporal consistency. Specifically, we select embeddings from the few-shot classification head ($\mathbf{V}_t^{cls}$) of the previous frame where detection confidence exceeds a threshold $\tau$ (shown in Figure~\ref{fig:model1}):
\begin{equation}
\mathbf{V}_{t-1}^{\text{det}} = \{\mathbf{v}_{t-1}^p |\hat{c}_{t-1,i}^p > \tau\},
\end{equation}
where $\hat{c}_{t-1,i}^p$ represents the classification probability for patch $p$ and class $i$ in frame $t-1$. The current frame embeddings $\mathbf{F}_t$ serve as queries, while the embeddings from Cls Head with high confidence score responsible for object detection (Figure ~\ref{fig:model1}), $\mathbf{V}_{t-1}^{\text{det}}$ from the previous frame serve as both keys and values. The cross-frame interaction operation is computed as:
\begin{equation}
\mathbf{A}_t = \text{Softmax}\left(\frac{\mathbf{F}_t (\mathbf{V}_{t-1}^{\text{det}})^T}{\sqrt{D}}\right).
\end{equation}

\noindent The resulting attention-weighted features are used to update the current frame representations:
\begin{equation}
\hat{\mathbf{F}}_t = \mathbf{F}_t + \mathbf{A}_t\mathbf{V}_{t-1}^{\text{det}}.
\end{equation}

\noindent These temporally enhanced features $\hat{\mathbf{F}}_t$ are then fed to the classification and localization heads for object detection in the current frame. By using attention between current frame features and previously detected objects representations, the model maintains better temporal consistency while detecting objects in the current frame.

\subsubsection{Few-Shot Detection Heads:}
\label{sec:fs_heads}
\noindent The temporally enhanced frame embeddings $\hat{\mathbf{F}}_t$ are processed through parallel classification and localization heads to produce the final detection results.

\noindent \textbf{Classification Head.} The classification head compares frame embeddings with support class prototypes to determine object categories. It projects frame embeddings to the classification space:
\begin{equation}
\mathbf{V}_t^{cls} = \mathbf{W}_{\text{cls}}\hat{\mathbf{F}}_t,
\end{equation}
where $\mathbf{W}_{\text{cls}}$ is a learnable projection matrix.

\noindent For each patch embedding $\mathbf{v}_t^p$ and class prototype $\mathbf{z}_i$, we compute cosine similarity:
\begin{equation}
s_{p,i} = \frac{\mathbf{v}_t^p \cdot \mathbf{z}_i}{||\mathbf{v}_t^p|| \cdot ||\mathbf{z}_i||}.
\end{equation}

\noindent Each patch is assigned a probability distribution over the $N$ classes and background using softmax normalization of similarity scores, translating feature similarity into class probabilities.

\noindent \textbf{Localization Head.} The localization head predicts bounding box coordinates for detected objects through a multi-layer perceptron:
\begin{equation}
\hat{\mathbf{B}}_t = \text{MLP}_{\text{box}}(\hat{\mathbf{F}}_t),
\end{equation}
where each row $\hat{\mathbf{b}}_t^p \in \mathbb{R}^4$ represents the predicted bounding box parameters $(x, y, w, h)$ for each corresponding patch $p$. For each patch $p$, if its maximum classification score exceeds threshold $\kappa$, we consider it as a valid detection.

\subsubsection{Training Objective:}
\noindent Our model is trained end-to-end by optimizing a combined loss function across all video frames:

$$
\begin{aligned}
\mathcal{L} = \sum_{t=1}^{T} \Big[ & \sum_{m=1}^{M_t} \lambda_{\text{cls}} \cdot \mathcal{L}_{\text{cls}}(\hat{\mathbf{c}}_t^{\sigma_t(m)}, c_t^m) \\
& + \sum_{m=1}^{M_t} \lambda_{\text{box}} \cdot \mathcal{L}_{\text{box}}(\hat{\mathbf{b}}_t^{\sigma_t(m)}, \mathbf{b}_t^m) \Big].
\end{aligned}
$$

\noindent The classification loss (\(\mathcal{L}_{\text{cls}}\)) uses cross-entropy to measure discrepancy between predicted and ground-truth class labels, while the localization loss (\(\mathcal{L}_{\text{box}}\)) combines L1 loss for coordinate distances and generalized IoU loss for handling varying box sizes. Following DETR~\cite{detr}, $\sigma_t(m)$ denotes the index of the prediction matched to ground-truth object $m$, and $M_t$ is the number of objects in frame $t$. The hyperparameters $\lambda_{\text{cls}}$ and $\lambda_{\text{box}}$ control the weighting of the classification and localization losses.

\section{Experiments and Results}

\subsection{Datasets}
\label{sec:aaai26dataset}
In this work, we utilized the FSVOD-500, FSYTV-40, VidOR, and VidVRD datasets, as proposed and repurposed in~\cite{fan2022few,Kumar2024}. These datasets are specifically designed for few-shot video object detection, providing a structured evaluation framework with disjoint training, validation, and testing classes. Detailed statistics of the datasets are provided in Table~\ref{tab:cvpr23dataset_summary}.

\subsection{Baselines}
We evaluate our approach against temporal and non-temporal based few-shot object detection methods. The selection of baselines aims to provide comprehensive comparisons across different architectural paradigms and learning approaches. We exclude some results on certain datasets due to the unavailability of public codebases, such as TACF~\cite{han2023temporal} on VidVRD and VidOR.

\noindent \textbf{Non-temporal Methods.} To establish strong non-temporal baselines, we compare against several image-based few-shot detectors as discussed and compared in \cite{fan2022few}. Traditional object detectors like Faster R-CNN~\cite{FasterRCNN} serve as fundamental baselines to validate the benefits of few-shot learning techniques. We include specialized few-shot detectors from ~\cite{fan2022few} such as FSOD~\cite{fan2020few}, RDN~\cite{RDN}. Further we included CoAE~\cite{hsieh2019one} designed for one-shot object detection and we extended it for few-shot detection by finding the best confident detections. CoAE leverage meta-learning and attention mechanisms to handle limited training data. Recent transformer-based approaches like OWL-ViT~\cite{owl} and UP-DETR~\cite{wang2021unsupervised} are included to evaluate the effectiveness of large-scale pre-training. We also compare against a retrieval-based methods~\cite{radenović2016cnn} that consider few-shot detection as the retrieval from the pool of proposals.

\noindent \textbf{Temporal Methods.}
For temporal modeling comparisons, we include several video-based approaches. First, we incorporate methods adapted in \cite{fan2022few}, including MEGA \cite{MEGA}, Ctracker \cite{CTracker}, FairMOT \cite{FairMOT}, and CenterTrack \cite{CenterTrack}. Second, we evaluate modern tracking-by-detection approaches, DeepSort+YOLOv8 \cite{deepsort}, TrackFormer \cite{trackformer}, and ByteTrack \cite{bytetrack} combined with CLIP \cite{clip} to measure similarity between detected object tubes and support images. For these approaches, we train models using examples from the support set to identify spatio-temporal tubes, then apply average pooling to compute similarities with support images, selecting tubes with maximum similarity across support images. Third, we compare against SoTA one/few-shot video detectors, including FSVOD \cite{fan2022few}, TACF \cite{han2023temporal}, and QDETRv \cite{Kumar2024}, which provide direct comparisons with our proposed method. Finally, we adapt Video Owl-ViT \cite{video-owlvit23} by replacing its text-encoder with an OWL-ViT image-encoder and incorporating our few-shot detection and classification heads.

\subsection{Performance Measure}
Following~\cite{fan2022few}, we use Average Precision (AP) as evaluation metric. AP quantifies object detection accuracy by measuring the area under the precision-recall curve. AP50 and AP75 are variants that evaluate performance at specific Intersection over Union (IoU) thresholds of 50\% and 75\%, respectively.
\begin{table}[t!]
\centering
\begin{tabular}{lcccc}
\toprule
Split & Vid-OR & VidVRD & FSVOD & FSYTV\\
\midrule
\#Train Videos & 6,164 & 758 & 2553 & 1627 \\
\#Train Object Queries & 57,599 & 4,395 & - & - \\
\#Train Object Categories & 75 & 30 & 320 & 30 \\
\#Test Videos & 32 & 42 & 949 & 608 \\
\#Test Object Queries & 407 & 112 & - & - \\
\#Test Object Categories & 9 & 5 & 100 & 10\\
\bottomrule
\end{tabular}
\caption{Overview of the datasets used in this work.}
\label{tab:cvpr23dataset_summary}
\end{table}
\subsection{Implementation Details}
We implement our model using a OWL ViT-L/16 for the vision encoder, pretrained with CLIP. The temporal fusion module employs a 4-head cross-attention mechanism with 1024-dimensional hidden states. The classification and localization heads use 2-layer MLPs with 512-dimensional hidden states. During training, we use AdamW optimization with a 1e-5 learning rate, 0.01 weight decay, and cosine scheduling with a linear warmup. The loss weighting parameters are set to $\lambda_{\text{cls}}=2$, $\lambda_{\text{box}}=5$. We set $\tau = 0.94$ and $\kappa = 0.98$ in our experiments. Our model was converged in four days using two Nvidia-A6000 GPUs.

\begin{table*}[!th]
\scriptsize
\centering
\begin{tabular}{ll ccc ccc ccc ccc}
\toprule
&\multirow{2}{*}{Method}  & \multicolumn{3}{c}{FSVOD-500} & \multicolumn{3}{c}{FSYTV-40} & \multicolumn{3}{c}{VidOR} & \multicolumn{3}{c}{VidVRD} \\
\cmidrule(lr){3-5} \cmidrule(lr){6-8} \cmidrule(lr){9-11} \cmidrule(lr){12-14}
& & $AP$ & $AP_{50}$ & $AP_{75}$ & $AP$ & $AP_{50}$ & $AP_{75}$ & $AP$ & $AP_{50}$ & $AP_{75}$ & $AP$ & $AP_{50}$ & $AP_{75}$\\
\midrule
\multirow{7}{*}{\rotatebox{90}{\textbf{Non-temporal }}}&
FR-CNN  & 18.2 & 26.4 & 19.6 & 9.3 & 15.4 & 9.6  & 21.6 & 29.1 & 23.2 & 14.3 & 18.6 & 14.2 \\
&FSOD & 21.1 & 31.3 & 22.6 & 12.5 & 20.9 & 13.0 & - & - & - & - & - & - \\
&RDN & 18.2 & 27.9 & 19.7 & 8.1 & 13.4 & 8.6 & - & - & - & - & - & - \\
&CoAE &18.4&29.3&18.9&9.8&18.4&10.1&35.8&43.7&36.1&30.2&38.1&31.8\\
&Retrieval-Based &1.3&2.4&1.6&0.6&1.1&0.8&1.4&2.6&1.3&1.8&2.5&2.0\\
&Owl-ViT &14.8&25.3&16.7&10.6&19.8&11.2&30.1&37.3&31.2&23.6&32.1&23.9\\
&UP-DETR &20.1&27.6&21.2&11.8&19.4&12.1&39.7&47.5&40.0&11.6&18.3&12.5\\
\midrule
\multirow{12}{*}{\rotatebox{90}{\textbf{Temporal}}}
&DeepSort+YOLOv8 + CLIP &13.2&21.4&14.4&5.3&9.4&6.2&23.2&30.1&23.6&23.4&33.1&24.1\\
&TrackFormer + CLIP &14.2&22.6&15.3&6.2&10.5&6.8&25.3&32.7&26.2&25.4&33.8&27.3\\
&ByteTrack  + CLIP &14.7&23.9&15.8&7.9&15.6&8.1&24.9&31.4&25.3&25.6&36.2&26.2\\
&MEGA  & 16.8 & 26.4 & 17.7 & 7.8 & 13.0 & 8.3 & - & - & - & - & - & -  \\
&CTracker  & 20.1 & 30.6 & 21.0 & 8.9 & 14.4 & 9.1&-&-&-&-&-&- \\
&FairMOT & 20.3 & 31.0 & 21.2 & 9.6 & 16.0 & 9.5&-&-&-&-&-&- \\
&CenterTrack & 20.6 & 30.5 & 21.9 & 9.5 & 15.6 & 9.7&-&-&-&-&-&- \\
&TACF& \underline{26.9} & \underline{39.2} & \underline{27.9} & \underline{15.9} & 24.2 & \underline{17.9} &-&-&-&-&-&-\\
&Video Owl-ViT  & 25.8 & 36.5 & 26.1 & 15.7 & \underline{26.3} & 15.9 & 44.3 & 52.9 & 45.7 & \underline{44.2} & \underline{53.7} & \underline{45.4}\\
&FSVOD &  25.1 & 36.8 & 26.2 & 14.6 & 21.9 & 16.1& \underline{45.1}& \underline{54.3}& \underline{46.2}&40.7&53.5&43.7\\
&QDETRv & 26.1 & 33.8 & 23.7 & 14.1 & 22.8 & 15.3 &43.4&51.8&44.6& 42.8& 50.8& 41.7 \\
&Ours & \textbf{30.6} & \textbf{42.9} & \textbf{32.1} & \textbf{21.2} & \textbf{29.8} & \textbf{23.5} &\textbf{49.4}&\textbf{57.3}&\textbf{50.2}&\textbf{48.7}&\textbf{57.8}&\textbf{49.2} \\
\bottomrule
\end{tabular}
\caption[Performance Comparision in 5-shot setting]{Performance comparison on FSVOD-500, FSYTV-40, VidOR, and VidVRd in 5-shot set-up. Our Object-aware Video OWL-ViT consistently surpasses both state-of-the-art and baseline methods across all datasets and metrics.}

\label{tab:main}
\end{table*}

\subsection{Results and Discussion}
\subsection{Main Results}
Our proposed approach achieves state-of-the-art performance across all four benchmark datasets. On FSVOD-500, we achieve AP of 30.6\%, AP${50}$ of 42.9\%, and AP${75}$ of 32.1\%, surpassing TACF by margins of 3.7\%, 3.7\%, and 4.2\%, highlighting robustness in localization and classification. Similarly, on FSYTV-40, our method yields 21.2\% AP, 29.8\% AP${50}$, and 23.5\% AP${75}$, outperforming TACF by 5.3\%, 5.6\%, and 5.6\%. Performance gains are more pronounced on VidOR and VidVRD, 49.4\% AP, 57.3\% AP${50}$, and 50.2\% AP${75}$ for VidOR, and 48.7\% AP, 57.8\% AP${50}$, and 49.2\% AP${75}$ for VidVRD, setting new benchmarks. These results underscore the effectiveness of our object-aware temporal modelling in capturing complex spatiotemporal dependencies with minimal supervision.

\begin{table*}[!th]
\scriptsize
\centering
 \begin{tabular}{llcccccccccccc}
\toprule
&\multirow{2}{*}{Method}& \multicolumn{3}{c}{FSVOD-500} & \multicolumn{3}{c}{FSYTV-40} & \multicolumn{3}{c}{VidOR} & \multicolumn{3}{c}{VidVRD}\\
\cmidrule(lr){3-5} \cmidrule(lr){6-8} \cmidrule(lr){9-11} \cmidrule(lr){12-14}
       && $AP$ & $AP_{50}$ & $AP_{75}$ & $AP$ & $AP_{50}$ & $AP_{75}$ & $AP$ & $AP_{50}$ & $AP_{75}$ & $AP$ & $AP_{50}$ & $AP_{75}$  \\
\midrule
\multirow{7}{*}{\rotatebox{90}{\textbf{1-Shot}}}
&CoAE &17.2&27.8&17.2&6.7&17.3&9.8&33.2&42.5&35.2&29.6&37.6&31.5\\
&UP-DETR &17.3&24.3&18.4&9.7&16.9&9.9&35.6&45.2&37.2&8.5&15.2&9.3\\
&DeepSort+YOLOv8 + CLIP &12.2&20.1&12.8&4.1&7.9&4.7&21.4&28.4&22.7&21.9&31.7&22.8\\
&FSVOD    &20.7 &29.4 &21.4 &11.4 & 18.6 & 12.1& 36.8 & 45.3&36.3 &34.5 &45.2  &35.2       \\
&QDETRv   & \underline{25.2} & \underline{34.9}  & \underline{25.8} & \underline{13.6} & \underline{24.1}  & \underline{14.2} & \underline{41.8} & 50.3 & \underline{42.2} & \underline{40.3} & \underline{49.7} & 40.1   \\
&Video OWL-ViT & 23.8 & 33.8 & 24.7 & 12.4 & 22.7& 13.2& 40.3& \underline{51.7} & 41.5 & 39.8 & 47.6& \underline{40.2}\\
&Ours &\textbf{27.4} & \textbf{39.1}& \textbf{28.9}& \textbf{18.3}& \textbf{26.2}&\textbf{19.7} &\textbf{45.2} &\textbf{54.1} &\textbf{47.6} &\textbf{45.3} &\textbf{54.5} & \textbf{46.8} \\
\midrule
\multirow{7}{*}{\rotatebox{90}{\textbf{3-Shot}}}
&CoAE &17.8&28.2&17.6&6.9&17.5&9.8&33.3&42.6&35.6&29.9&37.8&31.9\\
&UP-DETR &17.9&25.1&18.8&10.2&17.5&10.3&36.2&46.8&38.3&9.3&16.1&10.4\\
&DeepSort+YOLOv8 + CLIP  &12.7&20.4&13.7&4.3&8.3&4.9&22.0&28.8&23.1&22.0&31.9&23.4\\
&FSVOD   &21.9 &30.8 &22.1 &12.3 & 19.2 & 12.9& 37.6 & 46.6 &37.8 &35.4 &46.1  &36.4       \\
&QDETRv  & \underline{25.6} & \underline{35.2}  & \underline{26.3}  & \underline{13.8} & \underline{24.5}  & \underline{14.5} & \underline{42.1} & 50.8 & \underline{42.7} & \underline{40.8} & \underline{50.3} & 40.6  \\
& Video OWL-ViT & 24.1 & 34.6& 25.2 & 12.7 & 23.2  & 13.8 & 40.6& \underline{52.5} &41.8 &40.4 & 48.3 & \underline{40.7}    \\
& Ours &\textbf{28.2} & \textbf{40.5}& \textbf{31.1}& \textbf{19.6}& \textbf{27.2}&\textbf{20.9} &\textbf{46.4} &\textbf{55.6} &\textbf{48.2} &\textbf{46.5} &\textbf{55.7} & \textbf{47.2} \\
\midrule
\multirow{7}{*}{\rotatebox{90}{\textbf{10-Shot}}}
&CoAE &20.1&29.9&19.7&10.2&19.8&12.3&37.2&44.9&38.2&32.3&39.7&32.7\\
&UP-DETR &23.2&30.2&23.7&13.7&22.3&15.2&41.8&50.2&43.1&13.7&21.2&15.4\\
&DeepSort+YOLOv8 + CLIP &16.6&23.7&17.2&7.8&11.5&9.4&26.3&32.9&25.8&25.1&35.6&26.6\\
&FSVOD &  \underline{27.2} & \underline{38.5} & \underline{28.6} & \underline{16.1} & \underline{25.8} & \underline{17.9}& \underline{47.3}& \underline{56.7}& \underline{48.4}& \underline{44.7}& \underline{54.9}& \underline{44.9}\\
&QDETRv & 26.2 & 33.8 & 23.8 & 14.2 & 23.1 & 15.6 &43.9&52.1&45.2&43.1&53.8&44.2   \\
&Video OWL-ViT & 26.7 & 35.8 & 27.1 & 14.2 & 25.1& 14.9 & 43.7 & 54.2& 44.1 & 43.7 & 52.8 & 44.3   \\
& Ours & \textbf{33.2} & \textbf{45.2} & \textbf{34.9} & \textbf{23.5} & \textbf{32.2} & \textbf{25.2} &\textbf{51.8}&\textbf{59.6}&\textbf{53.3}&\textbf{50.3}&\textbf{59.2}&\textbf{51.7}  \\
\bottomrule
\end{tabular}
\caption[Few-shot performance comparison]{Few-shot performance comparison. Our method consistently outperforms SoTA and implemented baselines across all settings and datasets, demonstrating its effectiveness across diverse settings and datasets.}
\label{tab:aaai26_few-shot}
\end{table*}

\begin{table}[!th]
\scriptsize
    \centering
\begin{tabular}{c c c c c c c}
\toprule
 \multicolumn{1}{l}{}  & \multicolumn{3}{c}{FSVOD-500}  & \multicolumn{3}{c}{FSYTV-40}\\
 \cmidrule(r){2-4} \cmidrule(r){5-7}
Temopral Fusion & $AP$ & $AP_{50}$ & $AP_{75}$ & $AP$ & $AP_{50}$ & $AP_{75}$\\
 \midrule
  \xmark  & 25.4 & 37.1 & 26.9 & 19.2 & 26.8& 21.4\\
    \cmark & \textbf{30.6} & \textbf{42.9} & \textbf{32.1} & \textbf{21.2} & \textbf{29.8} & \textbf{23.5} \\
 \bottomrule
\end{tabular}
\caption[Advantage of Object Aware Temporal-fusion]{Advantage of Object Aware Temporal-fusion. Object-Aware Temporal Fusion improves detection performance across all metrics on FSVOD-500 and FSYTV-40 datasets. Removing the Temporal fusion (\xmark) significantly reduces the performance shows the efficacy of object aware decoder.
\label{tab:obj-aware_decoder}}
\end{table}

\begin{figure}[!th]
    \centering
    \includegraphics[width=0.7\linewidth]{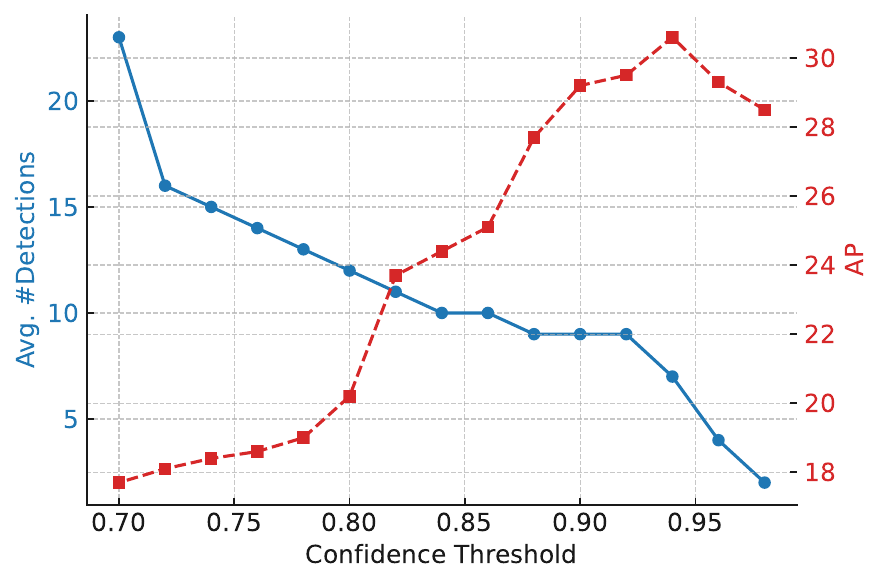}
    \caption[Threshold effect]{\label{fig:thr_oadec} Threshold effect ($c^p > \tau$) during the object-aware temporal consistency. }
\end{figure}

\subsection{Temporal vs. Non-temporal Methods}
The performance comparison clearly demonstrates the advantage of temporal methods over non-temporal. While non-temporal methods like UP-DETR and FSOD show competitive performance, they fundamentally fail to capture temporal dependencies crucial for video object detection. This is evidenced by the consistent performance gap between across all datasets. For instance, on FSVOD-500, even the best non-temporal method (UP-DETR with 20.1\% AP) falls significantly short compared to our approach (30.6\% AP), representing a substantial 10.5\% absolute improvement. Similarly, on FSYTV-40, our method outperforms UP-DETR by 9.4\%. This highlight the critical importance of modeling temporal information in video object detection, particularly in the few-shot set-up where limited examples make temporal continuity an essential cue for accurate detection.

\subsection{Comparison with adapted Temporal Methods}
Despite leveraging strong trackers (ByteTrack, DeepSort, TrackFormer) with CLIP's rich semantic representations, these methods consistently underperform compared to dedicated video object detection architectures. The performance gap is substantial across all datasets: on FSVOD-500, the best tracker-based approach (ByteTrack+CLIP) achieves only 14.7\% AP, less than half of our method's 30.6\% AP. Similar patterns emerge on other datasets, with gaps of 13.3\% AP on FSYTV-40, 24.5\% AP on VidOR, and 23.1\% AP on VidVRD. This significant under performance reveals that tracking methods struggle with few-shot detection when adopted without explicit few-shot modules.

\subsection{Improvements over Specialized Few-Shot Methods:}
Our approach demonstrates significant improvements over previous specialized SoTA few-shot video detection methods, which represent the most direct competitors to our work. Compared to FSVOD, which pioneered dedicated architectures for few-shot video object detection, we achieve consistent gains across all datasets: 5.5\% AP on FSVOD-500, 6.6\% AP on FSYTV-40, 4.3\% AP on VidOR, and 8.0\% AP on VidVRD. Similar, observation we got for specialized few-shot method, TACF. Similarly, our approach significantly outperformed the adapted Video OWL-ViT method. The consistent performance gains over these specialized methods highlight several key advantages of our approach: (1) more effective temporal aggregation through our object-aware frame encoder, (2) better feature matching between support and query examples using few-shot designed heads.

\begin{figure*}[t!]
    \centering
      \includegraphics[width=0.95\textwidth]{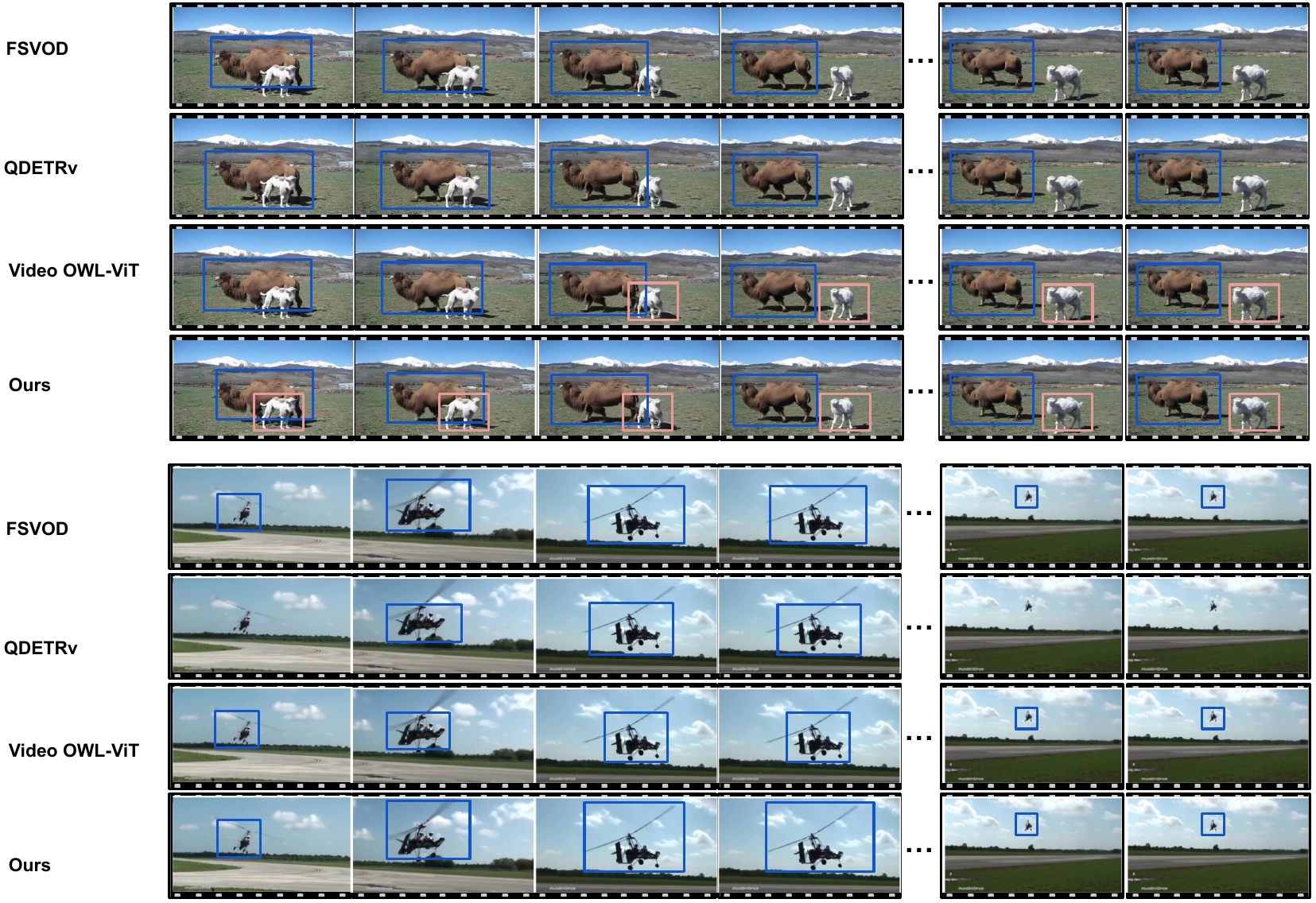}
  \caption[Qualitative Results]{\label{fig:aaai26_qual}. Qualitative comparison of detection results across SoTA methods for two test videos from FSVOD-500. Our approach accurately detects multiple instances of the \emph{Bactrian} camel (top) having a different visual appearance and precisely localizes the \emph{Autogyro} (bottom), even as it moves farther from the camera in the subsequent frames. Notably, competing methods miss some camel instances, while Video OWL-ViT fails to detect all instances in the first two frames.}
\end{figure*}

\subsection{Advantage of Temporal Fusion (Object Aware Decoder)}
In Table~\ref{tab:obj-aware_decoder} on FSVOD-500, AP rises from 25.4\% to 30.6\% (+5.2\%), AP$_{50}$ from 37.1\% to 42.9\% (+5.8\%), and AP$_{75}$ from 26.9\% to 32.1\% (5.2\%) when temporal fusion used. On FSYTV-40, AP improves from 19.2\% to 21.2\% (+2.0\%), AP$_{50}$ from 26.8\% to 29.8\% (+3.0\%), and AP$_{75}$ from 21.4\% to 23.5\% (+2.1\%). These gains highlight the decoder’s enhanced temporal reasoning detection performance.

\subsection{Effect of Threshold ($\tau$) on Temporal propagation}
Figure~\ref{fig:thr_oadec} demonstrates the relationship between confidence threshold and detection performance. As the threshold decreases from 0.98 to 0.70, the average number of detections increases from 2 to 23, indicating more objects are detected at lower confidence levels. The AP follows a non-linear pattern, initially improving as the threshold decreases, reaching a peak of 30.6 at a threshold of 0.94, before gradually declining to 17.7 at 0.70. This pattern suggests that while lower thresholds increase detection quantity, they also introduce false positives that reduce precision. The optimal threshold (0.94) represents the balance point between detection coverage and precision, with performance degrading when the threshold is excessively relaxed. Since the optimum is a property of the confidence distribution of the vision--language scores, and that distribution shifts with object size, scene clutter and category granularity, the best value of $\tau$ is dataset-dependent to some degree. In practice the same value transfers across our four benchmarks without per-dataset retuning, which suggests that the sensitivity is mild for datasets of broadly similar composition, though a domain differing substantially in these respects would warrant repeating this ablation. We did not explore an automatic mechanism for setting $\tau$. The most natural option would replace the absolute threshold with a per-frame quantile of the score distribution, propagating the top-$q\%$ of detections, which self-calibrates when the score distribution shifts and moves the tuning burden to $q$, a considerably less sensitive quantity. A small learned gate over the detection embedding and its score, trained jointly with the detection loss, would remove the hyper-parameter altogether at the cost of a more complex propagation module.
\begin{figure}[!th]
    \centering
    \includegraphics[width=0.70\linewidth]{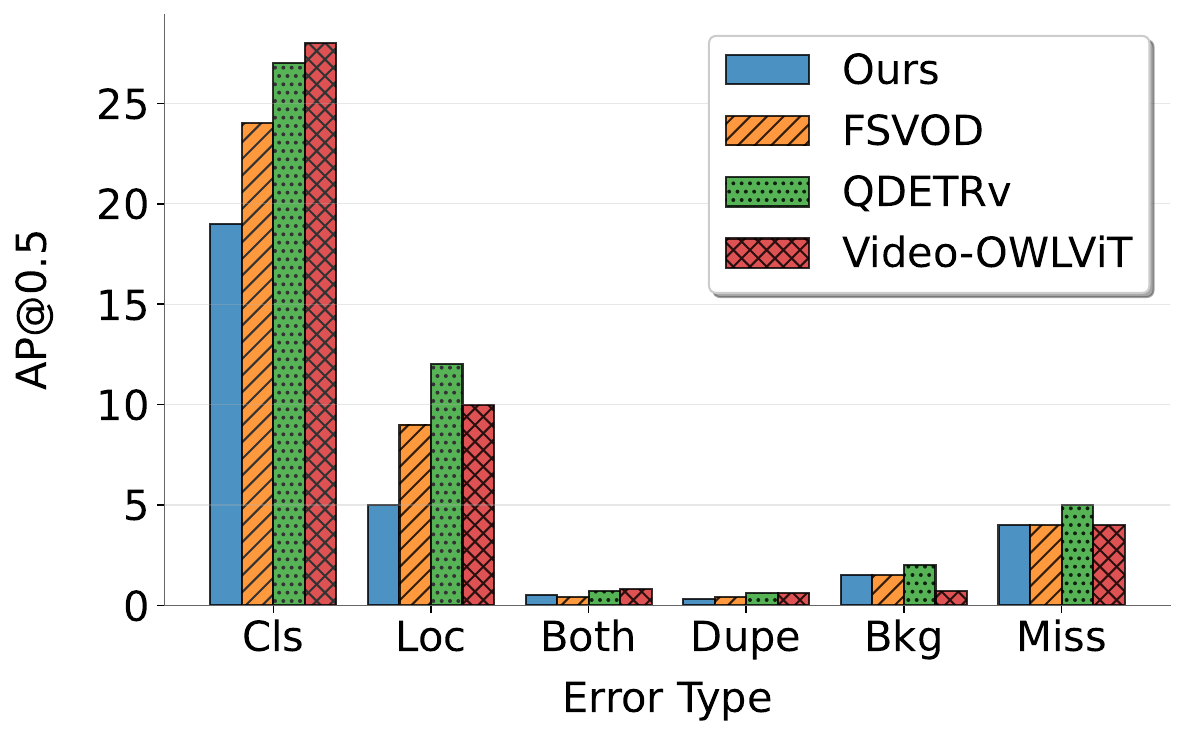}
    \caption[Error Type Analysis]{\label{fig:error_type} Error type analysis on FSVOD val set. A lower AP score is better. }
\end{figure}

\subsection{Few-shot Performance Comparison} Table~\ref{tab:aaai26_few-shot} shows consistent improvement of our method across baselines in 1-shot, 3-shot, and 10-shot settings. In the 1-shot case, we achieve 27.4\% AP on FSVOD-500 and 18.3\% on FSYTV-40, outperforming FSVOD by 6.7\% and 6.9\%, respectively. In the 10-shot setup, our method attains 33.2\% AP on FSVOD-500 and 23.5\% on FSYTV-40, with gains of 6.0\% and 7.4\% over FSVOD. Similar improvements across other benchmarks further validate our approach.

\subsection{Error Type Analysis}
Following TIDE~\cite{tide-eccv2020}, we analyze error types in Fig.~\ref {fig:error_type}. Classification errors dominate all approaches (19-28 AP@0.5), with our method achieving lowest classification (19) and localization (5) errors. Video-OWLViT excels in background handling (0.7) but suffers from high classification errors (28), while QDETRv performs the poorest overall. Improving feature discrimination and localization precision are key challenges. \\

\subsection{Qualitative Results} Figure~\ref{fig:aaai26_qual} shows qualitative results on two videos from FSVOD-500. Our method demonstrates several advantages. In the first video, our approach successfully detects multiple instances of the same object category (bactrian camel), while competing methods miss some instances. Video OWL-ViT fails to detect all instances in the first two frames, highlighting the effectiveness of our object-aware decoder. In the second video featuring an autogiro, most methods can detect the object, but our approach provides more accurate bounding boxes. QDETRv particularly struggles when the object becomes smaller in the frame.

\subsection{Efficiency Analysis.}
Our method achieves a strong balance between accuracy and efficiency, processing at 14 FPS with 8.1 GB of memory while attaining 30.6 AP. This is 40\% faster than VideoOWL-ViT (10 FPS, 8.4 GB) and 27\% faster than QDETRv (11 FPS, 7.6 GB), while providing substantially higher accuracy (+4.8 AP and +4.5 AP, respectively). Although FSVOD operates at 16 FPS with lower memory usage (5.8 GB), our approach delivers a +5.5\% AP improvement with only a minor trade-off. Compared to other baselines, DeepSort runs at 9 FPS (7.9 GB), CenterTrack at 12 FPS (6.2 GB), and UP-DETR at 13 FPS (7.4 GB).

Table~\ref{tab:aaai26_efficiency} collects these measurements. The comparison against QDETRv is worth stating explicitly, since our framework builds on a heavier vision--language backbone and might be expected to pay for it in latency. It does not: our method is 27\% faster than QDETRv while gaining 4.5 AP, for only 0.5\,GB of additional memory. The reason is architectural. We dispense with explicit object-tube proposals, which are the expensive component in tube-based methods, and the filtering mechanism propagates only high-confidence object features, so the number of tokens carried across frames remains small. The one genuine trade-off is against FSVOD, which is faster (16 FPS) and leaner (5.8\,GB) at 5.5 AP lower accuracy; for a memory-constrained edge deployment where the difference between 5.8\,GB and 8.1\,GB is the binding constraint, FSVOD remains defensible, whereas with an 8\,GB budget available our method dominates on both axes.

\begin{table}[!t]
\scriptsize
\centering
\begin{tabular}{lccc}
\toprule
\textbf{Method} & \textbf{FPS} & \textbf{Memory (GB)} & \textbf{AP (FSVOD-500)} \\
\midrule
\textbf{Ours} & \textbf{14} & 8.1 & \textbf{30.6} \\
QDETRv & 11 & 7.6 & 26.1 \\
Video OWL-ViT & 10 & 8.4 & 25.8 \\
FSVOD & 16 & 5.8 & 25.1 \\
UP-DETR & 13 & 7.4 & 20.1 \\
CenterTrack & 12 & 6.2 & 20.6 \\
DeepSort+YOLOv8+CLIP & 9 & 7.9 & 13.2 \\
\bottomrule
\end{tabular}
\caption[Inference cost comparison]{Inference cost and accuracy on FSVOD-500. Our method is both faster and more accurate than QDETRv and Video OWL-ViT; the only trade-off is against FSVOD, which is leaner but substantially less accurate.}
\label{tab:aaai26_efficiency}
\end{table}

\subsection{Comparison against a Generic Video-LLM}
To test the claim that specialised grounding still outperforms generic vision--language models on this task, we additionally evaluated Qwen3.5-VL~\cite{qwen} under the same protocol as our method, supplying the support image as the visual exemplar and scoring the predicted boxes with the same AP measure. The comparison is therefore like-for-like in the information available to each model. Qwen3.5-VL~\cite{qwen} reaches 15.6 AP on FSVOD-500 and 8.3 AP on FSYTV-40, well below our 30.6 and 21.2 respectively. We would not attribute this gap to model capacity. Qwen3.5-VL~\cite{qwen} matches an exemplar within a frame but has no mechanism for maintaining object identity across frames, so its detections are temporally inconsistent, which is precisely what the feature-propagation component supplies.

\section{Summary}
In this chapter, we introduced a novel approach for Few-Shot Video Object Detection that incorporates a language-aligned vision encoder and a filtering mechanism to selectively propagate high-confidence object features across frames. This mechanism efficiently enhances feature progression, reduces noise accumulation, and improves detection accuracy in a few-shot video setting. Through extensive experiments on four datasets, our approach consistently outperforms state-of-the-art methods, achieving superior detection performance and demonstrating its effectiveness in real-world video object detection tasks.

In the next chapter, we shift from grounding individual objects to grounding the interactions between them, that is, understanding how two objects relate and behave together.

\chapter{Few-Shot Relationships Grounding in Videos}
\label{chapthree}

Interpreting visual relationships is a core aspect of comprehensive video understanding. Given a query visual relationship as $<$subject, predicate, object$>$ and a test video, our objective is to localize the subject and object that are connected via the predicate. Given modern visio-lingual understanding capabilities, solving this problem is achievable, provided that there are large-scale annotated training examples available. However, annotating for every combination of subject, object, and predicate is cumbersome, expensive, and possibly infeasible. Therefore, there is a need for models that can learn to spatially and temporally localize subjects and objects that are connected via an unseen predicate using only a few support set videos sharing the common predicate. We address this challenging problem, referred to as few-shot referring relationships in videos for the first time. To this end, we pose the problem as a minimization of an objective function defined over a T-partite random field. Here, the vertices of the random field correspond to candidate bounding boxes for the subject and object, and T represents the number of frames in the test video. This objective function is composed of frame-level and visual relationship similarity potentials. To learn these potentials, we use a relation network that takes query-conditioned translational relationship embedding as inputs and is meta-trained using support set videos in an episodic manner. Further, the objective function is minimized using a belief propagation-based message passing on the random field to obtain the spatiotemporal localization or subject and object trajectories. We perform extensive experiments using two public benchmarks, namely ImageNet-VidVRD and VidOR, and compare the proposed approach with competitive baselines to assess its efficacy.

\begin{center}
    \centering
    \captionsetup{type=figure}
    \includegraphics[width=0.96\textwidth]{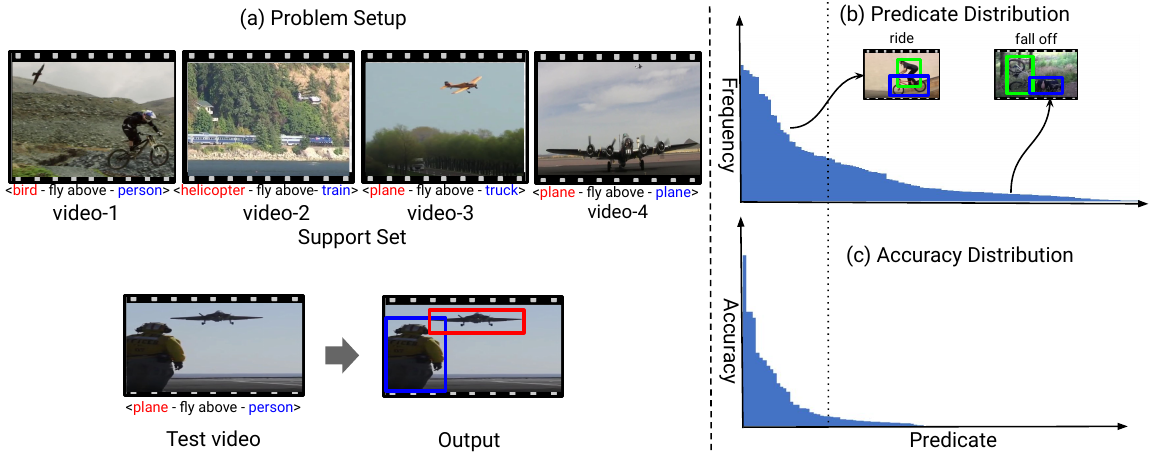}
       \captionof{figure}[The proposed problem setup]{\label{fig:cvpr23goal}\textbf {The proposed problem setup}.
    \textbf{(a).} Given a query visual relationship as $<$subject, predicate, object$>$ and a test video, our goal is to localize the subject and object on the test video with the help of a support set containing a few videos with the same predicate. In this example, the goal is to spatio-temporally localize the \emph{plane} (subject) and \emph{person} (object) that are connected via \emph{fly above} (predicate), with the help of a support set containing only four videos sharing predicate \emph{fly above}. It should be noted here that \emph{fly above} is unseen during training. We refer this problem as \emph{few-shot referring relationship in videos}. This problem setup is inspired by the real-world scenario where obtaining large-scale annotations for every visual relationship is practically infeasible. As shown in (b), a popular visual relationship video dataset, namely ImageNet-VidVRD, contains many predicates with very few examples, i.e., it has a long-tail distribution. Further, as shown in (c), the success of a modern visual relationship localization technique (vRGV)~\cite{xiao2020visual} is clearly proportional to predicate distribution in the training set. This calls a need for solving referring relationship tasks in a few-shot setup. We propose this task and present a novel principled solution.
}
\end{center}

The subsequent sections present the introduction, a review of related work, the proposed methodology, and a detailed analysis of results and discussions.

\section{\label{sec:cvpr_intro}Introduction}
Consider the following problem: given a video, a visual relationship query represented as a $<$\textit{subject, predicate, object}$>$ tuple, and a support set of a few videos containing the same predicate but not necessarily the same subjects and objects, our objective is to spatially and temporally localize both subjects and objects that are related via the predicate within the video. We refer to this problem as \emph{Few-shot Referring Relationship} and illustrate it in Figure~\ref{fig:cvpr23goal}. Solving this problem has the potential to benefit cross-task video understanding~\cite{zhukov2019cross} and video retrieval~\cite{dong2019dual,gabeur2020multi}, among other applications. Identifying its utility, referring relationship task for images has been first introduced by~\cite{krishna2018referring}. However, referring relationships in videos poses additional video-specific challenges, such as understanding dynamic visual relationships. Some of these challenges have been addressed in recent research by Xiao et al.~\cite{xiao2020visual}, but with a reliance on strong supervision.
Referring relationships in videos within a few-shot setup is an under-explored area. We aim to fill this research gap via our work.

Visual relationships inherently have long-tail distributions in any video collection. For example, Image-Net VidVRD~\cite{vidvrd} dataset includes approximately $18.9\%$ predicates with more than 100 instances but $20.5\%$ predicates with less than 10 instances. This phenomenon is also shown in Figure~\ref{fig:cvpr23goal}, where most predicates belong to the tail side of the distribution. The methods that work best for frequent visual relationships do not necessarily generalize well to unseen visual relationships. Moreover, in a real-world scenario annotating visual relationships for each combination of subject, object, and predicate are cumbersome, expensive, and possibly infeasible. Therefore, there is a need to study visual relationship tasks in a few-shot setup. For instance: only with a few examples of the \emph{fly above} predicate, such as videos containing ${<}$\emph{bird, fly above, person}${>}$, ${<}$\emph{helicopter, fly above, train}${>}$ as shown in Figure~\ref{fig:cvpr23goal} (a), a model should be able to generalize to the unseen visual relationship, such as ${<}$\emph{plane, fly above, person}${>}$. We propose a solution for \emph{Few-shot Referring Relationship} in videos in this work.

We pose the problem of a few-shot referring relationship in the video as a minimization of an objective function defined over a $T$-partite random field where $T$ is the number of frames in the test video. Furthermore, the vertices of the random field are treated as random variables and represent candidate bounding boxes for the subject and objects. The objective function consists of frame-level potentials and visual relationship similarity potentials, both of which are learned using a relation network that takes query-conditioned translational relationship embeddings as inputs. We meta-train the relation network using support set videos in an episodic manner. Further, the objective function is minimized using a belief propagation-based message passing on the random field to obtain subject and object trajectories. We perform extensive experiments on two public benchmarks, namely ImageNet-VidVRD~\cite{vidvrd} and VidOR~\cite{thomee2016yfcc100m}, and report the accuracy of localizing subject, object, and relation, denoted by $A^{sub}$, $A^{obj}$, and $A^r$, respectively, along with other popular measures used in the literature. Our proposed approach clearly outperforms the related baselines.

The contributions of this chapter are three folds. (i) We propose a novel problem setup for referring relationship in videos, where the model must learn to localize the subject and object corresponding to a query visual relationship that was unseen during training using only a few support videos. (ii) We propose a new formulation to solve this task based on the minimization of an objective function on $T$-partite random field where $T$ is the number of frames in the test video, and the vertices of the random field representing potential bounding boxes for subject and objects correspond to the random variables. (Section~\ref{sec:tpartite}). (iii) Additionally, to enrich query-conditioned relational embeddings, we present two aggregation techniques, namely global semantic and local localization aggregations. The use of these aggregation techniques results in enhanced relationship representations, which helps to obtain better trajectories for objects and subjects related via the query visual relationship. This is evidenced by extensive experiments and ablations. (Sections~\ref{sec:re} and~\ref{sec:rd}).

\section{Related Work}
\noindent\textbf{Visual Relationships:}
Interpreting visual relationships in images~\cite{lyu2022fine,liang2018visual,zhan2019exploring} as well as videos~\cite{zheng2022vrdformer,li2022dynamic,cherian20222,gao2021video,li2021interventional} have gained huge attention over the last few years. They have also been key components of large-scale popular datasets such as Visual Genome~\cite{krishna2017visual} and Action Genome~\cite{ji2020action}. Visual relationships were studied with respect to object segmentation to leverage spatial relations~\cite{galleguillos2008object} and to understand human-object interactions~\cite{yao2010modeling} via human-centric relationships. Krishna et al.~\cite{krishna2017visual} proposed the concept of scene graphs by combining multiple visual relationships in a graph structure. The structured representation of the scene graph is exploited for several tasks, including image retrieval~\cite{johnson2015image}. Shang et al.~\cite{vidvrd} extended scene graphs from images to videos. Videos have spatiotemporal nature introducing dynamic relations that are not present in images. Scene graphs in videos represent fine-grain information that helps in the downstream task for spatiotemporal reasoning. Several works have been introduced that utilize visual relationships to improve downstream tasks such as an image or video captioning~\cite{yao2018exploring}, visual grounding~\cite{hu2017modeling}, visual question answering~\cite{lu2018r}. To localize subjects and objects connected via a query visual relationship, Krishna et al.~\cite{krishna2018referring} proposed referring relationship for images. Their proposed method used an iterative message-passing mechanism between subject and object using language priors to ground the query relations. Further, Xiao et al.~\cite{xiao2020visual} extended referring relationship to videos and refer it as visual relationship grounding in videos. In a similar spirit, but, in a \emph{few-shot set-up}, we propose to spatiotemporally localize subjects and objects in a video for a given relationship.

\noindent\textbf{Few-Shot Learning:}
Few-shot learning in the literature can be grouped into: (i) metric-based ~\cite{koch2015siamese,vinyals2016matching,snell2017prototypical} and (ii) model-based~\cite{finn2017model,ravi2016optimization,sung2018learning} methods. We limit our discussion to only closely related works. Vinyals et al.~\cite{vinyals2016matching} proposed matching networks that learn to compare using a small support set. Further, Sung et al.~\cite{sung2018learning} proposed a relation network that learns from a few labeled images of the support set by comparing the query image. In this work, we have used a relation network to learn relationship similarity using the given support set videos, and this enabled us to obtain both frame-level and visual relationship similarity potentials.
\begin{figure}[t!]
    \centering
      \includegraphics[width=0.7\textwidth]{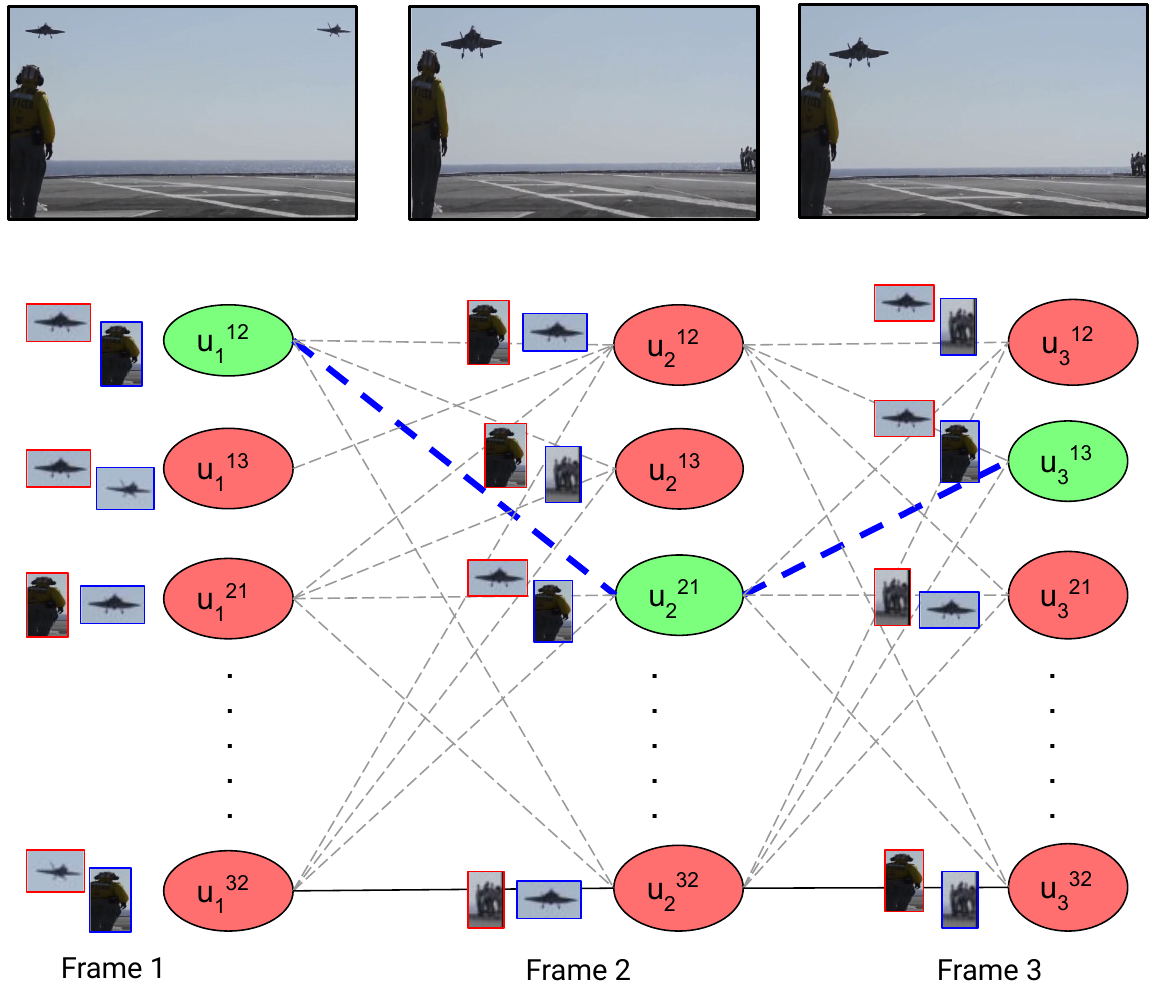}
    \caption[Illustration of $T$-partite random field]{\label{fig:graph_construction} \textbf{Illustration of $T$-partite random field.} Here, we illustrate graph construction using three frames ($T=3$) and three object bounding boxes ($M=3$). The \emph{ordered pair} of bounding boxes are represented as a node. For this example, each frame will result in $2 \times {3 \choose 2}=6$ nodes. Each node corresponds to a binary random variable ($u_{jk}^{i}$) that takes one of the two labels, i.e. \{select ($1$), reject ($0$)\}. The green- and red-filled nodes correspond to those random variables that the optimization framework aims to assign $1$ and $0$, respectively, and the dotted blue line indicates the ground truth tracklet. \textbf{[Best viewed in color]}. }
\end{figure}

\section{Proposed Method}
\noindent\textbf{Task Definition:} Given an unseen visual relationship consisting of subject ($s$), predicate ($p$), and object ($o$), i.e. $r = {<}s,~p,~o{>}$ as a query and a test video $v$ along with a small support set of $K$ videos $S_{sup} = \{{<}s_i,~p,o_i{>},~v_i\}_{i=1}^{K}$ containing the same predicate $p$, the goal is to obtain the sequence of bounding boxes (also known as trajectories) $T^{*}_s$ and $T^{*}_o$ corresponding to subject and object, respectively on the test video $v$.
We refer to this task as \emph{few-shot referring relationship} in videos.
As an example, in Figure~\ref{fig:cvpr23goal} (a), ${r=}<$\emph{plane}, \emph{fly above}, \emph{person}$>$ is a query that needs to be spatiotemporally localized on a test video using a support set that contains four videos of \emph{fly above} predicate. We pose this task as an optimization problem on a $T$-partite random field which we describe next.
\subsection{\label{sec:tpartite}Few-shot Referring Relationship as an optimization on a $T$-partite random field}
Given a test video $v$, we split it into $T$ frames. Then, we obtain $M$ most confident object bounding boxes on each frame using FasterRCNN~\cite{FasterRCNN}. Video $v$ can be represented as a sequence of extracted bounding boxes, $v = \{B_{i}^{j} | i\in [1, T], j\in [1, M] \}$. The pair of these bounding boxes in each frame is a candidate solution for referring relationship task. While finding the optimal solution using a brute force technique is combinatorial
and practically infeasible, we solve it using optimization on a $T$-partite random field. To this end, we construct a $T$-partite graph $G=({\cal V}, {\cal E})$ representing the test video as follows: for a frame $i$, we represent each pair of $M$ bounding boxes as nodes and their all possible next and previous-frame connections as edges. More precisely, the set of vertices ${\cal V}$ for a video $v$ contains ordered pair of bounding boxes as follows: $\{u_{i}^{jk}=(B_{i}^{j},B_{i}^{k}) : i, \in [1,T]; j,k \in [1,M]; j \neq k\}$, and the set of edges is defined as ${\cal E} = \{(u_{i}^{j,k}, u_{(i+1)}^{j,k}): i \in [1,T-1];j,k \in [1,M]; j \neq k\}$. Each vertex $u_{i}^{jk}$ of this graph is a binary random variable that takes one of two labels $\{\emph{select}~(1), \emph{reject}~(0)\}$ for the given query $r$. Figure~\ref{fig:graph_construction} illustrates the construction of graph $G$ for a three-frame test video. It is worth making the cost of this construction explicit, since the term ``$T$-partite random field'' may suggest a more expensive inference procedure than is actually required. Each frame contributes $M(M-1)$ binary nodes, and the edge set defined above connects only the \emph{same} ordered pair $(j,k)$ across consecutive frames. The graph therefore decomposes into $M(M-1)$ independent chains of length $T$. Belief propagation on a chain is exact and linear in its length, so inference costs $O(T \cdot M^{2})$ over binary variables and is trivially parallel across the chains. The cost is thus linear in the number of frames, which means that long videos are not the binding constraint; the quadratic term is $M$, so doubling the number of candidate boxes per frame quadruples the node count, and densely annotated settings are the harder case. Where a video is long enough for memory to become a concern, the natural mitigations are sampling keyframes, since a relationship persists across adjacent frames and not every frame need form a graph layer, and capping $M$ through proposal suppression. It should be noted here that the selected nodes (corresponding subject and object bounding box) from each frame form subject and object trajectories. To obtain optimal subject and object trajectories ($T^{*}_s$ and $T^{*}_o$) for a given visual relationship $r$, we solve following optimization problem:

\begin{equation}
\label{eq:obj}
\begin{split}
 T^{*}_s, T^{*}_o = \arg\min_{{\bm{\theta}},i,j,k} \sum_{i=1}^{T}\sum_{\substack{j,k=1\\j\neq k}}^{M}\Big(\Psi_i(u_{i}^{jk},r, {\bm{\theta}})\\+\sum_{l \in n(i)}\Psi_{il} (u_{i}^{jk}, u_{l}^{jk}, r, {\bm{\theta}})
  \Big).
 \end{split}
\end{equation}

Here, $n(i)$ represents the neighboring nodes to frame $i$, and ${\bm{\theta}}$ is a learnable parameter of a neural network that needs to be trained using support set videos. Specifically, we use support set videos to meta-train relation network ($R_{\bm{\theta}}$) in a few-shot way. We describe the relation network used in our framework and its training strategy in detail in Section~\ref{sec:rel_net}. Further, in the aforementioned objective function, the terms $\Psi_i$ and $\Psi_{il}$ denote frame-level and visual relationship similarity potentials that are defined next.

\subsubsection{Frame-level Potentials $(\Psi_{i})$}
We compute frame-level potentials  $\Psi_i(u_{i}^{jk},r,{\bm{\theta}})$ such that the cost of selecting a node to form a trajectory is low if the selected node is semantically similar to query predicate $r$, otherwise high. Mathematically,
\begin{equation}
\begin{split}
        \Psi_i(u_{i}^{jk},r,{\bm{\theta}}) = \frac{-\sum_{m =1}^{K}R_{\bm{\theta}} \Big(f_{r}({u_{i}^{jk}}), f_r({u_{i_m}^{jk}})\Big)}{K},
        \end{split}
\end{equation}
where, $u_{i_m}^{jk}$ is the relationship pair in the $m^{th}$ support set video connected via query predicate $r$ and $f_{r}({u_{i}^{jk}})$ is a visual relationship embedding in video $v$ for $j^{th}$ and $k^{th}$ bounding boxes of $i^{th}$ frame. Thus, $\Psi_i(u_{i}^{jk},r, {\bm{\theta}})$ gives the negative of average similarity of node $u_{i}^{jk}$ with respect to the relationship of $S_{sup}$ having predicate as $r$. The $R_{\bm{\theta}}$ returns a value closer to $1$ when it represents visually and semantically similar visual relationships, and closer to $0$, otherwise.

\subsubsection{Visual Relationship Similarity Potentials $(\Psi_{il})$}
To ensure that coherent (both semantically and visually) visual relationships are being selected across frames, we define visual relationship similarity potential such that the cost of selecting a similar visual relationship in the optimal tracklet is low. To this end, the visual relationship similarity potential
$\Psi_{il} (u_{i}^{jk}, u_{l}^{jk},{\bm{\theta}})$ is computed as follows:
\begin{equation}
    \label{p_cost}
    \Psi_{il} (u_{i}^{jk}, u_{l}^{jk}, {\bm{\theta}}) = -R_{\bm{\theta}} \Big(f_{r}({u_{i}^{jk}}),f_{r}({u_{l}^{jk}})\Big),
\end{equation}

where $f_{r}({u_{i}^{jk}})$ is a visual relationship embedding in the video $v$ for $j^{th}$ and $k^{th}$ bounding boxes of $i^{th}$ frame.
The computation of these relationship embeddings and details of the relation network follows next.

\subsection{\label{sec:re}Query-conditioned Relationship Embedding}
We describe the representation of a frame-level visual relationship in this section. To this end, we first obtain $M$ extracted objects for each frame of the video using Faster-RCNN~\cite{FasterRCNN}. Then, given a query relationship $r={<}s,p,o{>}$, for each frame, an object or subject representation corresponding to $j^{th}$ and $k^{th}$ bounding box of $i^{th}$ frame respectively are obtained by the following equations:
\begin{equation}
\label{eq:initial_f_s}
f_s(B_j^{i}) = \textbf{W}_{s_2}\Big(ReLU\Big(\textbf{W}_{s_1}[f_{app}^s(B_j^{i});f_{spa}^s(B_j^{i}); G(s)]\Big)\Big),
\end{equation}

\begin{equation}
\label{eq:initial_f_o}
f_o(B_j^{i}) = \textbf{W}_{o_2}\Big(ReLU\Big(\textbf{W}_{o_1}[f_{app}^o(B_j^{i}); f_{spa}^o(B_j^{i}); G(o)]\Big)\Big).
\end{equation}

Here, $\textbf{W}_{s_1}$, $\textbf{W}_{s_2}$, $\textbf{W}_{o_1}$ and $\textbf{W}_{o_2}$ are learnable parameters. In addition, the variables $f_{app} \in \mathbb{R}^{2048}$ and $f_{spa} \in \mathbb{R}^{4}$ represent the ROI appearance and spatial features of the corresponding bounding boxes, while $G(s) \in \mathbb{R}^{300}$ and $G(o) \in \mathbb{R}^{300}$ represent the GloVe word embeddings for the subject and object, respectively. Moreover, $[\cdot;\cdot;\cdot]$ denotes the concatenation operation.

From here onwards, for the sake of simplicity of notations, we represent $f_s(B_j^{i})$ and $f_o(B_k^{i})$ as $f_s$ and $f_o$ respectively.
It should be noted that $f_s$ and $f_o$ can be used directly for
obtaining relationship embeddings that can be used to compute frame-level and visual relationship similarity potentials. However, to further enrich these representations, we present the following two attention-based aggregation techniques:
\newline
\noindent\textbf{(i) Global Semantic Aggregation (GSA):}
The subject and object representations, i.e. $f_s$ and $f_o$ learned using eq. (\ref{eq:initial_f_s}) and (\ref{eq:initial_f_o}) are independent of other frames in the video. However, the global semantic context information from other frames may help in enriching subject and object representation. Therefore, we fused the $I3D$-features~\cite{carreira2017quo} of every frame by weighting them with a global attention vector $\alpha_{s}^g$ which is computed as follows:

\begin{equation}
\alpha_s^g = G_{Att}(f_{I3D}, G(s)).
\end{equation}

Where $G_{Att}$ is a learnable attention unit, and it is defined as follows:

\begin{equation}
\label{eq:s_j_g}
s_j^g = \textbf{W}_{gs_1}ReLU(\textbf{W}_{gs_2}[f_{I3D}^j; G(s)] + b_{gs}),
\end{equation}

\begin{equation}
\label{eq:a_s_j_g}
\alpha_{s_j}^g = softmax(s_j^g),
\end{equation}

where, $\textbf{W}_{gs_1}$, $\textbf{W}_{gs_2}$ and $b_{gs}$ are learnable parameters and $f_{I3D}^j$ represent the $I3D$-feature for $j^{th}$ frame. Finally, after aggregating the global semantic information the subject representation is obtained as:
\begin{equation}
    \label{eq:f_s_g}
    f_s^g = \sum_{j=1}^{T} (\alpha_{s_j}^g \odot f_{I3D}^j) \cdot f_s.
\end{equation}

Similarly, object representation after global semantic aggregation is obtained as:
\begin{equation}
    \label{eq:f_o_g}
    f_o^g = \sum_{j=1}^{T} (\alpha_{o_j}^g \odot f_{I3D}^j) \cdot f_o.
\end{equation}
Here, $f_s$ and $f_o$ are initial subject and object features obtained using eq. (\ref{eq:initial_f_s}) and (\ref{eq:initial_f_o}), respectively.
\newline
\newline
\noindent(ii) \textbf{Local Localization Aggregation (LLA):}
For adding the context from the adjacent frames as a local localization context, we considered a window size of five, i.e., for $i^{th}$ frame, we considered $i-2$ to $i+2$ frames. The local context helps with partially visible or occluded objects. We fuse the local spatial and ROI features from the adjacent frames weighted by the local attention vector for the subject $s$ ($\alpha_{s}^l$) which is computed as follows :
\begin{equation}
    \alpha_s^l = L_{Att}(f_l^{-2\leq t \leq 2}, f_s).
\end{equation}
Here, $f_s$ is the initial representation of subjects obtained from eq. (\ref{eq:initial_f_s}) and $f_l^{-2\leq t \leq 2}$ is obtained as:

\begin{equation}
f_s^{-2\leq t \leq 2} = \textbf{W}_{ls_2}ReLU(\textbf{W}_{ls_1}[(f_s^{t-2}-f_s^{t-1})\cdot (f_s^{t+1}-f_s^{t+2})],
\end{equation}

where $f_s^{t-2}$ to $f_s^{t+2}$ are stacked ROI and spatial features of all objects from their respective frame numbers. $\textbf{W}_{ls_1}$, $\textbf{W}_{ls_2}$, $\textbf{W}_{ls_3}$ and $\textbf{W}_{ls_4}$ are learnable parameters. Then, from eqs. (\ref{eq:s_j_g}) to (\ref{eq:f_o_g}), local localization information-aggregated subject and object features, $f_s^l$ and $f_o^l$, are obtained.
\newline
\newline
\noindent\textbf{Obtaining Relationship Embedding:}
To obtain relationship embedding, we first  enrich subject and object representations as follows:
\begin{equation}
    \label{eq:f_s_full_context}
    f_s =\textbf{W}_{sr_1}ReLU(\textbf{W}_{sr_2}[f_s^g;f_s^l]),
\end{equation}

\begin{equation}
    \label{eq:f_o_full_context}
    f_o = \textbf{W}_{or_1}ReLU(\textbf{W}_{or_2}[f_o^g;f_o^l]).
\end{equation}

Where, $\textbf{W}_{sr_1}$, $\textbf{W}_{sr_2}$, $\textbf{W}_{or_1}$ and $\textbf{W}_{or_2}$ are learnable parameters. Finally, for any pair of subject and object, $(j,k)$ of $i^{th}$ frame of a video $v$, the relationship embedding with respect to query visual relationship $r$ is computed as a translation vector in lower dimensional relation space similar to VTrasE~\cite{zhang2017visual} as follows:
\begin{equation}
\label{eq:final_emb}
  f_r{(u_{i_v}^{jk})} = \textbf{W}_{r_2}ReLU\Big(\textbf{W}_{r_1}ReLU\big([\textbf{W}_{rs}f_j - \textbf{W}_{ro}f_k\big);G(p)]\Big),
\end{equation}
where,  $\textbf{W}_{r_2}$, $\textbf{W}_{r_1}$, $\textbf{W}_{rs}$, and  $\textbf{W}_{ro}$ are learnable parameters and $f_j$, $f_k$ are subject and object representation obtained from eq. (\ref{eq:f_s_full_context}) and (\ref{eq:f_o_full_context}) respectively.

\subsection{\label{sec:rel_net}Learning Relation Network with Few Examples}
The proposed problem formulation has to learn the similarity between object pairs as the visual relationships using a few videos from $S_{sup}$. We have selected the Relation Network ($R_{\bm{\theta}}$)~\cite{sung2018learning}, a metric-based meta-learning approach, as our method of choice. After learning the similarity measure for the unseen predicate, $R_{\bm{\theta}}$ is used to compute both frame-level as well as visual relationship similarity potentials.

For a pair of object pairs $p_1 = (x_1, y_1)$ and $p_2 = (x_2, y_2)$, their representation as a visual relationship, $f_{r}^{p_1},f_{r}^{p_2}$ are obtained using eq.~(\ref{eq:final_emb}). Then pairwise similarity score is computed as:
\begin{equation}
R_{\bm{\theta}}(f_{r}^{p_1},f_{r}^{p_2}) = \textbf{W}_r^T\Phi(f_{r}^{p_1},f_{r}^{p_2}) + b,
\label{eq4}
\end{equation}

where $\textbf{W}_r$, $b$ are learnable parameters matrix and bias vector respectively. Further, $\Phi$ is computed using the following equation:

\begin{equation}
\label{eq5}
\Phi(f_{r}^{p_1},f_{r}^{p_2}) = tanh(\textbf{W}_1([f_{r}^{p_1};f_{r}^{p_2}]) + b_1)
\sigma(\textbf{W}_2[f_{r}^{p_1};f_{r}^{p_2}] + b_2) + ((f_{r}^{p_1}+f_{r}^{p_2})/2),
\end{equation}

where $\textbf{W}_1, \textbf{W}_2$, and $b_1, b_2$ are learnable parameters matrices, and bias vectors, and $\sigma$ and $tanh$ are sigmoid and hyperbolic tangent activation functions, respectively.

For each support set video, we extract ground truth positive and negative object pairs. Positive object pairs are connected via the query relation predicate $p$, while negative object pairs are picked randomly from a set containing pairs of objects that are not connected via predicate $p$. We used ground-truth spatial and its corresponding ROI features to get the relation embedding using eq. (\ref{eq:final_emb}).  $R_{\bm{\theta}}$ learns to return higher similarity between semantically similar pairs of relationships while lower similarity for other pairs. Let us consider a set of positive object pairs $p_r^+ = \{\{(x_{ij}^+, y_{ij}^+)\}_{j=1}^{l}\}_{i=1}^{k}$ and negative object pairs $p_r^- = \{\{(x_{ij}^-, y_{ij}^-)\}_{j=1}^{l}\}_{i=1}^{k}$  extracted from the $S_{sup}$, where $l$ is the number of positive or negative object pairs extracted from each support set video. Finally, the $R_{\bm{\theta}}$ is meta-trained on $S_{sup}$ using the following episodic loss.
\begin{equation}
\begin{split}
\mathcal{L} = \sum_{a=1}^{K}\sum_{b=1}^l\sum_{c=1}^{l}log\Bigg(\Big(1+e^{-R_{\bm{\theta}}(f_r^{+bc}, f_r^{+bc})}\Big)  \\\Big(1+e^{-R_{\bm{\theta}}(f_r^{+bc}, f_r^{-bc})}\Big)\Big(1+e^{-R_{\bm{\theta}}(f_r^{-bc}, f_r^{+bc})}\Big) \Bigg).
\end{split}
\end{equation}

\subsection{Trajectory Generation}
To generate the final trajectories of the subject and object, we used three different optimization approaches. In the first technique, we used only frame-level potential $\Psi_{i}$ to select a node with the optimal potential value, which was determined as the minimum value among all the nodes at frame $i$. We skipped a frame if the optimal potential value is higher than the fixed threshold ($= -0.5$). As an alternative to this, we also used a greedy solver by selecting a node from frame $i$  using the first technique, and then, we find the next node in the next frame $j$ with optimal $\Psi_{ij}$. We skipped a frame $j$ if the optimal $\Psi_{ij}$ potential value is higher than the fixed threshold ($= -0.5$). In our full model, we used both frame-level potential and visual relationship similarity potential and used belief propagation using message passing \cite{pearl2022fusion} to find the optimal trajectories.

\section{\label{sec:expts}Experiments and Results}
\subsection{\label{sec:data}Datasets}
We have used two video benchmark datasets, namely, VidOR~\cite{vidor,thomee2016yfcc100m} and ImageNet-VidVRD~\cite{vidvrd} in our experiments. ImageNet-VidVRD contains 1000 videos obtained from the ILVSRC2016-VID dataset~\cite{ILSVRC15}. It has 132 predicates and 35 object categories. VidOR is a large-scale dataset containing 10,000 videos obtained from YFCC100M collection. VidOR contains 50 predicates and 80 object categories. For our problem setting,
we split both datasets into disjoint sets based on predicates and videos by randomly assigning 35 and 15 predicates to the train and test sets of VidOR, while assigning 88 and 22 predicates to the train and test sets of ImageNet-VidVRD, respectively.

\subsection{\label{sec:pm}Performance Measure}
In this work, we adopt widely-used evaluation metrics in the referring relationship and video understanding literature, as described in \cite{krishna2018referring,chen2019weakly,xiao2020visual}. Specifically, we calculate the subject and object accuracy, which represents the percentage of correct trajectories returned by the model for the entire test set. A relation accuracy is defined as the percentage of correct subject and object pair trajectories. Here, a pair is considered correct if both subject and object trajectories are correct. For spatiotemporal accuracy, $A_{s-t}^{sub}$, $A_{s-t}^{obj}$, of the subject or object, a trajectory is considered correct if at least 50\% of the bounding boxes across frames have $\geq$ 0.5 intersection over union (IoU) with respect to ground truth bounding boxes. Similarly, for spatiotemporal accuracy of relation, $A_{s-t}^r$, a pair of subject and object is considered true only if both trajectories are spatiotemporally accurate (i.e. at least 50\% of the bounding boxes across frames have $\geq$ 0.5 IoU). For spatial accuracy, $A_{s}^{sub}$, $A_{o}^{obj}$, a trajectory is considered true if the average of IoUs of the bounding box with respect to ground truth is at least 0.5. The mean IoU for a trajectory is an average of the IoU score for all bounding boxes with respect to ground truth. The mean IoU of the subject or object, $mIoU_s^{sub}$, $mIoU_s^{obj}$ is the average of the mean IoU of the subject or object for the entire test set.

\subsection{\label{sec:cvpr_base}Baselines}
To the best of our knowledge, we are the first to approach referring relationships in videos in a \emph{few-shot setup}, and there are no existing baselines in the literature that can be directly compared to our approach. Therefore, we adopt closely related methods as our baselines to compare the effectiveness of our method as follows:
\subsubsection{(i) Few-Shot Visual Relation Co-Localization (VRC):} Few-shot VRC has originally been proposed for localizing common subjects and objects in a bag of images in a few-shot setup~\cite{teotia2021few}. We adopt it for videos by treating object trajectory pairs in the test and support set videos as a bag and performing visual relationship co-localization at the frame level.
\subsubsection{(ii) Visual vRGV~\cite{xiao2020visual}} tackles the same task as ours but differs significantly in supervision. We adapted vRGV for our few-shot problem setup for a fair comparison. Recall that in our problem setup, the model is trained in episodes, and each episode contains a support set of a few videos and a test video for evaluation. We utilize all available videos from the train set to train the vRGV model. During testing for each episode, we fine-tune the model on the support set and then perform localization for the given query visual relationship on the test video. We refer to these baselines as vRGV in a few-shot setup or vRGV-fs.
\subsubsection{(iii) Tracklet-based approach:}
In this section, we provide a detailed explanation of the proposed tracklet-based. Consider notations independent of the main method.
\\ \textbf{Tracklet Detector.}
For each video, using a pre-trained detector, we extract all tracklets (represented as a set $T_e$) features and spatio-temporal locations. Each entity $e_i \in T_e$ having length $l_i$ as number of frames is categorized by a time slot $(s_{i}^e, e_{i}^e)$ and bounding box coordinate $b_i \in R^{l_i\times4 }$.
For each entity $e_i$, tracklet feature $feat_i$ is a combination of spatial and appearance features. Spatial feature $feat_{i}^s \in R^{l_i\times 8}$ is formed as the concatenation of all box offset $\Delta b_i$ and coordinate $b_i$, where $\Delta b_{i,j}$ is defined between two consecutive frames as box coordinate offset, i.e., $\Delta b_{i,j} = b_{i, j+1} - b_{i,j}$. For appearance $feat_{i}^a \in R^{l_i\times d_a}$ each frame is extracted using RoIAlign~\cite{FasterRCNN} based on box locations. For entity $e_i$, the tracklet feature $feat_i \in R^{l_i\times d_e}$ is formed as,
\begin{equation}
 feat_i = f_a[f_b(feat_{i}^a);f_c(f_{i}^s)].
\end{equation}
where $f_a$, $f_b$ and $c$ are learnable linear models and [;] is concatenation operation.
\\ \textbf{Relationship Pair Proposal.}
The goal of this module is to propose relationship pair proposals. Given a video $v$, the first tracklet set $T_e$ is extracted using a pre-trained detector. An attention score is calculated for each tracklet for finding whether a tracklet contains a subject similar for the object. A tracklet having a high attention score is more likely to contain the desired object/subject category. To this end, corresponding to subjects, an attention vector is computed $\alpha_{s_{n\times 1}} = SubAtt(T_e, w2v(s))$, where $SubAtt$ is attention unit which takes all tracklet features along with word2vec features of query subject and predicate. Similarly for objects, we have $\alpha_{o_{n\times1}} = ObjAtt(T_e, w2v(o)$. To calculate attention vectors $\alpha$ first a score for each entity (subject/object) is computed as,

\begin{equation}
s_j = f_d(f_e[feat_j;w2v(s)]).
\end{equation}

where $f_d$ and $f_e$ are learnable linear functions. Finally, the attention vector is computed as,

\begin{align}
\alpha_s = SubAtt(T_e, w2v(s)), where, \newline \alpha_j = \frac{e^{s_j}}{\sum_{i=1}^{n} e^{s_i} }.
\end{align}

Similarly, an attention vector $\alpha_o$ corresponding to an object is computed. To make entity tracklet features aware of each other we use an attention-shifting mechanism~\cite{xiao2020visual}, where attention is shifted from one entity to every other entity. Attention shift from subject to object computed as, $\alpha_{so} = W_{so}\alpha_s$, similarly object to subject $\alpha_{os} = W_{so}\alpha_o$, where $W_{so}$ and $W_{os}$ are learnable parameters. Then entity tracklet features aware of each other computed as, $feat_s = f_{os}[\alpha_{os}; w2v(p)]$ and $feat_o = f_{so}[\alpha_{so}; w2v(p)]$, where $feat_s$ and $feat_s$ are objects and subject features respectably and $f_{os}$ and $f_{so}$ are linear learnable modules. Finally, probable query relationship pair proposals are computed as,
\begin{equation}
    feat_p = f_e(ReLu(f_d([\phi(feat_s;feat_o)], w2v(p)))).
\end{equation}
where $feat_p$ is a set pair proposals features, and $\phi()$ is function to enumerate all possible pair proposals.

\textbf{Metric Learner.}
This module aims to rank the pair proposals produced by the pair proposal module. A deep metric is learned with the help of support set videos in a few-shot way. Deep metric learns to project positive pair proposals (pair proposals with the same predicate) close in the feature space while negative pair proposals (pair proposals with different predicates) are far away.\\
\textbf{Feature Extraction.} Given a $k^{th}$ support set video, we have ground truth bbox tracklets of entities (subject) $feat_{k_s}^s \in R^{l_k\times 8}$ of query predicate. We extracted appearance feature $feat_{k_s}^{l_k\times d_a}$ based on bbox using a pre-trained detector. To incorporate spatial and visual features, the final subject features $feat_{k_s}$ is computed using equation 1. Similarly, object feature $feat_{k_o}$ is computed. Then positive predicate feature is computed as $feat_{p(+ve)} = MLP[ f_{k_s}; f_{k_s};w2v(p)]$. Similarly, for each supported video, negative predicate feature $feat_{p(-ve)}$ is computed where the predicate is not the same as in the query relationship.\\
A metric learner $\mathbf{ML}$ is defined as a linear differentiable model and trained using the ground truth positive and negative predicate feature in a supervised setting. The loss function for this module is defined as,

\begin{equation}
\begin{split}
        \mathcal{L}_1 = \sum_{i=1}^k\sum_j((ML(f_{p_i(+ve)}) - ML(f_{p_j(+ve)})) \\ + (ML(f_{p_i(-ve)}) -  ML(f_{p_j(-ve)})) \\ - (ML(f_{p_i(+ve)}) - ML(f_{p_j(-ve)})) \\ - (ML(f_{p_i(-ve)}) - ML(f_{p_j(+ve)}))).
        \end{split}
\end{equation}

The pair proposals produced by the relationship proposal module are ranked by the learned deep metric \textbf{$ML$}. The top-ranked pair proposal $feat_t$ has the lowest $l_2$ distance from all the positive ground truth predicates. The top-ranked pair is returned by the following equation.

\begin{equation}
    feat_t = \min_{i=1}^{n}(\sum_{j=1}^{k} ML(f_i) - ML(f_{k(+ve)})).
\end{equation}

\textbf{Reconstruction module.}
This module aims to ensure the effective use of word embedding of query relationships in the aforementioned modules. We took the inspiration of reconstruction as weak supervision~\cite{rohrbach2016grounding,xiao2020visual} to use query relationship embedding effectively. The first few (set as hyperparameters) top-ranked predicate feature returns from the previous module are stacked together to form an encoding for the LSTM, i.e., $f_{emb} = MLP(\psi(f))$, where $\psi$ return the top-ranked stacked predicate features. LSTM considers relation as a textual phrase and reconstructs it using a single layer of LSTM.
This module is trained using the following loss function,
\begin{equation}
    \mathcal{L}_2 = -\sum_{i=1}^{l_r}logP(w_i|w_{0:i-1}, f_{emb}).
\end{equation}

\textbf{Training and Inference.}
The top-ranked feature contains the tracklet information of the subject and object related to the query predicate. But the query visual relation might not span the whole duration of the subject and object and can be in multiple time slots. To correctly find the duration and probable multiple slots we learn another linear model to predict each frame whether it belongs to a query relationship or not. The linear model is defined as,
\begin{equation}
    r_d = softmax(MLP([f_{emb};f_t]))
\end{equation}
$r_d$ has a dimension equal to twice the max frame length in the dataset. For each common frame of subject and object, we have a score for starting of the relationship and similarly for the end of a relationship. This module is trained using the following supervised loss,
\begin{equation}
    \mathcal{L}_3 = -tIoU(bbox_{gt}, bbox_{pred})
\end{equation}
$tIOU$ computes the temporal intersection over the union of predicted predicates with respect to the ground truth bounding boxes.
The start and end of the query relation are extracted as,
\begin{equation}
    F_s = \{i \: ; r_{d}[i] > \epsilon\}_{i=1}^{d/2},
    F_e = \{i \: ; r_{d}[i] > \epsilon\}_{i=d/2}^{d}
\end{equation}

The frames which are not contained within the common duration of the subject and object are removed. The lowest frame number $i_s$ is selected from $F_s$ as the start of the relationship and the lowest frame from $i_e$ such that $i_e > i_s$ is selected from $F_e$ as the end of the relation. The frames contained in between $i_s$ and $i_e$ are removed from $F_s$ and $F_e$ to find another query relation instant in a similar way. \\
Finally, all modules are combined during training as an end-to-end differential function. The overall model is trained using the following loss function,
\begin{equation}
    \mathcal{L} = \mathcal{L}_1 + \lambda_1\times\mathcal{L}_2 + \lambda_2\times\mathcal{L}_3
\end{equation}

where $\lambda_1$ and $\lambda_2$ are hyperparameters.

\begin{table*}[!t]
    \centering
   \resizebox{\textwidth}{!}{ \begin{tabular}{lc ccccc ccccccc}
   \toprule
         \multicolumn{2}{c}{} &
         \multicolumn{5}{c}{\textbf{VidVRD}} &
         \multicolumn{5}{c}{\textbf{VidOR}} & \\
         \cmidrule(r){3-7}
         \cmidrule(r){8-12}
         Approach & Support Set Size & $A_{s-t}^{sub}$ &$mIoU_s$ &$Acc_{s-t}^{obj}$ &$mIoU_o$ & $A_{s-t}^{r}$ & $A_{s-t}^{sub}$ &$mIoU_s$& $A_{s-t}^{obj}$ &$mIoU_o$& $Acc_{s-t}^{r}$ \\
         \toprule

         tracklet-based & 2 & 12.3 &12.1 & 12.2 & 11.7 & 8.8 & 10.9 & 10.7& 10.1 & 10.3 & 7.3  \\

         tracklet-based & 4 & 12.6 & 12.9 & 12.8 &12.0 & 9.6 & 11.1 & 11.0 & 10.9 &10.4 & 8.2  \\

         tracklet-based & 8 & 13.4 &13.2& 13.7 &13.3& 10.4 & 12.8 & 12.6& 13.2 &13.0 & 9.7 \\
         \toprule
         ~~w/o vis.sim. potential, w/o GSA, w/o LLA & 2 &19.2 &18.2& 17.8 & 19.9& 15.9 & 19.3 &18.1 & 18.4 & 19.8 & 16.7 \\
         ~~w/o vis.sim. potential, w/o GSA, w/o LLA & 4 &19.6 &19.1& 18.2 & 20.8& 16.3 & 19.8 &18.7 & 18.5 & 20.4 & 17.3 \\
         ~~w/o vis.sim. potential, w/o GSA, w/o LLA & 8 &20.2 &19.6& 18.5 & 21.6& 17.1 & 20.2 &19.4 & 18.7 & 20.9 & 18.4  \\

         ~~w/o vis.sim. potential & 2 & 21.4 & 20.6 &22.1 & 21.3 & 16.9 & 19.3 & 21.4  &21.4 & 21.8 &17.2 \\
         ~~w/o vis.sim. potential & 4 & 22.6 & 21.9 &22.3 & 22.9 & 17.2 & 20.9 & 22.5  &22.7 & 22.1 &17.6 \\
         ~~w/o vis.sim. potential & 8 & 22.9 & 22.3 &22.8 & 23.7 & 17.8 & 21.3 & 22.8  &23.4 & 23.7 &18.6\\
         ~~Greedy Solver & 2 & 23.8 &23.2 &22.9 &25.2& 22.4 & 23.7 &21.7 & 23.4 & 24.7 & 20.4\\
         ~~Greedy Solver & 4 & 25.4 &24.7 &24.3 &26.4& 22.8 & 24.2 &23.8 & 24.5 & 25.8 & 21.7 \\
         ~~Greedy Solver & 8 & 25.7 &25.4 &25.1 &26.7& 23.1 & 24.8 &23.8 & 24.7 & 26.4 & 22.3\\
         ~~Full model & 2 & 24.5 &26.3 &25.7 &26.7& 24.3 & 24.8 &25.7 & 26.1 &26.8 & 22.6\\
         ~~Full model & 4 & 26.8 &27.1 &26.0 &27.9& 25.1 & 25.3 &26.2 & 26.3 &27.0 & 23.8   \\
         ~~Full model & 8 & 27.4 &28.3 &26.8 &28.2& 25.7 & 26.1 &27.3 & 26.8 &27.6 & 24.4\\

    \hline
    \end{tabular}}
    \caption[ Comparison of baseline and proposed method]{\label{tab:supp_res} Comparison of baseline and proposed method on different support set sizes. For every method (baseline and all variants of our method) the performance is increased with an increase in support set size.}
\end{table*}

\subsection{Implementation Details}
For frame-level detection, we utilized FasterRCNN~\cite{FasterRCNN}, which was pre-trained on MS-COCO with ResNet-101 as the backbone. We extracted the 30 most confident object bounding boxes for each frame. Our implementation was done using PyTorch. We optimized the model parameters using Adam~\cite{kingma2014adam} with an initial learning rate of 1e-5, while also employing a dropout rate of 0.3 to reduce overfitting. To prevent overfitting, we used early stopping. We trained the model on an Nvidia-RTX A6000 GPU.

\begin{table*}[!t]
    \centering
    \resizebox{\textwidth}{!}{ \begin{tabular}{lccccccc ccccccc}
    \toprule
         \multicolumn{1}{c}{} &
         \multicolumn{6}{c}{\textbf{ImageNet-VidVRD}} &
         \multicolumn{7}{c}{\textbf{VidOR}} & \\
        \cmidrule(r){2-8}
        \cmidrule(r){9-15}
         Method & $A^{sub}_{s-t}$ & $A^{sub}_{s}$ & $mIoU^{sub}_s$& $A^{obj}_{s-t}$ & $A^{obj}_{s}$ & $mIoU^{obj}_{s}$& $A^r_{s-t}$ & $A^{sub}_{s-t}$ & $A^{sub}_{s}$ & $mIoU^{sub}_s$& $A^{obj}_{s-t}$ & $A^{obj}_{s}$ & $mIoU^{obj}_s$ & $A^r_{s-t}$ \\
         \toprule
         VRC~\cite{teotia2021few}  & 8.4 & 10.7 & 11.9 & 8.6 & 10.3& 11.2 & 5.8 & 7.2 & 9.9 & 10.3 & 8.0 &10.1& 11.1 & 5.6 \\

         vRGV-fs~\cite{xiao2020visual} & 8.2 & 11.1 & 12.4 & 8.3 & 11.6& 12.9 & 6.1 & 7.5 & 10.3 & 10.9 & 7.3 &10.7 & 11.4 & 6.3   \\
         Tracklet-based &12.6 & 15.1& 15.9 & 12.8 & 14.8& 15.3 & 9.6 & 11.1 & 13.7& 14.2 & 10.9 & 12.6& 13.8 & 8.2 \\
         \midrule
         \textbf{Ours}   \\
         ~~w/o vis.sim. potential, w/o GSA, w/o LLA &19.6 & 21.4 &19.1& 18.2 & 18.7 & 20.8& 16.3 & 19.8 & 20.7 &18.7 & 18.5 & 19.2 & 20.4 & 17.3 \\
         ~~w/o GSA, w/o LLA & 25.3 & 26.9 & 25.8 &25.7&25.1& 26.5 & 22.8 & 24.5& 24.9 & 25.1& 24.9 & 25.1& 22.3 & 21.9 \\
         ~~w/o vis.sim. potential & 22.6& 22.8 & 21.9 &22.3&21.4& 22.9 & 17.2 & 20.9& 21.3& 22.5  &22.7 & 21.7 & 22.1 &17.6 \\
         ~~Greedy solver & 25.4 & 26.5 &24.7 &24.3& 23.8 &26.4& 22.8 & 24.2 & 25.6&23.8 & 24.5 &22.7 & 25.8 & 21.7 \\
         ~~Full model & \textbf{26.8} & \textbf{28.4} & \textbf{27.1} &\textbf{26.0} & \textbf{25.1} & \textbf{27.9} & \textbf{25.1} & \textbf{25.3} & \textbf{26.9} & \textbf{26.2} & \textbf{26.3} & \textbf{25.7} & \textbf{27.0} & \textbf{23.8}   \\
         \bottomrule
\end{tabular} }
    \caption[Performance of few-shot referring relationship in videos]{\label{tab:mainRes} Performance of few-shot referring relationship in videos. Each method is trained on the same split of train and test datasets with a support size of four videos. Ours (full model) represents the proposed method where global semantic aggregation (GSA) and local localization aggregation (LLA) are performed to enrich the query-conditioned relationship representation, a translational relation embedding is learned, and an objective function containing frame-level and visual similarity potential is optimized using belief propagation on the $T$-partite random field. For more details, refer Section~\ref{sec:expts}.}
\end{table*}

\subsection{\label{sec:rd}Results and Discussions}
We compare our approach with baselines discussed in Section~\ref{sec:cvpr_base} on VidOR and ImageNet-VidVRD datasets in Table~\ref{tab:mainRes}. We observe that the baselines modified using prior works, namely VRC and vRGV, exhibit weaker performance across all the evaluated metrics. The tracklet-based approach performs marginally better than these baselines. However, its performance is heavily bottle-necked with the quality of tracklets generated. Our approach, without the proposed global semantic and local localization aggregation itself, surpasses these baselines. We also observed a positive impact of frame-level and visual relationship similarity potential; for example, in Table~\ref{tab:mainRes}, the accuracy of localizing relationship ($A_{s-t}^r$) has increased by 2.3\% (absolute scale) in the full model with respect to the greedy solver on ImageNet-VidVRD dataset. The superior performance of our method can be attributed to the principled optimization framework, the proposed aggregation techniques, query-conditioned learning of relationship embeddings, and the effective utilization of support set videos to meta-train relation networks. To show the efficacy of our method and to justify the choice of different modules, we conduct the following ablations and analyses:
\subsubsection{Global semantic and local localization aggregation:}
We remove global semantic (GSA) and local localization aggregation (LLA) in our full model to analyze their effect. In other words, we used equations~(\ref{eq:initial_f_s}) and~(\ref{eq:initial_f_o}) to obtain the subject and object representation and analyze its effect on the ImageNet-VidVRD dataset in Table~\ref{tab:loc-glob}. We observe that our full model that uses these two aggregation techniques performs better as compared to the settings where one or both of these are removed.

\begin{table}[!t]
    \centering
    \begin{tabular}{c|c|cccc}
        \toprule
        GSA & LLA & $A_{s-t}^{sub}$ &

         $A_{s-t}^{obj}$ & $A_{s-t}^r$ &\\
        \midrule
        \xmark & \xmark & 21.3 & 22.1& 20.8 & \\
        \cmark & \xmark & 24.7 & 23.5 & 22.8 & \\
        \xmark & \cmark & 22.3 & 21.7& 21.1 & \\
        \cmark & \cmark & 26.8 & 26.0 & 25.1&\\
        \bottomrule
    \end{tabular}
    \caption[Ablation study]{\label{tab:loc-glob}Ablation study to demonstrate the importance of global semantic aggregation (GSA) and local localization aggregation (LLA) on the ImageNet-VidVRD dataset. Our full model (the last row) performs better as compared to the settings where one or both of these are removed.}
\end{table}

\begin{table*}[!t]
    \centering
    \begin{adjustbox}{width=1\textwidth}
     \begin{tabular}{lc ccccc cccccccccc}
       \toprule
       \multicolumn{1}{c}{} &
        \multicolumn{7}{c}{\textbf{Static}} &
        \multicolumn{8}{c}{\textbf{Dynamic}} & \\
        \cmidrule(r){2-9}
        \cmidrule(r){10-17}
       Metric & stop front & taller & right & larger & stand front & sit behind & above & left & walk behind & run beneath & move right & play & feed & faster & move front & fly above \\
       \midrule
       $A_{s-t}^{sub}$ & 30.1 & 28.8 & 28.6 & 30.9 & 29.7 & 26.5 & 31.4 & 25.8 & 19.2 & 21.5 & 23.7 & 27.4 & 28.5 & 22.4 & 25.9 & 22.8 \\
        $A_{s-t}^{obj}$ & 28.3 & 30.1 & 27.5 & 27.2 & 26.5 & 23.8 & 30.1 & 23.3 & 19.1 & 21.7 & 22.8 & 28.9 & 27.2 & 22.3 & 24.4 & 24.3 \\
       $A_{s-t}^{r}$ & 27.6 & 26.2 & 26.3 & 25.5 & 25.9 & 22.2 & 27.8 & 22.5 & 18.6 & 19.8 & 22.6 & 26.2 & 26.1 & 19.2 & 23.1 & 21.4 \\

     \bottomrule
    \end{tabular}
    \end{adjustbox}
\caption[Predicate-wise Performance]{\label{tab:rel_type_comp} Predicate-wise performance of the proposed method on ImageNet-VidVRD Dataset grouped into static and dynamic categories.}
\end{table*}

\subsubsection{Translational relation embedding:} In the second ablation, we justify the choice of translation visual relationship embedding. To this end, we replaced the translational embedding in eq.~(\ref{eq:final_emb}) with a simple concatenation of subject and object representation. As shown in Table~\ref{tab:abltion_emb}, translational relationship embedding used in our model is significantly more robust than simple concatenation embeddings.

\begin{table}[!t]
    \centering
    \begin{tabular}{c | cccc}
        \toprule
         Relationship Embedding & $A_{s-t}^{sub}$ &

         $A_{s-t}^{obj}$ & $A_{s-t}^r$ &\\
          \midrule
        Concatenation & 18.3 & 20.1 & 18.7 & \\
        Translation Embedding & 26.8 & 26.0 & 25.1&\\
        \bottomrule
    \end{tabular}
    \caption[Performance comparison of translational and Concat relationship embedding]{\label{tab:abltion_emb} Performance comparison of translational and Concat relationship embedding on ImageNet-VidVRD dataset. In this study, both embeddings are fused with the global and local context.}
\end{table}

\begin{figure}
    \centering
  \includegraphics[width=0.75\textwidth]{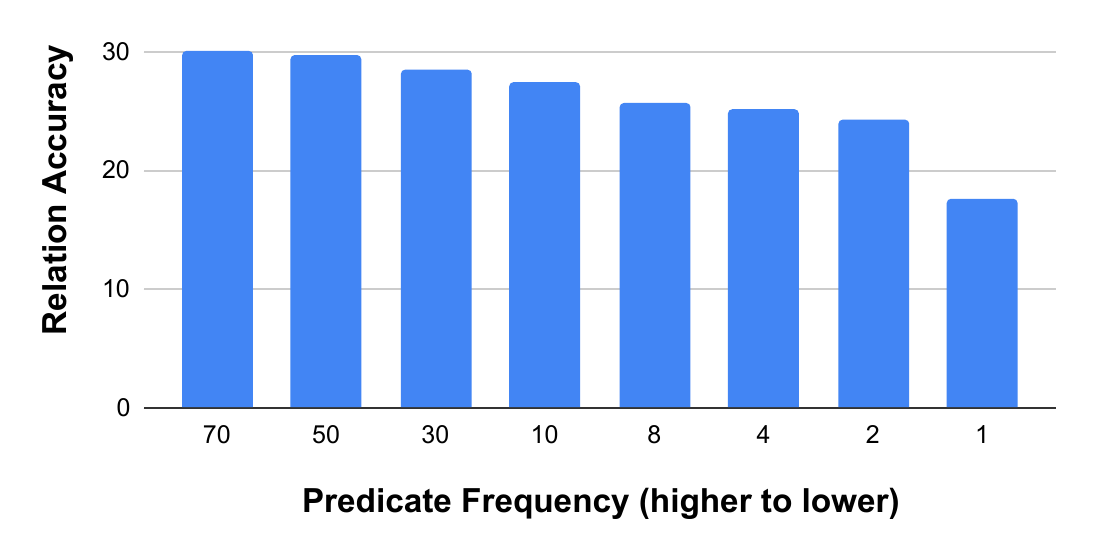}
    \caption[spatiotemporal localization accuracy]{\label{fig:bar_plot} Spatiotemporal localization accuracy of the proposed method with different support sizes on the ImageNet-VidVRD.}
\end{figure}

\subsubsection{Effect of visual relationship similarity potential:} To study the effect of visual relationship similarity potential in our optimization framework, we evaluate our model with and without this potential. Table~\ref{tab:mainRes} clearly indicates the utility of this potential where our full model that uses both visual relationship similarity and frame-level potential significantly surpass the variant that does not use visual relationship similarity potential.

\subsubsection{Effect of support set size/long-tail:} To analyze the effect of support set size, we perform experiments by varying its size. Specifically, we present a bar chart showing $A_{(s-t)}^r$ achieved using our approach on ImageNet-VidVRD dataset with a support set varying from 1 to 70 videos in Figure~\ref{fig:bar_plot}. We observe an intuitive gain with a larger support set.

\subsubsection{Unseen subjects or objects:} The predicate in the query visual relationship is always unseen in our setup. We further analyze cases where either subject or object is also unseen during training in  Table~\ref{tab:unseen_s_o}. The model performs reasonably well for localizing subjects and objects in videos, even in these cases with an intuitive drop in performance as compared to cases where only the predicate is unseen, and both subject and object are seen.

\subsubsection{Performance on static and dynamic relations:}
In order to analyze the performance of our approach for different types of relations, we present the predicate-wise results in Table~\ref{tab:rel_type_comp}. The results reveal that our proposed approach generally achieves better performance for static relationships. Nonetheless, despite the additional challenges posed by dynamic relationships, such as the presence of moving objects and subjects, occlusion, motion blur, and deformation, our approach demonstrates reasonable results for dynamic predicates highlighting its robustness. This gap follows directly from the structure of the objective. The frame-level potential $\Psi_{i}$ is computed from a single frame's box pair and therefore carries no motion information, while only the pairwise potential $\Psi_{ij}$ links consecutive frames. Static predicates such as ``beneath'' and ``above'' are consequently decidable from $\Psi_{i}$ alone, whereas dynamic predicates such as ``chase'' and ``follow'' depend on $\Psi_{ij}$, which is the weaker of the two signals in the few-shot regime where potentials are meta-learned from only a handful of support videos. This also indicates the natural remedy, namely richer motion features in the pairwise potential.

\subsubsection{Qualitative Results:} 
\begin{figure*}[t!]
    \centering
      \includegraphics[width=0.95\textwidth]{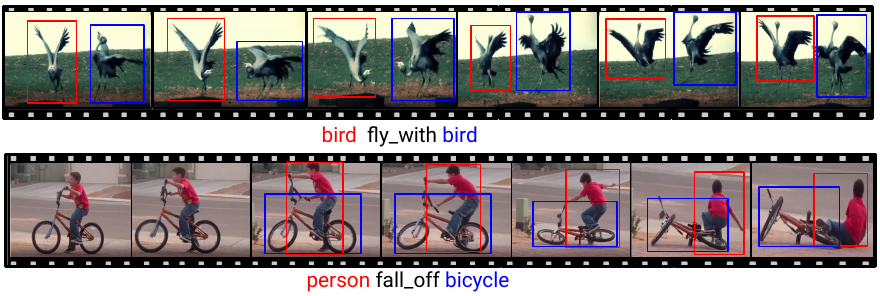}
    \caption[Qualitative result on a selection of test videos]{\label{fig:visRes}Qualitative result on a selection of test videos. Each subject and object of the query relationship is spatiotemporally localized on the test videos. Red and blue color bounding boxes represent the subject and object, respectively. \textbf{[Best viewed in color]}.}
\end{figure*}

Figure~\ref{fig:visRes} shows a selection of qualitative results of our approach on ImageNet-VidVRD dataset where query visual relationships, $<$\emph{bird}, \emph{fly with}, \emph{bird}$>$,  and $<$\emph{person}, \emph{fall off}, \emph{bicycle}$>$ are referred in the respective video by localizing the subject and object connected to the query predicate. Our method is successful in localizing subjects and objects connected via dynamic relationships indicating the effectiveness of our approach.

\begin{table}[!t]
    \centering
    \begin{tabular}{c|c|cccc}
        \toprule
        subject & object & $A_{s-t}^{sub}$ &

         $A_{s-t}^{obj}$ & $A_{s-t}^r$ &\\
        \midrule
        \cmark & \cmark & 28.3 & 27.6 & 27.3&\\
        \cmark & \xmark & 23.7& 21.4& 18.6 & \\
        \xmark & \cmark & 19.4& 20.5& 17.2& \\
        \bottomrule
    \end{tabular}
    \caption[Performance comparison for unseen subject and object]{\label{tab:unseen_s_o} Performance comparison when subject or object is unseen during training on ImageNet-VidVRD dataset.}
\end{table}

\section{Summary}
In this chapter, we approached the problem from frame-level object detection to video-level trajectory generation by optimizing an objective function on a random field in a few-shot way. We used global semantic and local localization aggregation to enhance query-conditioned translation visual relationship embedding. The objective function is minimized using belief propagation on the random field. We performed experimental comparisons along with ablation studies to show the efficacy of our approach. We firmly believe that our work will open up several future research towards the larger exciting goal of comprehensive cross-task video understanding.

In the next chapter, we move from spatial reasoning to temporal reasoning and focus on identifying when key actions or moments occur within a video.

\newcommand{\modelloc}{MATR}
\newcommand{\datas}{SportsMoments}
\newcommand{\actdata}{ActivityNet-VRL}

\chapter{Moment Grounding in Videos}
\label{chapfour}

Video-to-video moment retrieval (\emph{Vid2VidMR}) is the task of localizing unseen events or moments in a target video using a query video. This task poses several challenges, such as the need for semantic frame-level alignment and modeling complex dependencies between query and target videos.
To tackle this challenging problem, we introduce \modelloc{} (\underline{M}oment \underline{A}lignment \underline{TR}ansformer), a transformer-based model designed to capture semantic context as well as the temporal details necessary for precise moment localization. \modelloc{} conditions target video representations on query video features using dual-stage sequence alignment that encodes the required correlations and dependencies.
These representations are then used to guide foreground/background classification and boundary prediction heads, enabling the model to accurately identify moments in the target video that semantically match with the query video. Additionally, to provide a strong task-specific initialization for \modelloc{}, we propose a self-supervised pre-training technique that involves training the model to localize random clips within videos.
Extensive experiments demonstrate that \modelloc{} achieves notable performance improvements of 15.4\% in R@1 and 11.4\% in mIoU on an absolute scale compared to the state-of-the-art methods on the popular \actdata{} dataset. Additionally, on our newly proposed dataset, \datas{}, \modelloc{} shows a 16.4\% gain in R@1 and a 16.3\% gain in mIoU on an absolute scale over strong baselines.

The subsequent sections present the introduction, a review of related work, the proposed methodology, and a detailed analysis of results and discussions.

\section{Introduction}\label{sec:iccv_intro}

\begin{figure}[!h]
    \centering
      \includegraphics[width=0.8\textwidth]{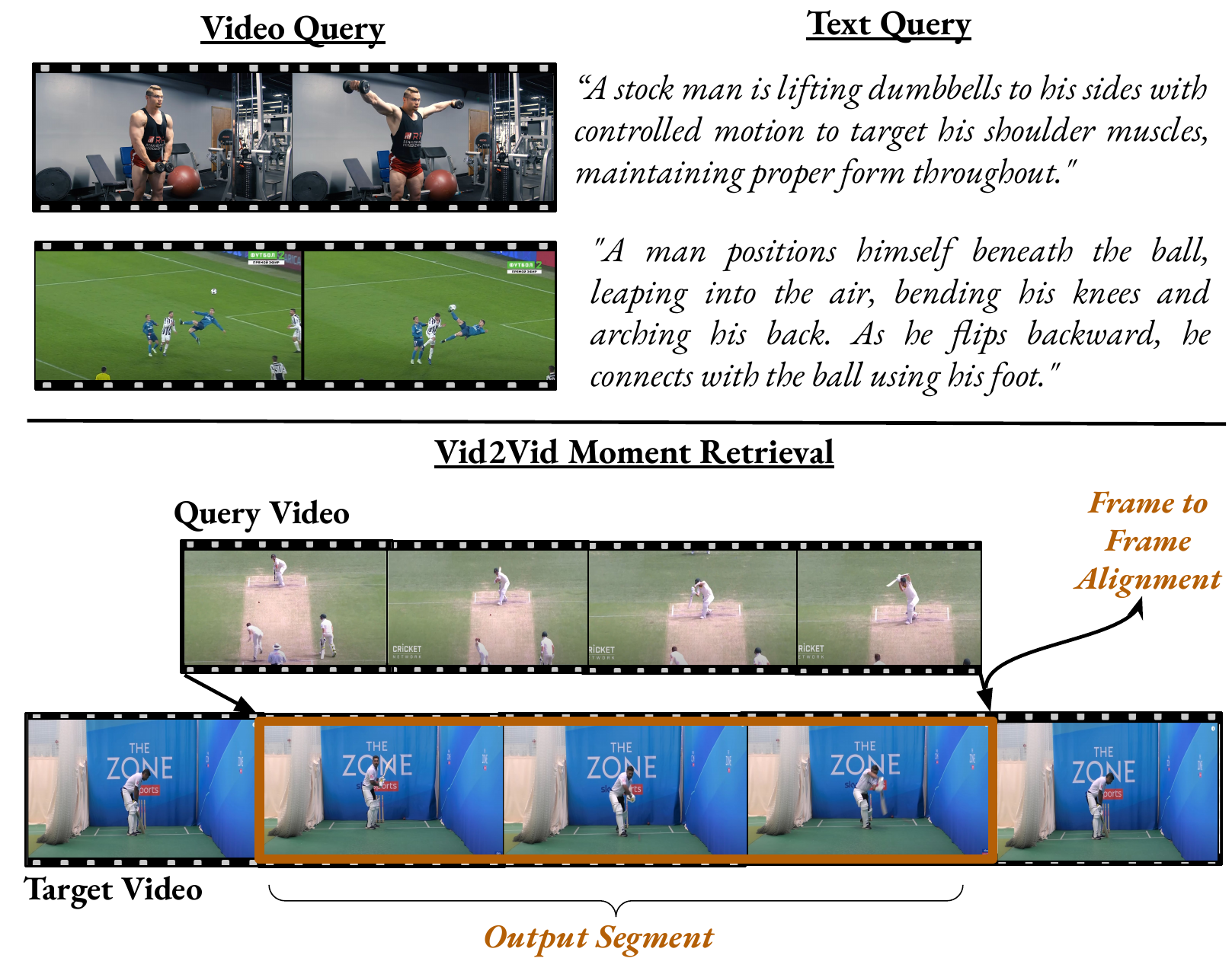}
    \caption[Video-to-video Moment Retrieval goal]{\label{fig:iccv25goal} (\textbf{Top}): Activity where text falls short to explain complex action or events, such as \textit{dumbbell side raises} and \textit{bicycle kick}, necessitating the need for a more intuitive query modality for localizing semantically matching moments. (\textbf{Bottom}):
 Shows the Vid2Vid moment retrieval setting, which is the goal of this work.
    }
\end{figure}

Video moment retrieval is the task of temporally localizing the start and end times of a moment\footnote{A moment is a continuous segment of frames within a target video that best represents the actions, events, or interactions described by the query.} in a target video described by a given query. Although text-based video moment retrieval has been extensively explored~\cite{moment_detr,QD-DETR,cg-detr,univtg,Taskweave}, it often poses challenges for users attempting to describe specific moments verbally. For example, consider a spectacular \emph{bicycle kick} in soccer (Fig.~\ref{fig:iccv25goal} top). Although an expert can search by naming or describing it in detail, a beginner in soccer may struggle to describe this impressive move accurately. They might say something like ``The player kicked the ball while in the air", which lacks the nuance needed to convey the athleticism and artistry involved. Such a description may lead to poor results for someone searching for similar moments. In contrast, if a soccer trainee shows a short video clip of a \textit{bicycle kick}, the input is rich and crisp, making it easier for a retriever to locate similar moments in a target soccer match video. This approach aligns with how users naturally would prefer to search content within the same (video) modality,
making it a more effective method for retrieving specific moments. Fig.~\ref{fig:iccv25goal} (bottom) shows a video \textit{cover drive} query in cricket that is localized in a target video.

Therefore, in this chapter, we study video-to-video moment retrieval (\emph{Vid2VidMR}) where the aim is to temporally localize a moment in the target video with a high semantic match between the moment and the query video. \emph{Vid2VidMR}, a task formally introduced by Feng et al.~\cite{video_reloc}, has many potential application areas, such as sports video analytics, educational content creation and e-learning, and surveillance systems.
This task is challenging and requires semantic frame-level alignment and modeling complex dependencies between query and target videos. This task is further complicated by the immense search space of thousands of frames, making efficient yet accurate processing essential.
The need for a semantic understanding of video content and variety in video length, context, and action speed calls for adaptive models capable of generalizing across diverse scenarios. Addressing these challenges demands advanced temporal video modeling techniques.

To address the aforementioned challenges, we propose \modelloc{}
 (\underline{M}oment \underline{A}lignment \underline{TR}ansformer) -- a method that uses explicit `dual-stage sequence alignment' to capture the required correlation and temporal details essential for accurate moment localization. By conditioning target video representations on query features, \modelloc{} produces query-aligned representations that encode the necessary correlations and temporal dependencies between the two videos. We use differentiable dynamic time warping~\cite{dtw,DropDTW} loss for aligning the query and target videos, and represent the target video by conditioning it on the query video to focus on correlated temporal features. These representations guide a classification head to discriminate relevant moments from the irrelevant background and a boundary prediction head to mark the start and end of the identified moment in the target video. Further, to enhance our model's generalization using unlabeled videos, we introduce a self-supervised pre-training strategy which involves training \modelloc{} to localize randomly sampled clips within the same video, enabling it to learn the moment localization skill in a self-supervised manner.

We evaluate \modelloc{} on public \actdata{}~\cite{video_reloc} benchmark and on our newly introduced dataset on the sports domain, viz. \datas{}, covering two of the most popular sports, namely soccer and cricket. The proposed dataset \datas{} contains complex video queries that are hard to describe in text, as shown in Fig.~\ref{fig:iccv25goal}. Also, \datas{} provides a challenging large-scale benchmark to provide a comprehensive study on vid-to-vid moment retrieval task.
Our approach achieves significant performance gains, with an improvement of 15.4\% and 11.4\% in R@1 and mIoU, respectively, on \actdata{} outperforming the state-of-the-art methods. Furthermore, \modelloc{} outperforms the implemented strong baselines on our proposed data set with 16.4\% gains in R@1 and 16.3\%  gains in mIoU, all on an absolute scale.

Our contributions are as follows: (i) We introduce \modelloc{}, which uses explicit dual-stage sequence alignment within a transformer framework between target and query video to capture temporal correlations and dependencies for accurate moment localization. (ii) We propose a self-supervised pre-training objective that enhances model initialization by understanding rich video structure without requiring any labeled data. (iii) We conduct extensive experiments and ablations to study the efficacy of our framework against competitive baselines and state-of-the-art methods. Our findings offer valuable insights into our design choices, and our approach advances the state-of-the-art on \emph{Vid2VidMR}.

\section{Related Work}
\label{s:rel_work}
\subsection{Video Moment Retrieval (VMR)} VMR has recently gained significant interest in the research community~\cite{DiDeMO,TVR,CTRL,moment_detr,QD-DETR,univtg,Taskweave,cg-detr,uvcom,UnLOC,UMT,mun2020LGI,chapvidmr}. Unlike video action understanding tasks such as action classification~\cite{TwoStreamNets,TSM,TPN,C3D,carreira2017quo,S3D,SlowFast,TQN} or temporal action localization~\cite{tal2019graph,talenriching,tal}, VMR focuses on identifying segments that semantically align with a broader range of queries, which may describe complex and context-specific moments. This focus on semantic alignment rather than predefined action categories enables VMR to address more flexible and nuanced video understanding scenarios, ranging from identifying specific events to capturing abstract concepts in videos. The complexity of VMR stems from the need to understand both the query's semantics and the temporal dynamics within target videos, requiring sophisticated cross-modal reasoning capabilities. Based on the query modality, existing VMR methods can be broadly grouped into (i) textual query-based approaches like Moment-DETR~\cite{moment_detr}, QD-DETR~\cite{QD-DETR}, Uni-VTG~\cite{univtg}, and (ii) video query-based approaches like GDP~\cite{GDP}, FFI+SRM~\cite{FFI-SRM}, SRL~\cite{srl}. While text-based VMR has seen extensive research attention due to the relative simplicity of handling textual queries, video query-based approaches remain less explored despite their practical importance in applications like video surveillance, sports analytics, and content recommendation systems. Our work falls under video query-based VMR (\emph{Vid2VidMR}). However, in addition to developing a new approach tailored for video query, we also adapt several text-query-based methods to make them suitable for \emph{Vid2VidMR} and compare our method against both text and video query-based methods.

\subsection{Alignment in Videos} Alignment has been a thrust area in the video understanding community. It has been applied to a wide range of video understanding tasks, including video retrieval~\cite{al_wang2021t2vlad}, procedural steps alignment~\cite{DropDTW}, action recognition~\cite{al_dogan2018neural}, anomaly detection~\cite{TCC}, movie understanding~\cite{Argaw2023LongrangeMP}, and video synchronization~\cite{al_arao2003view,al_padua2008linear}. These alignment approaches vary in their formulation, ranging from classical dynamic programming methods like Dynamic Time Warping (DTW) to more recent neural network-based techniques that learn alignment patterns from data. The effectiveness of alignment methods is particularly evident in tasks requiring the establishment of temporal correspondence between different sequences or modalities. Researchers have also explored sequence alignment for the text-VMR task, e.g., Chowdhury et al.~\cite{mithun2019weakly} used sequence alignment between CNN representation of frames and GRU representation of text query to perform text-VMR. Their approach demonstrated that even simple alignment techniques could improve localization performance by establishing correspondences between textual descriptions and visual content. Jung et al.~\cite{Jung_2025_WACV} used alignment to enhance semantic understanding between query and target video at the abstract level for the text-VMR task. However, their alignment is not at the sequence level. Despite these advances, most existing works utilize alignment in a single stage of their pipelines, limiting their ability to capture both low-level visual similarities and high-level semantic relationships simultaneously. Compared to the existing literature, \modelloc{} goes beyond traditional sequence alignment by introducing a dual-stage alignment mechanism, leveraging transformer-based feature fusion, and maintaining flexibility with multiple alignment strategies. These innovations enable it to achieve more accurate and context-aware video moment retrieval compared to existing approaches.\\

\subsection{Video Encoder Pre-training}
Recent works~\cite{achiam2023gpt,girdhar2023imagebind} have shown that pre-training on web-scale datasets can be highly effective for various downstream tasks in natural language processing as well as computer vision. The ability to transfer knowledge from large-scale datasets to specific tasks has become a cornerstone of modern deep learning approaches. The video representation learning community has also leveraged self-supervised pre-training techniques designed for training effective encoders~\cite{vul1,vul2,vul3,Kumar2024,vul5,vul6,wang2021unsupervised}, which can be finetuned for enhanced performance on downstream tasks. Unlike image-based pre-training, video pre-training must capture both spatial and temporal dynamics, making it particularly challenging yet rewarding when successful. These approaches have exploited temporal coherence and visual consistency in videos to learn meaningful features. For example, studies like~\cite{wang2021unsupervised,Kumar2024} explored the random spatio-temporal trajectory detection (constructed with a sequence of random patches) for object detection.
Inspired by these methods, which harness task-specific pre-training objectives for various video analytics tasks, we propose a novel self-supervised pre-training objective specially designed for the video-to-video moment retrieval task.

\section{Background: Dynamic Time Warping}
\label{subsec:dtw_softdtw}

Dynamic Time Warping (DTW)~\cite{saoke78} is a widely used method for aligning and comparing two sequences, such as time series, that may vary in length or exhibit local accelerations and decelerations. Unlike the Euclidean distance, which requires sequences to be of the same length and aligned in time, DTW is robust to temporal shifts and distortions, making it a popular choice in speech, gesture, and time-series analysis~\cite{soft-dtw}.

Let $\mathbf{X} = [\mathbf{x}_1, \ldots, \mathbf{x}_m] \in \mathbb{R}^{m \times d}$ and $\mathbf{Y} = [\mathbf{y}_1, \ldots, \mathbf{y}_n] \in \mathbb{R}^{n \times d}$ denote two sequences of possibly different lengths $m$ and $n$, where $d$ is the feature dimension. The pairwise dissimilarity between sequence elements is encoded in a cost matrix $\mathbf{C} \in \mathbb{R}^{m \times n}$, with entries $C_{i,j} = \delta(\mathbf{x}_i, \mathbf{y}_j)$ for a chosen distance function $\delta$ (e.g., squared Euclidean).

An alignment between $\mathbf{X}$ and $\mathbf{Y}$ is represented by a binary matrix $\mathbf{A} \in \{0,1\}^{m \times n}$, where $\mathbf{A}_{i,j} = 1$ indicates that $\mathbf{x}_i$ is matched to $\mathbf{y}_j$. The set of all valid alignments, $\mathcal{A}_{m,n}$, enforces monotonicity and continuity constraints so that alignments are contiguous and order-preserving.

DTW computes the minimum total alignment cost over all valid alignments:
\begin{equation}
   \nonumber \mathrm{DTW}(\mathbf{X}, \mathbf{Y}) = \min_{\mathbf{A} \in \mathcal{A}_{m,n}} \langle \mathbf{A}, \mathbf{C} \rangle,
\end{equation}
where $\langle \mathbf{A}, \mathbf{C} \rangle = \sum_{i=1}^m \sum_{j=1}^n \mathbf{A}_{i,j} C_{i,j}$ is the Frobenius inner product. This optimization can be efficiently solved by dynamic programming in $\mathcal{O}(mn)$ time and space.

\paragraph*{\textbf{Limitations}} DTW's flexibility enables robust sequence comparison and alignment, even when sequences are stretched or compressed in time. However, a key limitation of DTW is that it is not differentiable with respect to its inputs, as the minimum over alignments introduces discontinuities. This non-differentiability makes DTW unsuitable as a loss function in gradient-based learning frameworks.

\paragraph*{\textbf{Soft-DTW Differentiable Relaxation}} To address this, \emph{soft-DTW}~\cite{soft-dtw} replaces the hard minimum in DTW with a differentiable soft-minimum (log-sum-exp) operator, parameterized by a smoothing parameter $\gamma > 0$. The soft-minimum operator is defined as:
\begin{equation}
  \nonumber  \min\nolimits^\gamma \{a_1, \ldots, a_K\} = -\gamma \log \sum_{k=1}^K \exp\left(-\frac{a_k}{\gamma}\right).
\end{equation}
The soft-DTW cost is then given by:
\begin{equation}
   \nonumber \mathrm{soft\text{-}DTW}_\gamma(\mathbf{X}, \mathbf{Y}) = \min\nolimits^\gamma \left\{ \langle \mathbf{A}, \mathbf{C} \rangle : \mathbf{A} \in \mathcal{A}_{m,n} \right\}.
\end{equation}
As $\gamma \to 0$, soft-DTW recovers the original DTW; for $\gamma > 0$, it aggregates the costs of all possible alignments, weighting them exponentially by their cost. This smoothing ensures that soft-DTW is differentiable everywhere with respect to its inputs, enabling its use as a loss function for end-to-end training of models that output sequences.

\section{The \modelloc{} Model}
\label{sec:method}
Our objective is to temporally localize a moment in a target video $V_t$ using a query video $V_q$. We represent the target and query videos as sequences of $M$ and $N$ uniformly sampled frames, respectively. We refer to this problem as \emph{Vid2VidMR}. In this work, we present \underline{M}oment \underline{A}lignment \underline{TR}ansformer (MATR) -- a transformer-based architecture that leverages explicit alignment for precise moment localization. MATR transforms the target video into a query-aligned representation via a transformer framework to exploit finer-level correlation and dependencies for precise moment localization.

\subsection{Motivation behind MATR Architecture}
Video moments vary significantly in duration, motion patterns, and visual appearance, making their retrieval inherently challenging. Additionally, preserving the correct temporal order of events is crucial for accurate localization, as moments often involve complex interactions that unfold over time. A robust \emph{Vid2VidMR} model must, therefore, capture both high-level semantic relationships and fine-grained frame dependencies to ensure precise alignment between the query and target video.

\begin{figure*}[t!]
    \centering
      \includegraphics[width=\textwidth]{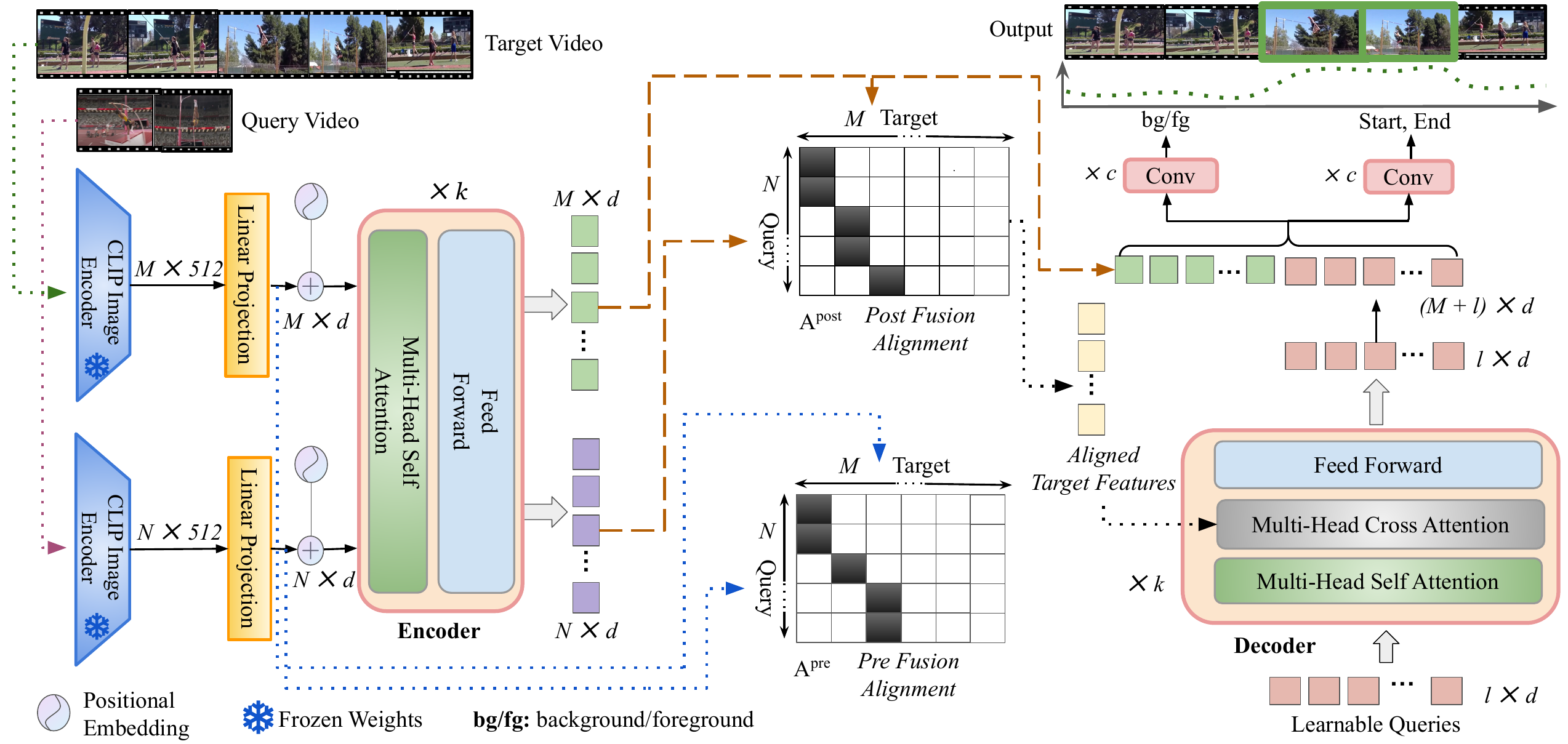}
  \caption[\underline{M}oment \underline{A}lignment \underline{TR}ansformer (MATR)]{\textbf{\underline{M}oment \underline{A}lignment \underline{TR}ansformer (\modelloc{})} Architecture. We represent the target video as a query-aligned representation capturing the required correlation and dependencies with query video at the abstract level (output of encoder) and finer level (output of decoder). These features are used to extract foreground target frames, as well as localize the relevant moment in terms of start and end times via two prediction heads. The alignment is computed using representations both before and after the encoder (shown using blue dotted and red dashed lines, respectively). \textbf{[Best viewed in color]}.}
  \label{fig:iccv25model}
  \end{figure*}

To address these challenges, \modelloc{} incorporates an explicit \emph{dual-stage sequence alignment} strategy within a transformer-based encoder-decoder framework. This strategy enables \modelloc{} to learn a \emph{query-aligned representation} of the target video by combining the abstract representation from the encoder with the refined fine-grained features from the decoder. In doing so, it facilitates precise moment localization by leveraging both global semantic alignment and fine-grained temporal dependencies between the query and target videos. Fig.~\ref{fig:iccv25model} illustrates the overall architecture of \modelloc{}, which we discuss in detail in the following.
\subsection{Architecture Details}
\noindent{\textbf{Input Representation.}}
We uniformly sample frames every 2 seconds to obtain $M$ and $N$ frames from the target and query videos, respectively. We encode these frames using a frozen CLIP~\cite{clip} (ViT-B/32) encoder along with a linear projection module. CLIP is applied separately to both videos to obtain target video embeddings $\mathrm{E}_t \in \mathbb{R}^{M \times d}$ and query video embeddings $\mathrm{E}_q \in \mathbb{R}^{N \times d}$. The linear projection module is a two-layer perceptron, each with layer normalization and dropout. It maps $512$-dimensional CLIP embeddings to $d$-dimensional outputs. The resulting projected features ($\mathrm{E}_t$ and $\mathrm{E}_q$) are concatenated along the sequence length dimension to form the input to the transformer encoder, denoted as $\mathrm{E}_c=[\mathrm{E}_t;\mathrm{E}_q] \in \mathbb{R}^{(M + N) \times d}$.

\noindent{\textbf{Encoder.}}
The combined target and query video representation ($E_c$) is processed through a series of $k$ standard Transformer~\cite{vaswani2017attention} encoder layers, each comprising a multi-head self-attention mechanism and a feed-forward network. Further, following the prior works~\cite{detr,Img_Transformer,Attn_ConvNet}, fixed positional encodings are added to the input of each attention layer to preserve temporal order. The encoder generates a fused representation $[\mathrm{E_t^g}; \mathrm{E_q^g}]$ of the target video conditioned on the query video. Here $\mathrm{E_t^g}\in \mathbb{R}^{M \times d}$ and $\mathrm{E_q^g}\in \mathbb{R}^{N \times d}$ represent the target and query parts respectively.

\noindent{\textbf{Dual-stage Sequence Alignment}}.
The transformer encoder computes effective features by performing a joint understanding (or fusion) of target and query video frames. The final goal of \emph{Vid2VidMR} is to localize a moment in the target video that semantically matches the content of the query video. In other words, we would prefer the features of the moment in the target to align strongly with the features of the query video. Therefore, we perform dual-stage sequence alignment, i.e, before and after the encoder. In our model, we incorporate several alignment techniques—including soft-DTW~\cite{soft-dtw}, TCC~\cite{TCC}, DropDTW~\cite{DropDTW}, and OTAM~\cite{CaoJCCN20}, etc. to enable robust temporal matching between query and target videos. While our framework supports all these methods, we describe the alignment process here using soft-DTW for clarity.

\noindent \underline{Pre-fusion alignment:} Before encoder fusion, given the target and query video feature sequences, $\mathrm{E_t} = [\bm{e}_1^t, \ldots, \bm{e}_M^t] \in \mathbb{R}^{M \times d}$ and $\mathrm{E_q} = [\bm{e}_1^q, \ldots, \bm{e}_N^q] \in \mathbb{R}^{N \times d}$, soft-DTW outputs a binary alignment matrix $\mathrm{A}^{\text{pre}}$ and an alignment cost matrix $\mathrm{C}^{\text{pre}}$. $\mathrm{A}^{\text{pre}}\in \{0,1\}^{M \times N}$ is an alignment matrix such that:
\begin{equation}
    \nonumber\mathrm{A}^{\text{pre}}_{i,j} =
    \begin{cases}
      1 & \text{if } \bm{e}_i^t \text{ is matched to } \bm{e}_j^q, \\
      0 & \text{otherwise.}
    \end{cases}
\end{equation}
further, soft-DTW finds the optimal alignment by minimizing the pre-fusion alignment loss \( \mathcal{L}_{\text{align}}^{\text{pre}} \), defined as:

\begin{equation}
\nonumber\mathcal{L}_{\text{align}}^{\text{pre}}= \mathrm{soft\text{-}DTW}_\gamma(\mathrm{A}^{\text{pre}}_{i,j}, \mathrm{C}^{\text{pre}}_{i,j}),
\end{equation}
\noindent such that
aligned frames in the target are contiguous. Here $\gamma$ is a smoothing factor for soft-min operator. The value of $\gamma$ in Soft-DTW is selected empirically to balance smoothness and alignment fidelity. Note that $\mathrm{C}^{\text{pre}}$ is the alignment cost matrix with elements defined by cosine similarity:
\begin{equation}
    \nonumber\mathrm{C}_{i,j} = 1 - \frac{\langle \bm{e}_i^t, \bm{e}_j^q \rangle}{\|\bm{e}_i^t\| \|\bm{e}_j^q\|}.
\end{equation}
Pre-fusion alignment enhances the semantic representation of target and query video. These enhanced representations are further processed by the encoder. The encoder outputs the fused representation of the target and query.

\noindent\underline{Post-fusion alignment:} Given the post-fusion target and query video feature sequences, $\mathrm{E_t^g} \in \mathbb{R}^{M \times d}$ and $\mathrm{E_q^g} \in \mathbb{R}^{N \times d}$, soft-DTW outputs a binary alignment matrix $\mathrm{A}^{\text{post}}$ and an alignment cost matrix $\mathrm{C}^{\text{post}}$, soft-DTW achieves optimal alignment between $\mathrm{E_t^g}$ and $\mathrm{E_q^g}$ by minimizing the post-fusion alignment loss \( \mathcal{L}_{\text{align}}^{\text{post}} \), defined as:

\begin{equation}
\nonumber\mathcal{L}_{\text{align}}^{\text{post}}= \mathrm{soft\text{-}DTW}_\gamma(\mathrm{A}^{\text{post}}_{i,j}, \mathrm{C}^{\text{post}}_{i,j}).
     \label{eq:optimal_alignment_pre}
\end{equation}
Post-fusion alignment refines target video features by leveraging fused query-target representations, ensuring fine-grained semantic matching for precise moment localization.

\noindent\textbf{Decoder}. The decoder in \modelloc{} further refines the target video representation by processing aligned query-target features, enabling fine-grained temporal matching and enhancing frame-level precision for accurate moment localization. To this end, the post-fusion alignment matrix $\mathrm{A}^{\text{post}}$ defines a contiguous sub-sequence $\mathrm{E_t^g}[s:e]$ which aligns best with the query video. These aligned target features, capturing query-aligned fine-grained information, are passed as input to the decoder for further refinement.
The decoder consists of $k$ Transformer layers, each containing a multi-head self-attention layer, a multi-head cross-attention layer, and a feed-forward network. The input to the decoder comprises fixed size $l$ learnable query vectors denoted by $\mathrm{Q} \in [q_1,\ldots, q_{l}] \in \mathbb{R}^{l \times d}$. These queries guide the extraction of refined features relevant to moment localization. As these queries are processed through each decoder layer, they are refined to capture intricate temporal dependencies, with positional encodings applied at each attention layer. The cross-attention layers enable interaction between $\mathrm{E_t^g}[s:e]$ from the encoder (which serve as keys and values) and representations of moment queries (which serve as queries). The final decoder output, $\mathrm{E_t^l} \in \mathbb{R}^{l \times d}$, provides a refined fine-grained set of target video features conditioned on the query video.

Finally, the encoder and decoder representations are combined as $\mathrm{E_f} = [\mathrm{E_t^g}; \mathrm{E_t^l}]$, making the final target video representation query-aligned. The final \emph{query-aligned representation}, combining the abstract and fine-grained semantics, serves as input for the prediction heads.

\noindent{\textbf{Prediction Heads.}}
The prediction heads consist of three ($c = 3$) sequential convolutional layers each having $d$, \( 1\times3 \) kernels followed by a ReLU activation. Finally, a sigmoid activation function is applied to produce the foreground predictions \( \hat{f}_i \). The model is trained using the following binary cross-entropy loss $\mathcal{L}_{fg}$ to distinguish foreground (relevant moment) from background predictions:
\begin{equation}
\nonumber\mathcal{L}_{fg} = -\left( f_i \log \hat{f}_i + (1 - f_i) \log (1 - \hat{f}_i) \right),
\end{equation}
where \( f_i \) is the ground truth label, with \( f_i = 1 \) and \( f_i = 0 \) indicating foreground and background, respectively.

In addition to foreground classification, the boundary prediction head is designed to estimate the start and end boundaries for the moment at every position. This head shares the same initial structure as the classification head but differs in the final output, producing two channels for left and right boundary offsets relative to the current position. Specifically, given \( \mathrm{E_f} \), the boundary prediction head outputs predicted offsets \( \hat{d}_i = \{\hat{d}_i^L, \hat{d}_i^R\} \) for each position \( i \in [0,M-1]\), where \( \hat{d}_i^L \) and \( \hat{d}_i^R \) represent the left and right boundary offsets, respectively. The boundary prediction head is trained with a combination of smooth $L1$ and generalized intersection over union (IoU) loss~\cite{rezatofighi2019generalized}, specifically applied to foreground positions (\( f_i = 1 \)), defined as:%
\begin{equation}
\nonumber\mathcal{L}_{\text{seg}} = \mathds{1}_{f_i = 1} \left[ \lambda_{\text{L1}} \mathcal{L}_{\text{L1}} \left( \hat{d}_i, d_i \right) + \lambda_{\text{IoU}} \mathcal{L}_{\text{IoU}} \left( \hat{d}_i, d_i \right) \right].
\end{equation}
Here, \( d_i \) denotes the ground truth offset. The parameters \( \lambda_{\text{L1}} \) and \( \lambda_{\text{IoU}}\) weigh the contributions of the smooth L1 and IoU losses, ensuring that boundary localization focuses on the predictions identified as foreground.

Convolutional layers operate along the temporal axis over the concatenated encoder query-conditioned and decoder-aligned target features. This preserves temporal continuity, enabling fine-grained predictions without boundary artifacts.

\noindent{\textbf{Overall Loss.}}
The overall loss function \( \mathcal{L} \) is computed as a combination of multiple objectives, accounting for alignment costs (pre- and post-fusion alignment), foreground/background classification, and boundary localization. For $S$ training samples, this multi-objective overall loss is defined as:
{\small \begin{eqnarray}
\nonumber\mathcal{L} = \frac{1}{S} \sum_{i=1}^{S} \left( \lambda_{fg}\mathcal{L}_{fg} + \lambda_{\text{seg}}\mathcal{L}_{\text{seg}}  + \lambda_{\text{align}}^{\text{pre}}\mathcal{L}_{\text{align}}^{\text{pre}} + \lambda_{\text{align}}^{\text{post}}\mathcal{L}_{\text{align}}^{\text{post}} \right),
\end{eqnarray}}
\noindent where the \( \lambda \)s control the importance of each component.

\noindent\textbf{Inference.}
During inference, given a target video \( V_t \) and a query video \( V_q \), we pass both through the model to generate foreground probabilities \( \{\hat{f}_i\}_{i=0}^{M-1} \) and boundary predictions \( \{\hat{d}_i\}_{i=0}^{M-1} \) using the two prediction heads. To handle the densely generated boundaries, while predicting \( \hat{d}_i \), we apply 1-dimensional non-maximal suppression with a threshold of 0.7, filtering out highly overlapping boundaries and producing the final set of predictions. We choose \( \hat{d}_i \) corresponding to position $i$ with highest \( \hat{f}_i \).

\begin{figure}[t!]
    \centering
    \includegraphics[width=0.9\linewidth]{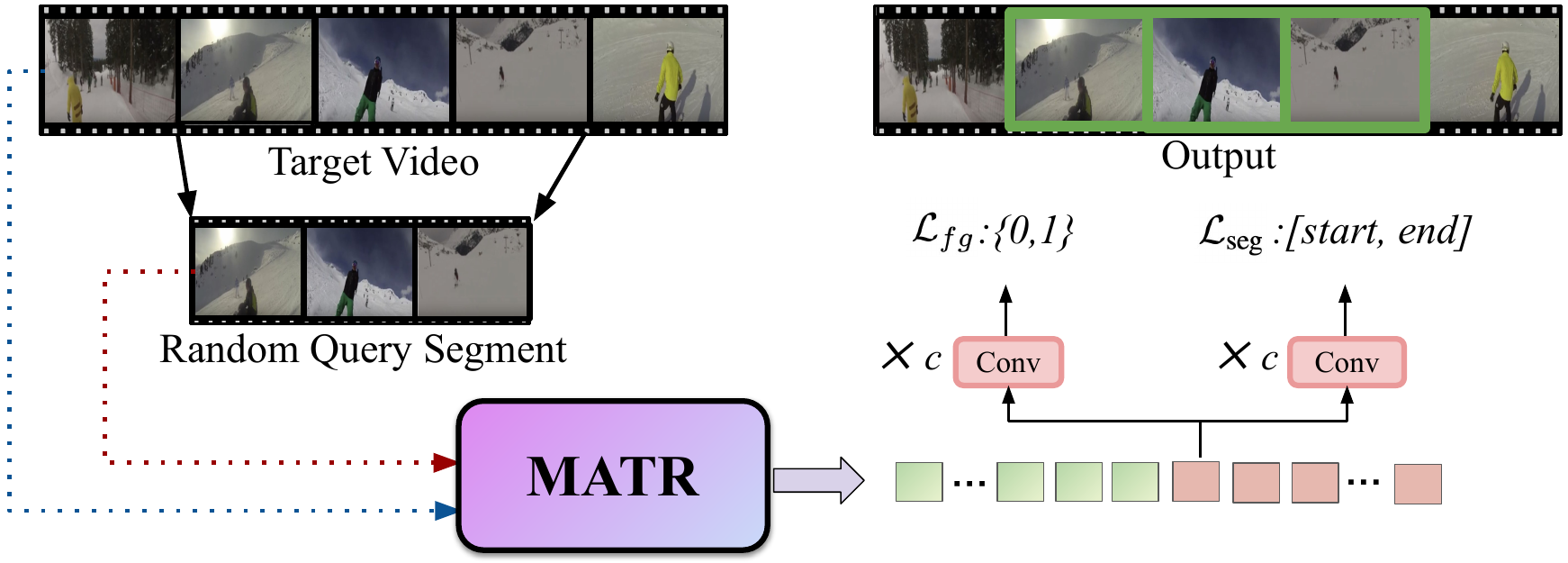}
    \caption[Self-supervised Pre-training Strategy]{\textbf{Proposed self-supervised pre-training strategy:} Given a target video, a query clip is randomly sampled from it and processed by the model. The model then predicts the boundaries of the selected clip within the target video, highlighting the corresponding frames with a green border. \textbf{(Best viewed in color)}}
    \label{fig:pre_train}
\end{figure}

\subsection{Self-supervised Pre-training Strategy}\label{s:pt_obj}
The video representation learning community has leveraged self-supervised pre-training techniques designed for training effective encoders~\cite{vul1,vul2,vul3,Kumar2024,vul5,vul6,wang2021unsupervised}, which can be finetuned for enhanced performance on downstream tasks. In this work, we introduce a self-supervised pre-training objective designed to improve temporal localization capabilities without relying on labeled data. Specifically, given a target video \( V_t \), a query clip \( V_q \) is randomly sampled from \( V_t \) as shown in Fig.~\ref{fig:pre_train}. The model is then trained to localize this query clip in \( V_t \). This pre-training objective closely aligns with the task of \emph{Vid2VidMR}. To enhance generalization capabilities and encourage the model to become robust to variations in both appearance and timing, we apply one of the following random augmentations to each of the query clips: reversing frames, adding gaussian noise, slowing down or speeding up the action, thereby doubling the number of pre-training samples. These augmentations introduce temporal and spatial variations, enabling the model to learn diverse representations.

The overall pre-training loss, \(\mathcal{L}_{pt}\), combines the foreground-background classification loss, the boundary prediction loss, pre-fusion and post-fusion alignment costs. The total loss is averaged over \( P \) pre-training samples as:
{\small\begin{eqnarray}
\nonumber\mathcal{L}_{pt} = \frac{1}{P} \sum_{i=1}^{P} \left( \lambda_{fg}^p\mathcal{L}_{\text{fg}} + \lambda_{seg}^p\mathcal{L}_{\text{seg}} + \lambda_{\text{align}}^{\text{pre}}\mathcal{L}_{\text{align}}^{\text{pre}} + \lambda_{\text{align}}^{\text{post}}\mathcal{L}_{\text{align}}^{\text{post}}\right),
\end{eqnarray}}
\noindent where the \( \lambda \)s control the importance of each component.

\section{Datasets}
We use the following two datasets to compare our proposed approach with existing \emph{Vid2VidMR} methods and other strong baselines:

\subsection{{\actdata{} Dataset~\cite{video_reloc}}}
\actdata{} is the popular benchmark dataset shared by Feng et al.~\cite{video_reloc}, based on the ActivityNet video understanding benchmark~\cite{ActivityNet} consisting of 200 action classes. The dataset is split into 160 disjoint classes for training and 20 classes each for the validation and test splits. Furthermore, the training set comprises 463K query-target video pairs, while the validation set and the test set have 829 and 978 query-target video pairs, respectively. As shown in Table~\ref{table:data_comp}, this class-disjoint split ensures that models are evaluated on unseen activity categories, providing a rigorous test of generalization capabilities.
The temporal characteristics of \actdata{} are detailed in Table~\ref{table:data_stats}, revealing that target videos have an average duration of 38.2 seconds (median 29.3 seconds), with substantial variation ranging from very short clips (1.3 seconds) to longer recordings (191.5 seconds). Query videos are significantly shorter, averaging just 8.2 seconds (median 7.2 seconds), with lengths varying from 0.5 to 57.2 seconds. This substantial difference between query and target video durations highlights the challenging nature of the temporal localization task, where models must identify short actions within much longer contexts. The extensive training set, combined with diverse activity categories and varying video lengths, makes \actdata{} a comprehensive benchmark for evaluating the generalization capabilities of \emph{Vid2VidMR} models across different types of human activities.

\subsection{{ Our proposed \datas{} Dataset}}\label{s:data}
One of the most promising applications for \emph{Vid2VidMR} lies in the area of sports analytics. Although there exist large-scale sports datasets such as Sports-1M~\cite{Sports1M}, they are primarily tailored for broader sports video classification tasks and do not contain fine-grained sports actions and moments, such as ``Cover Drive,'' ``Ducking a Bouncer,'' ``Goal Kick,'' or ``Penalty.'' Secondly, although \actdata{} covers a wide range of action categories (including sports actions), it only encompasses broader categories such as ``Playing Polo'' and does not focus on capturing higher-granularity events within a sport. Towards filling this gap, we introduce the \datas{} dataset, which consists of $\sim$770K query-target pairs annotated from 176.6 hours of complete match footage of two of the most popular sports, viz. soccer and cricket. We obtain a total of 80 full-length cricket and soccer match videos from YouTube. We then curate a list of 29 action classes comprising 13 soccer and 16 cricket actions, respectively. Given this list, we employed two annotators, each with strong knowledge of these sports, to mark the start and end timestamps for the specified actions in the videos.
We split \datas{} into training, validation, and test sets. The training split consists of approximately 750K pairs spanning 16 classes, while the validation and test splits each contain 10K pairs, covering 4 and 9 classes, respectively. Action classes are disjoint across the train, validation, and test sets, ensuring no overlap. Additionally, each split includes cricket and soccer actions for a well-rounded distribution. The statistics of the dataset are provided in Tables~\ref{table:data_comp} and \ref{table:data_stats}.

Additionally, we leverage the unlabeled videos from the Kinetics700~\cite{carreira2017quo} dataset for pre-training.

\begin{table}[!t]
\scriptsize
\centering
\begin{tabular}{l|l|c|c}
\hline
\textbf{Dataset} & \textbf{Split} & \textbf{\# Pairs} & \textbf{\# Classes} \\
\hline
\multirow{3}{*}{\datas{}} & Train & 750,393 & 16 \\
                             & Val   & 10,000 & 4 \\
                             & Test  & 10,000 & 9 \\
\hline
\multirow{3}{*}{\actdata{}~\cite{video_reloc}} & Train & 462,872 & 160 \\
                                  & Val   &  829 & 20 \\
                                  & Test  & 978 & 20 \\
\hline
\end{tabular}
\caption[Statistical Comparison of Datasets]{Statistical Comparison Between \actdata{} and the Proposed \datas{} Dataset. While \actdata{} contains greater class diversity (160 training classes), \datas{} offers substantially larger validation and test sets (10,000 pairs each versus 829/978), enabling more comprehensive evaluation of model generalization capabilities.}
\label{table:data_comp}
\end{table}

\begin{table}[!t]
\scriptsize
\centering
\begin{tabular}{l|l|c|c|c|c}
\hline
\textbf{Dataset} & \textbf{Type} & \textbf{Mean} & \textbf{Median} & \textbf{Max.} & \textbf{Min.} \\
\hline
\multirow{2}{*}{SportsMoment}
& Target Video & 98.3  & 98.2  & 208.8  & 35.4  \\
& Query Video  & 6.0   & 5.0   & 66.0   & 1.0   \\
\hline
\multirow{2}{*}{ActivityNet-VRL}
& Target Video & 38.2 & 29.3 & 191.5 & 1.3  \\
& Query Video  & 8.2  & 7.2   & 57.2  & 0.5   \\
\hline
\end{tabular}
\caption[Temporal Characteristics of Videos]{Temporal Characteristics of Videos in \datas{} and \actdata{} Datasets. SportsMoment contains significantly longer target videos (98.3s mean vs 38.2s) while maintaining shorter query segments (6.0s vs 8.2s).}
\label{table:data_stats}
\end{table}
\begin{figure}[th!]
    \centering
    \includegraphics[width=\linewidth]{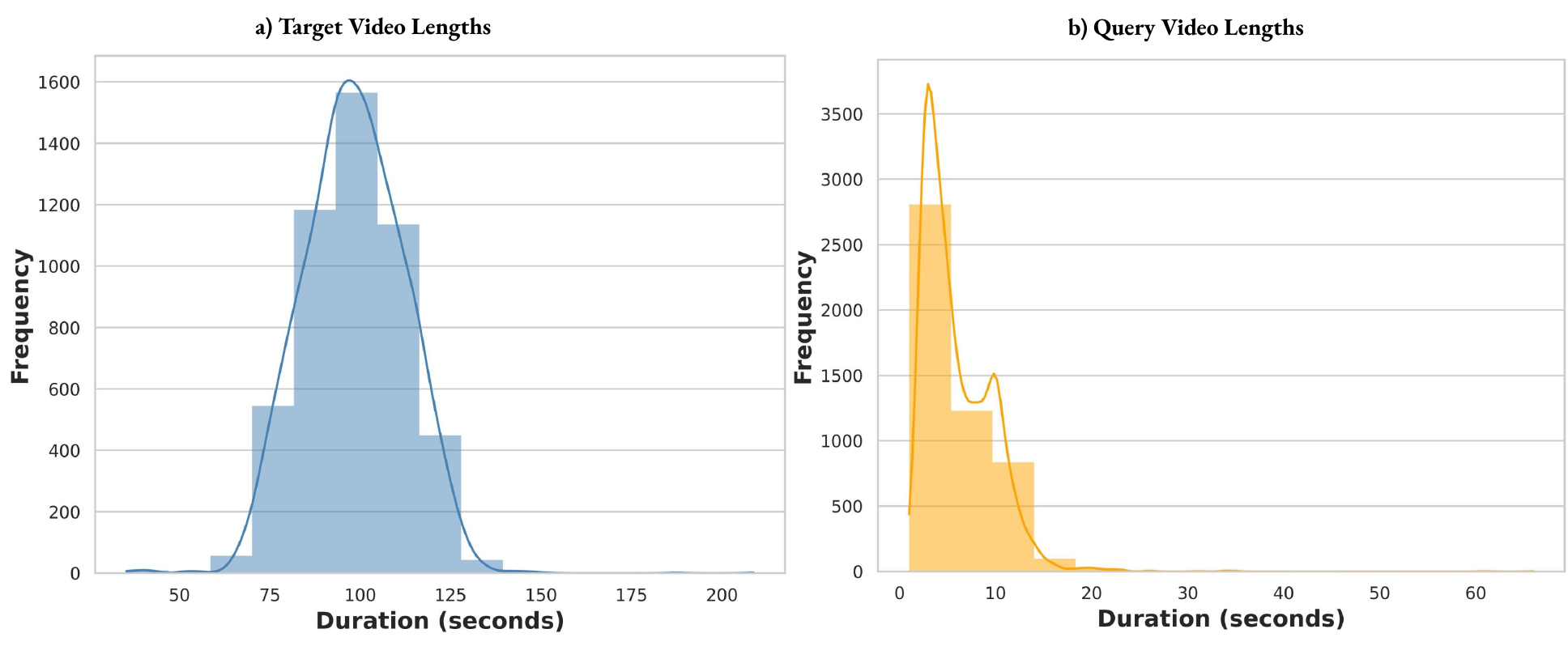}
    \caption[Distribution visualization]{We visualize the distribution of \textbf{a)} target video durations and \textbf{b)} query video durations in our proposed dataset, \datas{}.}
    \label{fig:vid_dis}
\end{figure}

\begin{figure}[th!]
    \centering
    \includegraphics[width=\linewidth]{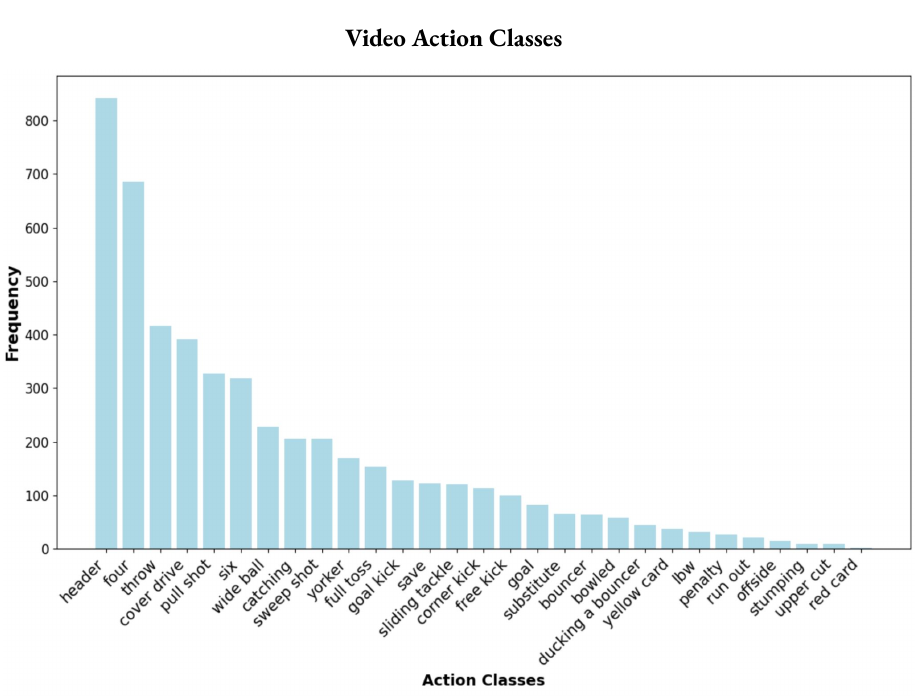}
    \caption[Frequency distribution of action classes]{Frequency distribution of action classes in the \datas{} dataset. The dataset comprises a wide range of actions across the two sports, including frequent activities such as "header," ``four," and ``throw," as well as less common ones like ``red card" and ``uppercut." The figure highlights the variation in class occurrences, reflecting the diversity and imbalance in real-world sports scenarios.}
    \label{fig:action_class_frequencies}
\end{figure}

\subsubsection*{SportsMoments Dataset Curation}
The construction of query and target videos follows a systematic process. Each annotated segment in the dataset serves as a potential query video associated with its corresponding action class. To create a target video, the annotated segment is offset by a duration ranging between 30 to 60 seconds on either side, i.e., both towards the start and end of the segment. Additionally, a scaling factor of 25\% of the annotated segment's duration is applied to this offset to account for variability in segment lengths.

During the annotation phase, to maintain data quality, we ensured that immediately overlapping segments (representing the same action occurring within 2 seconds) are merged into a single segment. Furthermore, since the dataset uses full-length matches for annotation, the number of overlapping segments across different classes is negligible, ensuring that the dataset remains clean and avoids ambiguities.

\textbf{Annotation verification and agreement.} Each annotator's work was independently reviewed by the other, and any segment on which the two disagreed was removed rather than adjudicated, so the released dataset contains only mutually confirmed annotations. Of the candidate segments produced in the first pass, 85.4\% of cricket segments and 95.0\% of soccer segments were confirmed. The disagreements are highly structured. In cricket, classes defined by \emph{outcome}, such as Four, Six, Bowled and LBW, reach 99.5\% agreement, whereas classes defined by \emph{batting technique}, such as Sweep Shot, Yorker and Cover Drive, reach only 69.3\%; almost all disagreement falls in the latter group. Outcomes are unambiguously signalled in broadcast footage through the scoreboard or the umpire signal, whereas technique classes require discriminating visually similar strokes. This is precisely the fine-grained difficulty the dataset is intended to capture, and it also locates where any residual label noise is concentrated. The higher soccer rate is consistent with this explanation, since the soccer taxonomy is largely outcome-defined. We note that, because the second annotator marked candidates as correct or incorrect rather than supplying an independent label, the figures above are verification statistics rather than a chance-corrected agreement coefficient such as Cohen's $\kappa$.

\textbf{Guarding against query--target leakage.} A query and its target are always drawn from different source videos, so a target never contains the query segment itself and retrieval cannot be solved by locating a copy of the query. In addition, the action classes in the train, validation and test splits are disjoint, so no class seen during training appears at test time. Together these two constraints ensure that the reported performance reflects generalisation to unseen actions rather than memorisation of specific footage.

Each pair in the dataset consists of a short clip acting as a query video coupled with a longer target video belonging to the same action class. For the creation of training pairs, we adopt a strategy similar to that used in \cite{video_reloc}. A query video can be paired with all target videos of the same class. Unlike \cite{video_reloc}, which generates such exhaustive pairings on the fly, we precompute and store all possible pairs during the dataset creation process.

For the test and validation splits, we curate a fixed set of 10,000 pairs for each split. These splits are designed to ensure fair representation of all classes, thereby enabling balanced evaluation across the dataset.

Compared to \datas{}, \actdata{} has a greater variety of classes in the training split, which supports more diverse class coverage. However, \datas{} has significantly more pairs overall in the validation and test splits, which provide a broader evaluation framework for our tasks. The target videos in \datas{} benchmark range from a minimum of 35.4 seconds to an upper bound of 208.8 seconds with the mean and median lengths being 98.3 and 98.2 seconds respectively. The query video is constrained to be atleast 1 second in duration as we believe this to be the minimum duration in which an action/event can be depicted coherently. Maximum length of a query video in our proposed dataset is 66 seconds consisting of overlapping scenes of the same action including replays from different perspectives. Mean and Median of the query videos are 6.0 seconds and 5.0 seconds, respectively.  For \actdata{}, the training split has 462,872 pairs covering 160 classes, while the validation and test splits have 829 and 978 pairs, respectively, each covering 20 classes.

We collected 64 cricket and 16 soccer full length matches collectively spanning 176.6 hours. We were able to extract 4999 segments from all the matches, with the average length of each segment being 6.0 seconds. Out of these 4999 segments 2925 fall under cricket and 2074 belong to soccer. Each segment is annotated with its start and end timestamp. The number of segments per-class is shown in Figure~\ref{fig:action_class_frequencies}. Table~\ref{table:data_comp} shows a comparison between the proposed \datas{} and the repurposed ActivityNet~\cite{video_reloc}.

\begin{table*}[th!]
\scriptsize
\centering
\begin{tabular}{|c|c|c|c|c|c|}
\hline
4a5Oou5u3io& GrIOmtbyHNc& 5j2tQfj2Osk& mo3H\_qusvxI& RpmSeHG8SJ0& cpU0GSxliSU\\
\hline
EOvpUWZC6ng& Txq8RoclyKM& -vYVBUyEuAA& IvjBlpb8GTA& ukuNlWxcR2k& LmqBbPwCy8o\\
\hline
zuPuDzEWI-A& OFbyNU6UQQs& iPXz5XWfj1k& IA5BNsNu6fw& 9sRb61qsoLs& X0we8220k74\\
\hline
MbdWVX2jUhA& 04vWaGaqw-4& 92s9HCGG4vQ& BMXfphiHj1s& 1yHWGw8DH4A& shyBGceNASI\\
\hline
YqKYpgZ9FWU& GxO1tlZKa8E& qElsVosGiXA& zUTUA0TKvfQ& G4vxbWF79f4& v9Akttif8gY\\
\hline
Mtnmvuhs2RA& 3K90O0Y9NZc& 9gyv2xh7qQw& 1-SCubr2yc4& ezQUPi74IIQ& cTqY53zWypk\\
\hline
qQfuOCns4P4& iGGElV40TcU& X6VNTk-Vl5w& iFYm\_ogzmdA& 6j1Oj5N7kwE& WkEvdVImkec\\
\hline
wTN82hUkhWk& rpPCjyVfQOo& L7olqH38dzw& farmGHpVCek& S825iLaZQhg& T-c3B0lJJvE\\
\hline
ku0Y8TA8L44& bSRB7ysDRQw& 90yyJvfbwAs& 9oRmmeKJ6WM& 6f9v2z5HYX4& eeaHNmIuP3c\\
\hline
5OGkvFUl3G8& ePmiyHWxMDk& CuEC8Twrcgo& nc9nA8fG87Q& 9dy7kpALSAE& EedISWx\_fCo\\
\hline
9lz5VV-8XEc& qEgOfHT\_MnI& snM7xGdWHSc& Us\_6Nzj3Fic& H7q7q1IwEY4& xYbklmu4ZsE\\
\hline
stWSFacouPM& RRk3S82oH9M& XJgOiFkOgzo& RKP\_Mg4piUA& 2rLIjqEM3Nw& rfl6elqYCKA\\
\hline
cdkiRWd15RA& FXF7c2LyGDo& hUBLgzSEaS4& X9ny1p0D61I& qWlfNQRd0uY& tDeoIdYvh6g\\
\hline
W\_j6PHiin\_0& fWCU\_oJyD38&&&&\\
\hline
\end{tabular}
\caption{YouTube IDs used for \datas{} dataset creation. We used 80 videos for dataset curation.}
\label{tab:video_ids}
\end{table*}

\subsubsection*{YouTube IDs used in \datas{}}
We also release the YouTube IDs of videos used in our dataset, 80 in total as shown in Table~\ref{tab:video_ids}.
For cricket, we have a total of 64 videos, 26 of which are full length twenty over games, 9 are fifty over games, 26 of which are highlights of different matches, one of which is a part of a full length match and 3 videos are that of a single day Test-match innings. The remaining 16 matches are that of full length football matches from various leagues.

\section{Experiments and Results}
\label{s:exps}

\subsection{Baselines}
\label{sec:iccv_base}
We experiment with an extensive set of strong baselines grouped into four categories as follows.

\subsubsection{\textbf{Fully-supervised \emph{Vid2VidMR} methods}} We compare with \emph{Vid2VidMR} methods like CGBM~\cite{video_reloc}, GDP~\cite{GDP}, SRL~\cite{srl}, FFI+SRM~\cite{FFI-SRM}, SST~\cite{SST} and Video-level match~\cite{video_reloc}. These approaches represent traditional fully-supervised methods that have been specifically designed for video-to-video moment retrieval tasks. Many of these methods rely on sophisticated attention mechanisms to establish correspondences between query and target video features, but they often lack the explicit alignment strategies that our approach incorporates. Their performance varies significantly based on the feature extraction backbones and matching strategies employed, with more recent methods generally achieving better results through improved feature representation learning. Further, Huo et al.~\cite{FFI-SRM} adapt text-VMR methods like VSLNet~\cite{VSLNet}, MABAN~\cite{MABAN} and 2D-TAN~\cite{2D-TAN} by replacing their text encoders with a C3D~\cite{C3D} feature extractor to extract query video features. This adaptation enables text-based methods to process video queries, providing a more direct comparison with our approach. While these adaptations demonstrate reasonable performance, they often struggle with effectively modeling the temporal dynamics present in video queries, as they were originally designed for a different modality of text. This highlights the need for approaches like ours that are specifically engineered to handle the unique challenges of video-to-video correspondence. All of these models have been trained on \actdata{}.

\subsubsection{Vision-Language Models (VLMs)}
VLMs like Video-LLaMa~\cite{videollama}, Video-LLaVa~\cite{videollava}, and TimeChat~\cite{Ren2023TimeChat} have shown promising results for multimodal text-video tasks like Video VQA, captioning, etc. These foundation models leverage large-scale pre-training on diverse multimodal datasets to develop general representations that can be applied to various downstream tasks with minimal adaptation. Their ability to understand complex semantic concepts across modalities makes them attractive candidates for video understanding tasks. Despite their impressive performance on various video understanding tasks, these models face inherent limitations when applied to video-to-video retrieval. The primary challenge stems from their architectural design, which is optimized for text-video interactions rather than video-video comparisons that require fine-grained temporal alignment capabilities. Unfortunately, they have not been pre-trained with video to video aligned data. This lack of video to video pretraining significantly impacts their ability to perform well on our task without substantial modification. To this end, we first represent the query video using a state-of-the-art captioner, i.e., mPLUG-OWL~\cite{mplugowl}. This conversion from video to text enables these text-centric models to process video queries indirectly. However, this approach inevitably loses some of the rich temporal and visual information present in the original query video, potentially limiting performance on tasks that require precise temporal alignment. Next, the VLMs are zero-shot prompted to generate start and end moment timestamps given the query caption and target video tokens.

\subsubsection{Text-VMR methods} Moment-DETR~\cite{moment_detr}, QD-DETR~\cite{QD-DETR}, CG-DETR~\cite{cg-detr}, BM-DETR~\cite{Jung_2025_WACV} and UniVTG~\cite{univtg} have been originally proposed for text-VMR. These methods represent the current state-of-the-art approaches for textual moment retrieval and serve as strong baselines when adapted to video queries. Each of these approaches employs a different architectural design, with more recent methods like UniVTG leveraging unified vision-text encoders for improved cross-modal understanding, while earlier methods like Moment-DETR primarily rely on transformer-based encoders with cross-attention mechanisms. We compare with five variants of these methods. The zero-shot variant (a) uses caption from mPLUG-OWL~\cite{mplugowl} to represent the query video, and leverages the pretrained checkpoints. This approach tests how well existing models can transfer to our task without any task-specific training. Variants (b) and (c) both use a caption from mPLUG-OWL~\cite{mplugowl} to represent the query video, but train a randomly initialized checkpoint (variant b) or finetune the pre-trained checkpoint (variant c), respectively. These variants evaluate whether task-specific training improves performance when using text representations of video queries. Though textual representations capture semantic content, they inevitably lose important visual and temporal nuances present in the original query videos, potentially limiting the upper bound on performance for these text-based variants. Variants (d) and (e) are equivalent to (b) and (c), where their CLIP text encoder is replaced by CLIP ViT/B-32 vision encoder, and therefore take the query video directly as input along with the target video. These latter variants provide the most direct comparison to our approach as they process video queries without converting to text intermediaries.

\subsubsection{Image-VMR methods} Our work focuses on video moment retrieval using `video' queries. As a single image query may not effectively capture the temporal aspects of a video query, this comparison may not be appropriate. The fundamental limitation here is that static images lack the ability to represent dynamic actions and temporal evolution that characterize video content. This limitation becomes particularly evident in scenarios involving complex activities with significant motion patterns or when distinguishing between visually similar actions that differ primarily in their temporal execution. However, we still design baselines by representing the video by its key-frame (more precisely, the middle frame) using Text-VMR methods, namely Moment-DETR, QD-DETR, and UniVTG. This simplified representation allows us to quantify the value of temporal information in query videos compared to static image representations.

\begin{table*}
\scriptsize
    \centering
    \begin{tabular}{ll ll cccc}
    \toprule
       &\multicolumn{3}{c}{} & \multicolumn{2}{c}{\actdata{}~\cite{ActivityNet}} & \multicolumn{2}{c}{\datas{}} \\
       \cmidrule(r){5-6} \cmidrule(r){7-8}
        &\multirow{1}{*}{Methods}   & Variants & Query Modality &  mIoU & R@1 & mIoU &  R@1 \\
        \midrule
        \multirow{10}{*}{\rotatebox{90}{Fully-supervised }}&Random  & - & Video & 7.3 & 16.2 & - & -\\
        &Video Match & -& Video &  12.4 & 24.3 & - & -\\
        &SST  & - & Video& 17.1 & 33.2 & - & -\\
        &CGBM  & - & Video& 25.7 & 43.5 & - & -\\
        &GDP & - & Video & 27.8 & \fbox{44.0} & - & -\\
        &SRL  & - & Video & 40.6 & 29.3 & - & -\\
        &2D-TAN  & - & Video & 45.3 & 39.6 & - & -\\
        &MABAN  & - & Video & 42.8 & 37.5 & - & -\\
       & VSLNet  & - & Video & 27.2 & 43.8 & - & -\\
        &FFI+SRM  & - & Video & \fbox{48.7} & 40.6 & - & -\\
        \midrule
        \multirow{4}{*}{\rotatebox{90}{VLMs}}&Video-LLaVA & - & Video & 15.1& 14.7 & 13.8 & 11.9\\
       & Video-LLaMA & - & Video & 14.7& 13.9 & 12.5 & 11.2\\
        &Video-LLaMA2 & -& Video & 17.6  & 15.2 & 14.7 & 13.4\\
       & TimeChat  & -& Video & 26.4 & 23.8 & 22.6 & 21.3\\
        \midrule

        \multirow{3}{*}{\rotatebox{90}{I-VMR}}&Moment-DETR & - & Image & 35.6 & 32.5 & 25.2 & 18.5\\
       & QD-DETR & - & Image & 37.2 & 38.1 & 27.7 & 20.7\\
        &UniVTG & -& Image & 38.8 & 42.0 & 34.6 & 37.2\\
        \midrule

\multirow{25}{*}{\rotatebox{90}{Text-VMR }}&\multirow{6}{*}{Moment-DETR } & (a) Zero-shot & Text & 28.2 & 22.8 & 3.7 & 0.9\\
&& (b) Trained & Text& 37.1 & 35.8 & 31.4 & 29.2\\
&& (c) Finetuned & Text & 40.0 & 38.9 & 35.8 & 34.0\\
&& (d) Trained & Video & 35.9 & 34.7 & 28.7 & 24.2\\
&& (e) Finetuned & Video & 40.0 & 39.9 & 30.4 & 25.1\\
\cmidrule(r){2-8}
&\multirow{6}{*}{QD-DETR } & (a) Zero-shot & Text & 25.4 &  22.2 & 6.3 & 2.6\\
&& (b) Trained & Text & 38.4 & 39.0 & 29.8 & 27.2\\
&& (c) Finetuned & Text & 41.6 & 45.2 & 36.0 & 33.5\\
&& (d) Trained & Video & 39.7 & 41.0 & 27.2 & 25.1\\
&& (e) Finetuned & Video & 42.5 & 42.7 & 35.4 & 30.6\\
\cmidrule(r){2-8}
&\multirow{6}{*}{UniVTG } & (a) Zero-shot & Text& 32.4 & 26.7 & 11.4 & 6.0\\
&& (b) Trained & Text & 43.8 & 45.6 & 41.5 & 36.5 \\
&& (c) Finetuned & Text & 45.8 & 46.4 & 44.8 & 39.2 \\
&& (d) Trained & Video & 48.4 & 49.8 & 43.2 & 39.5\\
&& (e) Finetuned & Video & 49.1 & 50.7 & 43.6 & 41.8 \\
\cmidrule(r){2-8}
&\multirow{6}{*}{CG-DETR } & (a) Zero-shot & Text& 25.9 & 21.9 & 3.3 & 1.6\\
&& (b) Trained & Text& 39.4 & 38.4 & 35.1 & 31.3\\
&& (c) Finetuned & Text& 41.0 & 43.2 & 35.4 & 32.5\\
&& (d) Trained & Video& 40.0 & 41.7 & 36.6 & 35.1\\
&& (e) Finetuned & Video& 40.1 & 41.7 & 37.2 & 34.8\\

\cmidrule(r){2-8}
&\multirow{6}{*}{BM-DETR} & (a) Zero-shot & Text & 30.6 & 25.1 & 8.3 & 5.7\\
&& (b) Trained & Text & 44.2 & 46.1 &  42.6& 38.4\\
&& (c) Finetuned & Text & 46.1 & 46.7 & 45.2 & 39.6\\
&& (d) Trained & Video &  50.7 & 53.2 & 45.6 & 44.2\\
&& (e) Finetuned & Video & \underline{50.9} & \underline{53.7} & \underline{47.6} & \underline{44.9}\\
        \midrule
\multirow{10}{*}{\rotatebox{90}{Ours}}&\multirow{10}{*}{\modelloc{}}
& Trained (CLIP + soft-DTW)  & Text& 49.3 & 50.1 & 48.4  & 47.9\\
& &Finetuned (CLIP + soft-DTW)& Text & 49.7& 50.9& 49.1 & 48.3\\

& &Trained (CLIP + soft-DTW) & Video& 53.2 & 54.8 & 56.2 & 52.7\\
& & Finetuned (CLIP + soft-DTW) & Video& 56.8 & 57.1 & 59.2 & 56.5\\
& &Trained (CLIP + OTAM) & Video& 57.1 & 57.4 & 61.4 & 57.9 \\
& &Finetuned (CLIP + OTAM) & Video& 58.2 & 58.7 & 62.2 & 58.6 \\
& &Trained (Slow-Fast + soft-DTW) & Video& 54.7 & 55.9 & 58.3 & 55.4  \\
& &Finetuned (Slow-Fast + soft-DTW) & Video& 57.6 & 58.3 & 60.4 & 58.2  \\
& &Trained (Slow-Fast + OTAM) & Video&  58.3& 57.9 & 62.2 & 59.1\\
& &Finetuned (Slow-Fast + OTAM) & Video&  \textbf{60.1}& \textbf{59.4} & \textbf{63.9} & \textbf{61.3}\\

       \bottomrule
    \end{tabular}
    \caption[Main Table]{Comparing \modelloc{} with fully-supervised \emph{Vid2VidMR} methods, VLMs, image-VMR methods, and variants of Text-VMR methods. Trained implies random initialization. Finetuned implies initialization using a pre-trained (respective to each architecture) checkpoint. Query video can be represented using text captions or as video itself. Since \modelloc{} is inherently designed to take video query as input, experimenting with text captions is not needed. For more details of baselines and their variants, please refer to Section~\ref{sec:iccv_base}. Results for the overall best, best among implemented baselines and previously reported SOTA are highlighted in bold, underline and box, respectively.}
    \label{tab:main_table}
\end{table*}

\subsection{Implementation Details}
\label{s:imple-det}
We choose a hidden dimension of $1024$ with $k=4$ layers in both encoders and apply a dropout rate of $0.1$ and  $0.5$ within the transformer and linear projection layers, respectively. Model weights are initialized using Xavier initialization~\cite{xavier}. To optimize the model parameters, we utilize the AdamW optimizer~\cite{adam} with an initial learning rate of 1e-4 and a weight decay of 1e-4. Training was done on two NVIDIA A6000 GPUs. We trained our model for 200 epochs with a batch size of 1200 while using \actdata{}. For \datas{}, we used 40 epochs with a batch size of 40. We set the number of learnable queries $l=10$ and all $\ lambda$'s in both pre-training and finetuning losses to 1.

\begin{table}[!t]
\scriptsize
    \centering
\begin{tabular}{l c c c c c}
\toprule
       & \multicolumn{2}{c}{Recall} & \multicolumn{3}{c}{mAP} \\
       \cmidrule(r){2-3} \cmidrule(r){4-6}
Method & R1@0.5 & R1@0.7 & mAP@0.5 & mAP@0.75 & Avg. \\
\midrule
Moment-DETR  & 53.9 & 34.8 & - & - & 32.2 \\
QD-DETR  & 62.7 & 46.7 & 62.2 & 41.8 & 41.2 \\
UniVTG  & 59.7 & 45.8 & 61.9 & 41.9 & 38.8 \\
BM-DETR & 63.6 & 46.8 & 63.7 & 43.5 & 44.2\\
CG-DETR & \textbf{67.4} & 52.1 & \textbf{65.6} & \textbf{45.7} & \textbf{44.9} \\
Ours (\modelloc{}) & 64.3 & \textbf{54.2} & 61.1 & 44.0 &  42.5\\
\bottomrule
\end{tabular}

\caption[Text-VMR Comparison]{Comparative Analysis of \modelloc{} versus State-of-the-Art Text-VMR Models on the QVHighlights Dataset. While primarily designed for video queries, our approach demonstrates competitive performance on text queries. }
\label{tab:tvmr_table}
\end{table}

\begin{table}[!t]
\scriptsize
    \centering
\begin{tabular}{c c c c c c}
\toprule
 \multicolumn{2}{l}{}  & \multicolumn{2}{c}{\actdata{}}  & \multicolumn{2}{c}{\datas{}}\\
 \cmidrule(r){3-4} \cmidrule(r){5-6}
Pre-Fusion  & Post-Fusion & mIoU & R@1 & mIoU & R@1 \\
\hline
\xmark & \xmark  & 49.7 & 50.2 & 49.1& 46.6\\
\xmark & \checkmark & 52.3 & 52.9 & 53.1 & 48.2\\
\checkmark & \xmark & 50.4 & 51.2 & 52.4 & 46.7\\
\checkmark & \checkmark & \textbf{53.2} & \textbf{54.8} & \textbf{56.2} &\textbf{52.7} \\
\hline
\end{tabular}
\caption[Impact of Pre-Fusion and Post-Fusion]{Impact of Pre-Fusion and Post-Fusion Alignment Strategies on \modelloc{} Performance. This study is done with CLIP features, soft-DTW alignment and without pre-training.}
\label{tab:abl_align}
\end{table}

\begin{table}[!t]
\scriptsize
    \centering
\begin{tabular}{c c c c c c}
\toprule
 \multicolumn{2}{l}{}  & \multicolumn{2}{c}{\actdata{}}  & \multicolumn{2}{c}{\datas{}}\\
 \cmidrule(r){3-4} \cmidrule(r){5-6}
Pre-training & Augmentation & mIoU & R@1 & mIoU & R@1\\
 \midrule
   \xmark & - & 53.2  &  54.8 & 56.2 & 52.7\\
  \checkmark & \xmark & 54.1  &  55.6 & 56.9 & 53.6\\
    \checkmark & \checkmark & \textbf{56.8} & \textbf{57.1} & \textbf{59.2} & \textbf{56.5}\\
 \hline
\end{tabular}
\caption[Impact of Pre-training]{Impact of Pre-training and Data Augmentation on \modelloc{} Performance Across both Datasets. This study is done with CLIP features and soft-DTW alignment.}
\label{tab:pre-train}
\end{table}

\subsection{Results and Discussion}

Table~\ref{tab:main_table} shows our main results where we compare \modelloc{} with fully-supervised \emph{Vid2VidMR} methods, VLMs, image-VMR methods, and five variants of each of the Text-VMR methods. As mentioned before, these variants correspond to zero-shot, finetuned, or trained setups. As usual, the trained setup implies random initialization, while the finetuned setup implies initialization using a pre-trained checkpoint (different for each architecture). Query video can be represented using text (T) captions or as the video (V) itself. For MATR, we show results with various video feature extraction backbones and sequence alignment methods. Following previous work, we report the standard mean Intersection over Union (mIoU) and Recall@1 IoU=0.5 metrics.

\subsubsection{Comparison with fully-supervised \emph{Vid2VidMR} methods} For a fair comparison, we directly use the reported results from original papers for fully-supervised methods on \actdata{}. Consequently, we do not report their results on \datas{}. As shown in Table~\ref{tab:main_table}, \modelloc{} outperforms the best method, FFI+SRM, with an 11.4\% gain in mIoU and achieves a 15.4\% improvement in R@1 over GDP, which had the highest recall. These results underscore \modelloc{}'s superior ability to accurately localize moments using video queries, validating its effectiveness across diverse and challenging benchmarks.

\subsubsection{Comparison with VLMs}
Table~\ref{tab:main_table} shows that our method \modelloc{} outperforms all four VLM-based baselines by a significant margin. This performance gap indicates the limitations of general-purpose vision-language models when applied to specialized temporal tasks without task-specific architectures. In comparison, the best-performing baseline, TimeChat~\cite{Ren2023TimeChat}, achieves 26.4\% mIoU and 23.8\% R@1 on \actdata{}, and 22.6\% mIoU and 21.3\% R@1 on \datas{}. The performance difference between TimeChat and our method (33.7\% mIoU difference on \actdata{} and 41.3\% mIoU on \datas{}) highlights the importance of architectures specifically designed for temporal reasoning and alignment. Other methods, such as Video-LLaVA~\cite{videollava}, Video-LLaMA~\cite{videollama}, and Video-LLaMA2~\cite{videollama2}, perform significantly worse. Despite the impressive general capabilities of these models in video understanding tasks, their performance on the fine-grained temporal localization required for moment retrieval remains limited. These results show that task-specific models for \emph{Vid2VidMR} are significantly better.

\subsubsection{Comparison with Text-VMR Methods}
Table~\ref{tab:main_table} shows that amongst all the \emph{five variants}, the finetuned variants are typically better than the trained variants, i.e., variant (c) is better than (b), and (e) is better than (d). On both \actdata{} and \datas{}, BM-DETR achieves the best results among variants that use video queries. However, our \modelloc{} model outperforms BM-DETR by 9.2\% in mIoU and 5.7\% in R@1 on \actdata{}. On \datas{}, \modelloc{} shows a further improvement, surpassing BM-DETR by 16.3\% in mIoU and 16.4\% in R@1. These gains highlight the superior performance of \modelloc{} compared to the text-VMR methods when used with video query input. When text captions are used to represent the query, we observe similar trends as in the video query input case. Among the baselines, BM-DETR performs the best. However, \modelloc{} outperforms it by 3.6\% in mIoU and 4.2\% in R@1 on \actdata{}. On \datas{}, \modelloc{} achieves 3.9\% higher mIoU and 8.7\% higher R@1 than BM-DETR. These results demonstrate the superior performance of \modelloc{} compared to the text-VMR methods when used with text query input for \emph{Vid2VidMR}. One surprising finding is that Moment-DETR and QD-DETR perform worse with video queries compared to text queries, even after fine-tuning, especially for \datas{}. On the other hand, UniVTG and CG-DETR show expected trends, i.e., performance is better with video queries than with text queries for these two models.

\subsubsection{Comparison with Image-VMR}
Table~\ref{tab:main_table} shows that the best-performing approach, UniVTG~\cite{univtg}, achieves a mIoU of 38.8\% and 42.0\% R@1 on ActivityNet-VRL and 34.6\% mIoU and R@1 37.2\%  on \datas{}. This performance gap is particularly evident when comparing across different evaluation metrics and datasets, highlighting the inherent limitations of image-based representations for temporal tasks. Although this is comparable to previous fully supervised methods, it still lags significantly behind our \modelloc{}. The substantial performance difference (21.3\% mIoU and 17.4\% R@1 on \actdata{}, 29.3\% mIoU and 24.1\% R@1 on \datas{}) demonstrates that video queries contain crucial motion patterns and temporal context that static images simply cannot capture. This further goes to show the importance of temporal dynamics captured by video queries as compared to static image queries.

\subsubsection{Comparison on Text-VMR task}
Experimental results in Table~\ref{tab:tvmr_table} demonstrate that our proposed model achieves competitive performance on the Text-VMR task on the QVHighlights dataset, particularly excelling in temporal boundary precision with the highest R1@0.7 score of 54.2\%. While CG-DETR maintains superiority in most metrics (R1@0.5, mAP@0.5, mAP@0.75, and average mAP), our approach shows a notable 2.1\% improvement over CG-DETR in R1@0.7, suggesting better precision in temporal localization. The trade-off appears to be a slight decrease in mAP metrics compared to CG-DETR, though our model still outperforms earlier approaches like Moment-DETR and competes well with other recent methods.

\subsubsection{Advantage of Dual Sequence Alignment}
Table~\ref{tab:abl_align} shows the impact of different alignment strategies in \modelloc{} on both datasets. We evaluate the effect of incorporating alignment at the pre-fusion and post-fusion stages. The pre-fusion alignment helps establish initial cross-modal correspondences, while post-fusion alignment refines these relationships after contextual encoding. Our results indicate that the combined use of both pre-fusion and post-fusion alignment achieves the best performance, suggesting that alignment is important to ensure both before and after the encoder. This dual alignment strategy enables the model to progressively refine cross-modal relationships at multiple levels of abstraction. Models with alignment either before (pre) or after (post) show moderate gains over neither. The incremental improvements demonstrate that each alignment stage contributes uniquely to the model's ability to localize moments. Removing post-fusion alignment hurts more than removing pre-fusion alignment, suggesting that post-fusion alignment is more important. This indicates that alignment after contextual encoding captures more semantically meaningful relationships between modalities. Omitting alignment entirely results in the lowest scores. The significant performance drop (3.5\% mIoU on \actdata{} and 7.1\% mIoU on \datas{}) when removing both alignment stages emphasizes their complementary nature. This highlights the importance of dual alignment for effectively capturing relevant temporal information for video query-based moment localization.

\subsubsection{Effect of Pre-training} \label{s}
We analyze the impact of pre-training and data augmentation on model performance in Table~\ref{tab:pre-train}. On \actdata{}, without pre-training, the model achieves 53.2\% mIoU and 54.8\% R@1. Adding pre-training without augmentation yields 54.1\% mIoU and 55.6\% R@1, demonstrating the benefit of leveraging large-scale video-text datasets prior to task-specific fine-tuning. The best results are obtained by using pre-training with augmentation, leading to a significant boost to 56.8 mIoU and 57.1\% R@1, a 3.6\% improvement in mIoU over the baseline. Similar trends are observed on the \datas{} dataset, where the full configuration achieves 59.2\% mIoU and 56.5\% R@1, confirming the generalizability of our approach across different datasets. During augmentation for each sampled clip, one of the following transformations is applied: reversing frames, adding Gaussian noise, slowing down the action, or increasing the action speed. These augmentations enhance the model's robustness to temporal variations and noise, which are common challenges in real-world video understanding scenarios. The consistent performance improvements across both datasets highlight the importance of our two-stage training strategy in developing robust video moment retrieval systems.

\begin{table}[!t]
\scriptsize
    \centering
\begin{tabular}{l cccc}
\toprule
 \multicolumn{1}{l}{}  & \multicolumn{2}{c}{\actdata{}}  & \multicolumn{2}{c}{\datas{}}\\
 \cmidrule(r){2-3} \cmidrule(r){4-5}
 & mIoU & R@1 & mIoU & R@1\\
\midrule
Pre-fusion alignment ($\mathrm{A^{\text{pre}}}$) & 34.1 & 32.6 & 31.6 & 29.8 \\
Post-fusion alignment ($\mathrm{A^{\text{post}}}$) & 38.3 & 35.2& 36.4 & 33.5\\
Prediction-heads & \textbf{56.8} & \textbf{57.1} & \textbf{59.2} & \textbf{56.5}\\
 \bottomrule
\end{tabular}
\caption[Advantage of Predicting Using Prediction Heads]{Advantage of predicting using prediction heads on the decoder. This study is done with CLIP features, soft-DTW alignment and with pre-training.}
\label{tab:pred_study}
\end{table}

\begin{table}[!t]
\scriptsize
    \centering
\begin{tabular}{l c c c c}
\toprule
 \multicolumn{1}{l}{}  & \multicolumn{2}{c}{\actdata{}}  & \multicolumn{2}{c}{\datas{}} \\
 \cmidrule(r){2-3} \cmidrule(r){4-5}
Alignment & mIoU & R@1 & mIoU & R@1 \\
 \midrule
    None  &  50.4 &  51.8 & 52.1 & 48.6 \\
    TCC \cite{TCC}  & 52.4 & 52.7  &  53.9 & 52.8 \\
    Needleman-Wunsch \cite{NW}  & 50.8& 51.2& 52.6 & 49.5\\
    LCSS~\cite{LCSS} & 50.9 & 52.6& 53.7 & 49.6 \\
    Drop-DTW~\cite{DropDTW} & 54.2 & 54.7 & 55.6 & 54.7 \\
    Soft-DTW~\cite{soft-dtw}  & 56.8 & 57.1 & 59.2  & 56.5 \\
    OTAM~\cite{CaoJCCN20} & \textbf{58.2} & \textbf{58.7} & \textbf{62.2} & \textbf{58.6} \\
 \hline
\end{tabular}
\caption[Performance Comparison of Sequence Alignment Methods]{Performance Comparison of Sequence Alignment Methods. This comparison is done with CLIP features.}
\label{tab:align_loss_abl}
\end{table}

 \begin{figure*}[!t]
    \centering
      \includegraphics[width=0.93\linewidth]{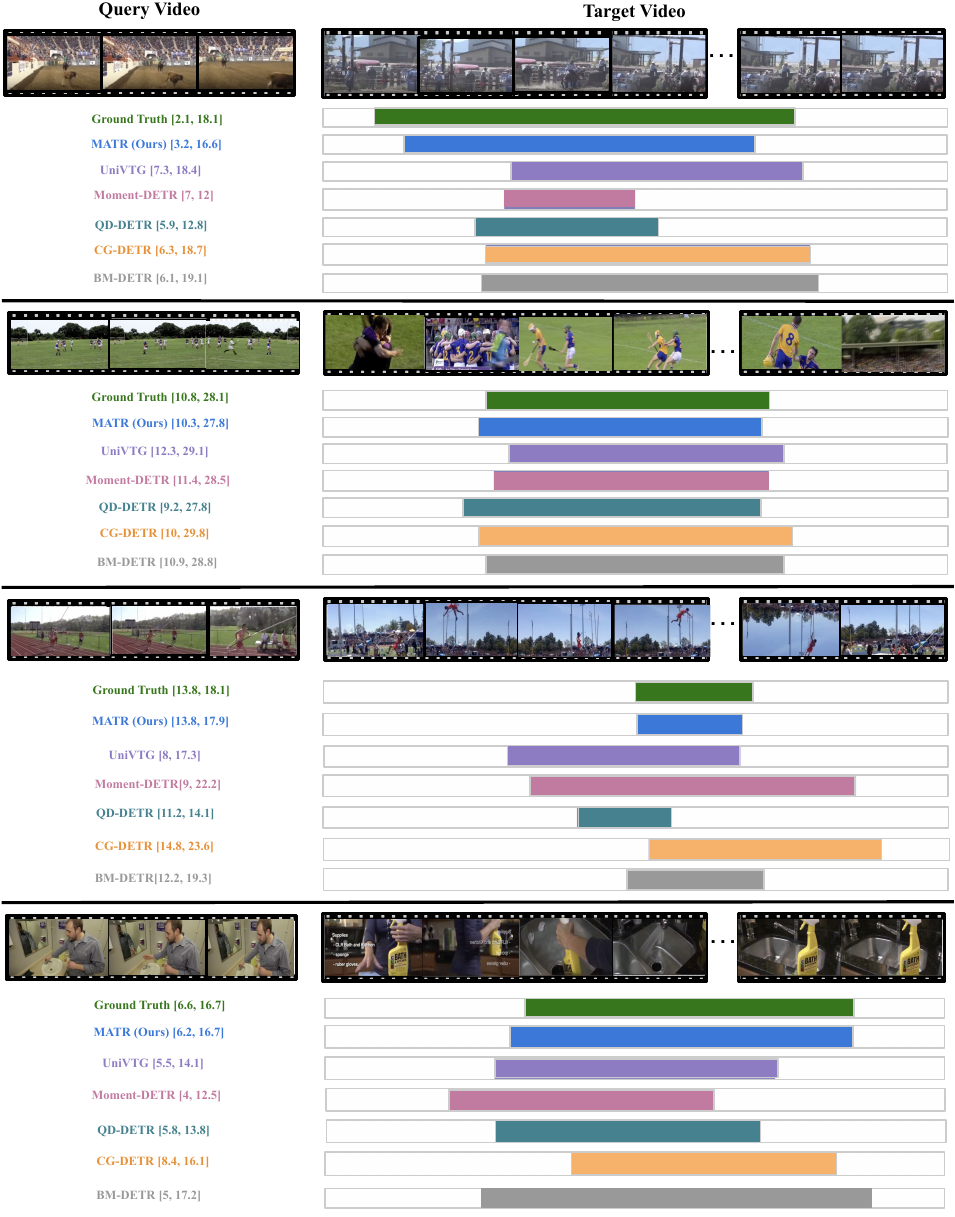}
  \caption[Qualitative Results]{Visualization of \emph{Vid2VidMR} on four query-target samples. Our proposed \modelloc{} model shows improved generalization capabilities over the best-performing baseline methods (variant (e), i.e., finetuned with video query input). The visualization demonstrates how our approach more accurately aligns with ground truth temporal boundaries across diverse activity types and durations. Start and end times for the ground truth and predicted moments are shown in the brackets. }

  \label{fig:iccv25_qual}
  \end{figure*}

\begin{table}[!t]
\scriptsize
    \centering
\begin{tabular}{l c c c c}
\toprule
 \multicolumn{1}{l}{}  & \multicolumn{2}{c}{\actdata{}}  & \multicolumn{2}{c}{\datas{}}\\
 \cmidrule(r){2-3} \cmidrule(r){4-5}
Feature & mIoU & R@1 & mIoU & R@1\\
 \midrule
    ResNet \cite{resnet} & 42.3 & 42.8 & 50.6& 48.1 \\
    I3D \cite{carreira2017quo}   & 43.7 & 43.4 & 52.2 & 50.6 \\
    S3D \cite{S3D} & 46.2 & 47.5 & 53.9& 52.1\\
    C(2+1)D \cite{C(2+1)D} & 46.6& 47.1 & 53.3 & 50.9\\
    CLIP \cite{clip} &  53.2 & 54.8 & 56.2 & 52.7 \\
    Slow-Fast \cite{SlowFast} & \textbf{54.7} & \textbf{55.9} & \textbf{58.3} & \textbf{55.4} \\
    CLIP \cite{clip} + Slow-Fast \cite{SlowFast} & 52.5 & 53.4 & 52.6 & 51.4 \\
 \hline
\end{tabular}
\caption[Comparison of Video Feature Extraction Backbones]{Comparison of Video Feature Extraction Backbones. Slow-Fast networks achieve superior performance across both datasets. This study was done with soft-DTW alignment and without pretraining.}
\label{tab:v_feat_abl}
\end{table}

\begin{table}[!t]
\scriptsize
    \centering
\begin{tabular}{l c c c c}
\toprule
 \multicolumn{1}{l}{}  & \multicolumn{2}{c}{\actdata{}}  & \multicolumn{2}{c}{\datas{}}\\
 \cmidrule(r){2-3} \cmidrule(r){4-5}
Noise Type & mIoU & R@1 & mIoU & R@1\\
 \midrule
    Frame Reverse &  42.7& 43.1 & 43.1& 44.6 \\
    Speed Variation & 53.3 & 55.9 & 57.4 & 55.3 \\
    Gaussian Noise & 55.8 & 56.3 & 58.5& 56.1\\
    None & \textbf{56.8}& \textbf{57.1}& \textbf{59.2}& \textbf{56.5}\\
 \hline
\end{tabular}
\caption[Robustness Analysis]{Robustness Analysis of \modelloc{} against Different Types of Perturbations. This study is done with CLIP features, soft-DTW alignment and with pre-training.}
\label{tab:robust_study}
\end{table}

\subsubsection{Predictions from Heads vs Alignment Matrices}
Table~\ref{tab:pred_study} shows that prediction heads outperform both pre-fusion and post-fusion alignment strategies across datasets. This highlights the limitations of direct alignment in capturing fine temporal boundaries. The transformer decoder refines coarse alignments and models complex temporal relationships, leading to significant gains in mIoU and R@1. While post-fusion alignment performs better than pre-fusion, suggesting that fused features improve cross-modal matching, both are modest compared to the improvements from prediction heads. These results emphasize the decoder’s critical role in transforming alignment signals into precise moment predictions.

\subsubsection{Ablation on Alignment Losses}
\modelloc{} can incorporate different alignment methods. Table \ref{tab:align_loss_abl} compares the performance of different alignment methods on both \actdata{} and \datas{} datasets. Our extensive ablation study comparing different alignment methods for \modelloc{} reveals that OTAM achieves the best performance across both datasets, with substantial improvements of 7.8\% and 10.1\% in R@1 over the baseline on \actdata{} and \datas{}, respectively. Temporal alignment is clearly beneficial, as all methods outperform the non-aligned baseline, with sophisticated approaches like OTAM and DTW showing particularly strong gains. Among the evaluated methods, OTAM consistently achieves the highest performance (58.2\% mIoU, 58.7\% R@1 on \actdata{} and 62.2\% mIoU, 58.6\% R@1 on \datas{}), followed closely by soft-DTW, which demonstrates the second-best results across both datasets. The performance improvements from simpler alignment methods like TCC and LCSS are more modest but still significant, confirming that the choice of alignment algorithm plays a crucial role in moment retrieval performance.

Although OTAM attains the best accuracy, we retain soft-DTW as the reference configuration throughout the remaining ablations. Soft-DTW is the method with which \modelloc{} was developed and the one for which this chapter gives the full derivation, and every other ablation in this section was run under it, so keeping it fixed leaves those comparisons mutually consistent. Re-running the full suite under OTAM would be unlikely to change any of the conclusions, since the two differ by a small and roughly constant offset. Soft-DTW is also preferable on its own merits in some settings: its $\gamma$-smoothed minimum aggregates over all alignment paths weighted by cost rather than committing to a single best path, which yields smoother gradients when frame-level similarities are noisy, and it is differentiable everywhere, which is the property that allows it to serve as $\mathcal{L}^{\text{pre}}_{\text{align}}$ and $\mathcal{L}^{\text{post}}_{\text{align}}$ and on which the dual-stage design depends.

\subsubsection{Ablation on Feature Backbones}
The ablation study in Table~\ref{tab:v_feat_abl} on feature backbones for \modelloc{} demonstrates that more advanced video feature extractors consistently yield better performance across both evaluated datasets. Specifically, Slow-Fast achieves the highest performance with 54.7\% mIoU and 55.9\% R@1 on the \actdata{} dataset, and 58.3\% mIoU and 55.4\% R@1 on the \datas{} dataset, showing a clear advantage over traditional architectures like ResNet and I3D. Interestingly, CLIP also performs strongly, particularly on the first dataset, suggesting that vision-language pre-training provides valuable transferable features for moment retrieval tasks. Surprisingly, the combination of CLIP and Slow-Fast features doesn't lead to further improvements and actually performs worse than either backbone individually, indicating potential feature redundancy or integration challenges that warrant further investigation in future work.

\subsubsection{Robustness study of \modelloc{}}
From the robustness study presented in Table~\ref{tab:robust_study}, we observe that our \modelloc{} demonstrates varying degrees of resilience to different types of noise. The model shows remarkable stability against Gaussian noise and speed variations, with performance drops of only 1.0\% and 3.5\% in mIoU, respectively, on \actdata{}, suggesting effective handling of these common real-world variations. Frame reversal causes a more significant performance decline (14.1\% mIoU drop on \actdata{}), indicating that temporal direction remains an important signal for the model. This comprehensive analysis demonstrates that while \modelloc{} is robust to several common video distortions, it appropriately remains sensitive to semantic query relevance.

The asymmetry between speed variation and frame reversal follows directly from the design. DTW exists to absorb local accelerations and decelerations, so a query performed faster or slower than the target is handled by the warping path; frame reversal, by contrast, violates the monotonicity constraint that all valid alignments enforce and cannot be represented at all. Both perturbations also appear as augmentations during self-supervised pre-training, which further explains the speed robustness. Perspective mismatch between query and target was not isolated under controlled perturbation, but it is present throughout the evaluation rather than absent from it: \datas{} is built from broadcast footage, and its longer query clips consist of overlapping views of the same action including replays from different camera angles. The qualitative results below make the same point on \actdata{}, where \modelloc{} localises correctly under significant viewpoint change, across multi-shot sequences with varying camera angles, and between different environments.

\subsubsection{Qualitative Results}
We present moment retrieval results on four samples in Fig.~\ref{fig:iccv25_qual}, where we compare our proposed approach to variant (e) (refer to Sec.\ref{sec:iccv_base} for details) of the four best-performing baselines. Our proposed \modelloc{} model exhibits the highest overlap with the ground truth for the given query, accurately identifying the correct temporal boundaries. Among the baseline methods, BM-DETR and UniVTG achieve strong overlap; however, they are still outperformed by our approach. \

\subsubsection{Efficiency Analysis}
\label{subsec:matr_efficiency}
Since the dual-stage design invokes sequence alignment twice, once before and once after the encoder, and each stage is $O(MN)$ in the query and target sequence lengths, it is reasonable to ask whether alignment dominates the cost. Table~\ref{tab:matr_efficiency} shows that it does not. All measurements are taken on an NVIDIA A6000 GPU with batch size 1, averaged over 100 inference runs. \modelloc{} requires 15.4 GFLOPs and 58.6M parameters, with 53.8\,ms inference and 2.5\,GB peak memory. It is therefore cheaper than BM-DETR (23.8 GFLOPs, 81.3M parameters, 71.5\,ms, 3.8\,GB) and CG-DETR (18.1 GFLOPs, 58.4\,ms, 3.1\,GB), both of which are single-stage, while achieving 9.2\% higher mIoU than BM-DETR. Against BM-DETR specifically, \modelloc{} uses 27.9\% fewer parameters, 35.3\% fewer FLOPs and 34.2\% less memory, and is 24.7\% faster. The reason alignment does not dominate is that it operates on sequences of at most a few hundred sampled frames rather than on raw video.

\begin{table}[!t]
\scriptsize
\centering
\setlength{\tabcolsep}{4pt}
\begin{tabular}{@{}lccccc@{}}
\toprule
Method & Params & FLOPs & Inference & Memory & Throughput\\
 & (M) & (G) & Time (ms) & (GB) & (FPS)\\
\midrule
Moment-DETR & 45.2 & 12.3 & 48.5 $\pm$ 2.1 & 2.1 & 20.6\\
QD-DETR    & 62.8 & 16.7 & 52.3 $\pm$ 2.4 & 2.6 & 19.1\\
UniVTG     & 38.4 & 10.2 & 45.8 $\pm$ 2.0 & 2.0 & 21.8\\
CG-DETR    & 69.5 & 18.1 & 58.4 $\pm$ 2.6 & 3.1 & 17.1\\
BM-DETR    & 81.3 & 23.8 & 71.5 $\pm$ 3.5 & 3.8 & 14.0\\
\midrule
\modelloc{} (Ours) & 58.6 & 15.4 & 53.8 $\pm$ 2.3 & 2.5 & 18.6\\
\bottomrule
\end{tabular}
\caption[Efficiency analysis of MATR]{Efficiency analysis of \modelloc{} and baselines. All measurements conducted on an NVIDIA A6000 GPU with batch size 1, averaged over 100 inference runs.}
\label{tab:matr_efficiency}
\end{table}

\begin{table}[!t]
\scriptsize
\centering
\setlength{\tabcolsep}{4pt}
\begin{tabular}{@{}lcccccc@{}}
\toprule
\multirow{2}{*}{Sampling Rate} & \multicolumn{2}{c}{\actdata{}} & \multicolumn{2}{c}{\datas{}} & Inference & Memory\\
\cmidrule(lr){2-3} \cmidrule(lr){4-5}
 & mIoU & R@1 & mIoU & R@1 & Time (ms) & (GB)\\
\midrule
1 frame/4s & 51.3 & 52.1 & 53.8 & 50.2 & 30.1 & 1.7\\
1 frame/3s & 54.2 & 55.4 & 56.4 & 53.6 & 41.2 & 2.1\\
1 frame/2s (default) & \textbf{56.8} & \textbf{57.1} & \textbf{59.2} & \textbf{56.5} & 53.8 & 2.5\\
1 frame/1s & 56.2 & 56.8 & 59.5 & 56.1 & 95.6 & 4.4\\
1.5 frames/s & 55.8 & 56.3 & 58.1 & 55.4 & 137.4 & 6.3\\
2 frames/s & 55.4 & 55.9 & 57.6 & 54.8 & 178.9 & 8.1\\
\bottomrule
\end{tabular}
\caption[Impact of frame sampling rate]{Impact of frame sampling rate on \modelloc{} performance and cost. Inference time and memory measured on an NVIDIA A6000 GPU.}
\label{tab:matr_sampling_rate}
\end{table}

\subsubsection{Cost as a Function of Sequence Length}
Table~\ref{tab:matr_sampling_rate} reports how accuracy and cost vary with the frame sampling rate, which controls the sequence length $M$ for a target video of fixed duration. Inference time ranges from 30.1\,ms at 1 frame/4s to 178.9\,ms at 2 frames/s, and peak memory from 1.7\,GB to 8.1\,GB. The default rate of 1 frame/2s sits at the accuracy knee: doubling the rate to 1 frame/1s raises inference time by 77.7\% and memory by 76.0\% for a 0.3\% change in mIoU, and denser sampling degrades accuracy through redundancy between highly correlated adjacent frames. Sparser sampling at 1 frame/4s costs 5.5\% mIoU on \actdata{}, indicating that critical motion cues are missed. Under a fixed 48\,GB budget at the default resolution, this corresponds to a maximum processable target duration of roughly 8 minutes at the default rate, extending to roughly 33 minutes at 1 frame/4s. All \actdata{} (max 191.5\,s) and \datas{} (max 208.8\,s) videos are therefore comfortably within range at the default setting.

\subsubsection{Failure Case Analysis}
\label{subsec:matr_failure}
To characterise the practical limitations of \modelloc{}, we inspected the 100 test samples with the lowest mIoU and identified three dominant failure modes. First, \emph{perceptually different action representations}, where the same action appears visually distinct across query and target, for example normal versus seated rope skipping, producing large cross-instance feature distances that weaken alignment. Second, \emph{viewpoint variation}, where query and target are recorded from substantially different perspectives, such as a third-person view against a first-person view. Third, \emph{speed variation}, where the action is performed at noticeably different speeds, for example a normal-speed pole vault against a slow-motion replay, causing soft-DTW to match incorrectly stretched sub-sequences. These modes are consistent with the robustness results reported above and indicate that viewpoint-invariant and speed-invariant representations remain the principal open problem for this task.

\subsubsection{Cross-Domain Generalisation}
\label{subsec:matr_crossdomain}
To assess robustness to cross-domain appearance shift, we converted the query videos in the test set into cartoon/anime-style footage using DCT-Net, leaving the target videos unchanged. \modelloc{} achieves 38.3\% mIoU under this setting, despite the frozen CLIP encoder never having seen cartoon-style imagery during training. This indicates that the model retains meaningful localisation ability even under a large stylistic gap between query and target, while the drop relative to the unperturbed setting shows that appearance-invariant and domain-adaptive representation learning remains an important direction for future work.

\section{Summary}
In this chapter, we introduced \modelloc{}, a well-motivated approach for \emph{Vid2VidMR}. By combining the abstract representation of the encoder and the refined fine-grained features of the decoder conditioned on aligned query features, \modelloc{} effectively captures both semantic and temporal information, enabling accurate moment localization. Our self-supervised pre-training strategy further improves the model initialization and boosts performance. Extensive experiments showed that \modelloc{} outperforms strong baselines on both \actdata{} and our novel \datas{} dataset, showcasing its effectiveness in tackling this challenging task. Although \modelloc{} excels at localizing moments using video queries, future research can extend it to multimodal queries and develop more efficient architectures to enhance scalability for \emph{Vid2VidMR}.

The following chapter presents the conclusion of our work, discusses its limitations, and outlines potential directions for future research.

\chapter{Conclusions and Future Scope} \label{chap7conclusionFW}

This thesis advances \textit{query-guided video understanding} across three complementary levels, object, relationship, and moment, enabling machines to answer \textit{where}, \textit{how}, and \textit{when} in open-world videos with minimal supervision. The motivation for this work is twofold. First, real-world deployments often encounter long-tailed and continuously evolving visual concepts, making exhaustive annotation and closed-set training impractical. Second, users increasingly demand targeted, example-driven retrieval, locating a specific object, interaction, or moment, without manually scanning long videos. To address these challenges, this thesis proposes query-driven grounding frameworks that leverage minimal supervision, temporal reasoning, and query-based generalization to perform robust video understanding.

\section{Summary and Discussion}

At the object level (\textit{understanding ``where''}), this thesis advances video object grounding through two complementary frameworks designed to localize novel objects in videos using minimal supervision. For the one-shot scenario, we introduce QDETRv (Query-Guided DETR for Videos, a transformer-based architecture that extends DETR with temporal modeling capabilities. QDETRv employs (i) query–context fusion, where query features are integrated with contextual video frames to enable effective temporal reasoning; (ii) a memory module that maintains temporal consistency across frames; and (iii) an unsupervised pretraining strategy using synthetic object trajectories with recurrent queries and feature reconstruction objectives. These components collectively enhance temporal modeling and generalization, allowing QDETRv to achieve substantial gains over competitive baselines. We further extend one-shot object grounding to few-shot object grounding. To this end, we proposed the Temporal Object-Aware Vision Transformer, which leverages large-scale, language-aligned, pre-trained vision encoders. This framework integrates semantic knowledge from vision–language models like OWL-ViT to enable effective knowledge transfer and robust object generalization over time. The model introduces a filtering mechanism that propagates high-confidence object features across frames, enabling efficient progression and reduced noise. It further incorporates feature interaction between current frames and past detections to maintain temporal coherence without explicit object-tube proposals, which are often computationally expensive and require task-specific training. Parallel classification and detection heads are optimized for few-shot settings, achieving state-of-the-art performance on FSVOD-500, FSYTV-40, VidOR, and VidVRD benchmarks. Together, these contributions advance the understanding of \textit{``where''} objects are located in videos.

At the relationship level (\textit{understanding ``how''}), this thesis extends individual object grounding to model interactions between objects through few-shot referring relationship grounding in videos. The task is formulated as inference over a $T$-partite random field, where $T$ denotes the number of frames in the video, and each vertex represents candidate subject–object pairs. Meta-learned potentials and query-conditioned translational embeddings are introduced to encode relational semantics and capture meaningful interactions between entities. The model achieves improved subject–object–predicate localization accuracy, with ablation studies validating the complementary roles of Global Semantic Aggregation (GSA), which captures long-range contextual dependencies, and Local Localization Aggregation (LLA), which focuses on fine-grained spatial cues and temporal consistency. Despite these advances, grounding dynamic predicates with complex temporal evolution remains challenging, and performance is influenced by the quality of subject–object proposals and the computational cost of message passing across frames. Future extensions could involve joint optimization of proposals and relationships using tubelet-level candidates, sparse or neural message passing for improved scalability, and end-to-end integration with open-vocabulary detectors to enhance better generalization to unseen subjects and objects. Moreover, extending the formulation to include temporal-tubes of objects for predicates would naturally connect ``how” and ``when,” enabling richer and temporally grounded scene graphs that capture evolving visual relationships over time.

At the moment level (\textit{understanding when}), the thesis introduces the Moment Alignment Transformer (MATR) for video-to-video moment retrieval. MATR aligns a short query clip with a longer target video using dual-stage sequence alignment, employing differentiable Dynamic Time Warping before and after fusion to refine temporal correspondences. The model achieves notable gains in mIoU and Recall@1 on ActivityNet-VRL and SportsMoments, demonstrating the importance of explicit sequence alignment and task-aligned self-supervised pretraining. However, computational cost increases with sequence length, and the current design also focuses on single-moment retrieval, overlooking multi-event queries. We further extended MATR to support text-to-video moment retrieval and achieved comparable, if not better, performance to the state-of-the-art methods. Future extensions could include hierarchical coarse-to-fine retrieval for efficiency, non-monotonic or skip-aware alignment for flexibility, and multimodal query fusion with text or audio to improve disambiguation and generalization. The restriction to single-moment retrieval deserves particular comment, since it follows the benchmark specification rather than a limitation of the architecture: Vid2VidMR as formalised in the literature, and as evaluated on ActivityNet-VRL, assumes one ground-truth moment per query--target pair. Two routes would lift it. The lighter of the two requires no retraining at all: a long target video could be partitioned into overlapping chunks, with MATR run on each chunk against the same query and predictions above a confidence threshold pooled across chunks to yield multiple moments. This would be parallel across chunks and would bound peak memory by chunk length rather than video length, at the cost of truncating moments that straddle a chunk boundary and requiring per-chunk scores to be normalised before pooling. The principled route would instead replace the single boundary head with a set-prediction head over the $l$ learnable queries that MATR already carries, trained with Hungarian matching against a variable number of ground-truth moments; this would also require a benchmark providing multi-moment ground truth, which neither dataset used here currently does.

\subsection{Transferability of Components Across Tasks}
As the three tasks were formulated and solved independently, experimentation revealed that two components were transferable across them. Language-aligned visual backbones from the CLIP family were useful at two of the three levels, as OWL-ViT in the few-shot object framework and as a frozen CLIP encoder in MATR; in both cases the gain came from semantic priors that the task-specific training data could not supply, in the object setting for separating visually similar categories and in the moment setting as the strongest single backbone on ActivityNet-VRL. Object-level temporal state propagation also transferred: the recurrent-query memory module in QDETRv and the confidence-thresholded feature propagation in the few-shot framework are the same principle at different levels of maturity, the second having been designed directly from the limitations of the first. Two components did not transfer, for structural rather than empirical reasons. The dual-stage alignment losses presuppose two sequences to align, whereas in the object and relationship tasks the query is a single image or a triplet, so there is no second sequence and the loss has no natural analogue. Belief propagation over the $T$-partite field is likewise tied to the pairwise subject--object candidate structure of the relationship task and has no counterpart in single-entity grounding. Cross-task joint training was not attempted, and closing that gap is precisely the aim of the unified architecture outlined in Section~\ref{sec:future_work}.

Overall, these contributions collectively demonstrate that temporal reasoning and query conditioning enable flexible, open-world grounding in videos under minimal supervision, forming a scalable foundation for structured video understanding.

\subsection{Training Cost and Reproducibility}
The computational requirements of each contribution are distributed across the implementation sections of the individual chapters, which makes the overall cost hard to assess; we therefore summarise them here. QDETRv is pretrained on synthetic trajectories over UCF101 together with image-level pretraining on ImageNet, and is trained on three NVIDIA RTX A6000 GPUs using Adam with a learning rate of 1e-5 for 200 epochs at a batch size of 350, initialised from an ImageNet-pretrained ResNet-50. The few-shot object framework builds on OWL-ViT ViT-L/16 used off the shelf, and converges in approximately four days on two A6000 GPUs using AdamW with weight decay 0.01 and a cosine schedule with linear warmup. The few-shot relationship framework is meta-trained episodically on support sets rather than pretrained on a separate corpus. MATR is pretrained on unlabeled Kinetics-700 with a frozen CLIP ViT-B/32 encoder and trained on two A6000 GPUs with AdamW at a learning rate of 1e-4. For reproducibility, the category-disjoint VidOR and VidVRD splits used throughout the object and relationship chapters are themselves a contribution of this thesis and are released alongside the code. For SportsMoments, the source video identifiers are listed in this chapter, and we additionally release the segment-level timestamp annotations and the pre-computed validation and test pair lists, which is what makes the benchmark reproducible in practice rather than merely re-collectable.

\section{Potential Applications}

The proposed methods have broad applicability across domains where query-based grounding can replace exhaustive manual analysis of videos. In sports analytics, these models can automatically extract highlights, track players, and retrieve tactical plays from exemplar clips. In manufacturing, they can detect defective components or rare fault patterns from a few visual examples, enabling rapid adaptation to new products. In education and skill learning, exemplar-based moment retrieval supports indexing and feedback in instructional videos, helping systems identify critical steps such as ``when the weld begins” or ``when the coffee is poured.” In healthcare, temporal grounding can detect and timestamp key events such as falls or procedural deviations from limited annotated data. In security and forensics, query-based localization allows analysts to locate specific objects or interactions across long surveillance videos with minimal supervision.

Together, these applications highlight the versatility of query-driven grounding systems, where users specify \textit{what to find} via queries rather than predefined class names, enabling adaptive and efficient video understanding across diverse domains.

\section{Future Directions}
\label{sec:future_work}
In this section, we outline research directions that extend and broaden query-guided video understanding.

\subsection{Unified “Any-Query, Any-Grounding”}
Our current methods have shown remarkable performance on three video grounding tasks: QDETRv handles image queries for object grounding, the relationship grounding framework processes visual triplets, and MATR operates on video/text queries for temporal grounding. One of the promising future directions of this thesis is to develop a unified architecture that can handle multiple query modalities such as image, video, and text, and generate diverse grounding outputs such as boxes, masks, relation spans, and moments through a shared backbone with task-agnostic query-target fusion. This would involve using a modality-agnostic encoder, similar to CLIP or Video-OWL, to process different types of queries while encoding the target video with temporal tokens. Dual-stage alignment could then be used to produce query-aligned target features that are consistently reused across all prediction heads. Building on this shared representation, the system could support multiple heads for detection, segmentation, relation trajectories, and moment boundaries, effectively extending the capabilities of architectures like QDETRv and MATR. Finally, incorporating class-agnostic matching and open-vocabulary priors, together with few-shot prototype heads and temporal filtering, would enable robust handling of uncertainty and open-set scenarios. However, we note that such an extension is non trivial and would require require additional rigorous experimental studies involving architectural integration, joint training, and appropriate multi-task evaluation, which is beyond the scope of the current thesis.

\subsection{Efficiency and Streaming Video}
A key challenge moving forward is to scale grounding to live video streams and long-duration content while operating under tight computational budgets. One promising direction is to explore streaming transformers that maintain causal state caching, along with keyframe-adaptive token sparsification and memory-augmented decoders that reuse high-confidence tokens across time to enable low-latency updates. We have taken initial steps in both directions: language-guided temporal token pruning reduces the number of video tokens a VideoLLM must process~\cite{kumar2025tokenpruning}, and a strictly causal state-space core supports streaming video anomaly detection~\cite{kumar2026streamingvad}. These ideas naturally extend the object-aware temporal propagation strategies used in few-shot VOD. Another opportunity lies in using lightweight prefilters that identify candidate windows based on query-conditioned signatures before applying more expensive grounding modules.

A related and largely untested axis is generalisation to fully unstructured, in-the-wild footage. The benchmarks used in this thesis are less curated than they may appear: VidOR is built from user-generated social-media video with handheld camera motion and clutter, FSVOD-500 and FSYTV-40 draw from YouTube, and SportsMoments consists of full-length broadcast matches with panning, cuts, replays and on-screen graphics. Category shift is also tested by construction throughout, since every benchmark uses class-disjoint train and test splits. What remains genuinely untested is egocentric video, where the camera is itself the actor and ego-motion, hand occlusion and objects leaving frame are constant, together with untrimmed long-form video at the scale of hours rather than the few-minute targets considered here. We would expect the few-shot object framework to degrade most gracefully, since its open-vocabulary priors are domain-robust and confidence-thresholded propagation naturally discards unreliable frames, and QDETRv to suffer most, given its documented sensitivity to occlusion and small objects. The relationship framework would be limited by proposal quality under ego-motion rather than by the random field itself, and MATR by the growth of the retrieval space in long-form video rather than by camera motion. Egocentric benchmarks such as Ego4D, whose episodic-memory and moment-queries tasks map closely onto the \emph{where} and \emph{when} formulations of this thesis, are the natural setting in which to test these expectations.

\subsection{Directions with Video LLMs}
Contemporary Video LLMs show considerable promise, yet they still fall short of specialized grounding methods when precise localization and timestamping are required, especially in the absence of strong video-to-video alignment or object-aware temporal modeling; a survey of current VideoLLM benchmarks and evaluation practices is presented in~\cite{kumar2025videollmsurvey}. A promising direction is to adopt a hybrid approach, in which Video LLMs handle intent parsing, disambiguation, and iterative refinement of the user’s query, while the actual fine-grained grounding is delegated to dedicated modules, such as object-aware few-shot VOD and dual alignment Vid2VidMR proposed in this thesis. Making such hybrid systems trustworthy also requires verifying that the evidence a Video LLM cites in its answers is actually grounded in the video~\cite{kumar2026hallucinated}.

In summary, this thesis establishes a foundation for query-driven, open-world video grounding and highlights the path toward unified, efficient, systems that can seamlessly process long, dynamic, and complex videos. Continued research along these directions promises not only better models but also more practical video understanding for real-world applications.

\chapter*{List of Publications}
\addcontentsline{toc}{chapter}{List of Publications}

\section*{Publications forming part of this thesis}

\begin{enumerate}[label={[\arabic*]}, leftmargin=*, itemsep=4pt]
  \item \textbf{Yogesh Kumar}, Anand Mishra.
    \textit{Few-Shot Referring Relationships in Videos}.
    In Proceedings of the IEEE/CVF Conference on Computer Vision and Pattern Recognition (CVPR), pages 2289--2298, 2023~\cite{Kumar023-2}.

  \item \textbf{Yogesh Kumar}, Saswat Mallick, Anand Mishra, Sowmya Rasipuram, Anutosh Maitra, Roshni Ramnani.
    \textit{QDETRv: Query-Guided DETR for One-Shot Object Localization in Videos}.
    In Proceedings of the AAAI Conference on Artificial Intelligence (AAAI), volume 38, pages 2831--2839, 2024~\cite{Kumar2024}.

  \item \textbf{Yogesh Kumar}, Uday Agarwal, Manish Gupta, Anand Mishra.
    \textit{Aligning Moments in Time using Video Queries}.
    In Proceedings of the IEEE/CVF International Conference on Computer Vision (ICCV), pages 20215--20225, 2025~\cite{kumar2025matr}.

  \item \textbf{Yogesh Kumar}, Anand Mishra.
    \textit{Temporal Object-Aware Vision Transformer for Few-Shot Video Object Detection}.
    In Proceedings of the AAAI Conference on Artificial Intelligence (AAAI), volume 40, pages 5764--5772, 2026~\cite{kumar2026fsvit}.

  \item \textbf{Yogesh Kumar}, Uday Agarwal, Manish Gupta, Anand Mishra.
    \textit{Moment Alignment Transformer for Video-to-Video Moment Retrieval}.
    Computer Vision and Image Understanding (CVIU), article 104959, 2026~\cite{kumar2026matrcviu}.
\end{enumerate}

\section*{Other publications during the PhD (not part of this thesis)}

\begin{enumerate}[label={[\arabic*]}, leftmargin=*, itemsep=4pt, resume]
  \item Shreya Shukla, Prajwal Gatti, \textbf{Yogesh Kumar}, Vikash Yadav, Anand Mishra.
    \textit{Towards Making Flowchart Images Machine Interpretable}.
    In Proceedings of the International Conference on Document Analysis and Recognition (ICDAR), pages 505--521, 2023~\cite{floco}.

  \item Uday Agarwal, \textbf{Yogesh Kumar}, Abu Shahid, Prajwal Gatti, Manish Gupta, Anand Mishra.
    \textit{CHAPVIDMR: Chapter-Based Video Moment Retrieval using Natural Language Queries}.
    In Proceedings of the Indian Conference on Computer Vision, Graphics and Image Processing (ICVGIP), pages 1--9, 2024~\cite{chapvidmr}.

  \item \textbf{Yogesh Kumar}.
    \textit{VideoLLM Benchmarks and Evaluation: A Survey}.
    arXiv preprint arXiv:2505.03829, 2025~\cite{kumar2025videollmsurvey}.

  \item \textbf{Yogesh Kumar}.
    \textit{Language-Guided Temporal Token Pruning for Efficient VideoLLM Processing}.
    In Proceedings of the Conference on Empirical Methods in Natural Language Processing (EMNLP), Main Conference, 2025~\cite{kumar2025tokenpruning}.

  \item \textbf{Yogesh Kumar}.
    \textit{Catching Hallucinated Citations in Video-LLM Question Answering: A Self-Verification Pipeline and Verifier Ablation Study}.
    arXiv preprint arXiv:2608.15574, 2026~\cite{kumar2026hallucinated}.

  \item \textbf{Yogesh Kumar}.
    \textit{Strictly Causal Streaming Video Anomaly Detection with a Theoretically-Grounded State-Space Core}.
    arXiv preprint arXiv:2608.24810, 2026~\cite{kumar2026streamingvad}.
\end{enumerate}

\begingroup
\setlength{\parskip}{2pt}
\bibliographystyle{unsrt}
\bibliography{references}
\endgroup

\end{document}